\documentclass[runningheads]{llncs}

\usepackage[mobile]{eccv}

\pdfpagewidth=\paperwidth
\usepackage{eccvabbrv}

\usepackage{colortbl}
\usepackage{pifont}
\usepackage{xspace}

\usepackage{wrapfig}

\usepackage{amssymb}
\usepackage{bbm}
\usepackage{glossaries}
\usepackage{adjustbox}

\usepackage{tabularx}
\usepackage{booktabs}
\usepackage{siunitx}
\usepackage{multirow}
\usepackage{caption}
\usepackage{bm}
\usepackage{fontawesome5}
\newcolumntype{Y}{>{\centering\arraybackslash}X} 

\definecolor{ouryellow}{RGB}{255,176,0}
\definecolor{ourorange}{RGB}{254,97,0}
\definecolor{ourpink}{RGB}{220,38,127}
\definecolor{ourpurple}{RGB}{120,94,240}
\definecolor{ourblue}{RGB}{100,143,255}
\definecolor{iceblue}{RGB}{165, 214, 217}

\def\method{DINOvTree\xspace}
\def\benchmark{BIRCH-Trees\xspace}

\newacronym{cnn}{CNN}{Convolutional Neural Network}
\newacronym{vit}{ViT}{Vision Transformer}
\newacronym{sota}{SOTA}{state of the art}
\newacronym{vfm}{VFM}{Vision Foundation Model}
\newacronym{uav}{UAV}{Unoccupied Aerial Vehicle}
\newacronym{chm}{CHM}{Canopy Height Model}
\newacronym{dsm}{DSM}{Digital Surface Model}
\newacronym{dtm}{DTM}{Digital Terrain Model}
\newacronym{dwa}{DWA}{Dynamic Weight Average}
\newacronym{dbh}{DBH}{Diameter at Breast Height}
\newacronym{mlp}{MLP}{Multilayer Perceptron}
\newacronym{mtl}{MTL}{Multi-Task Learning}
\newacronym{f1}{F1}{F1-Score}
\newacronym{acc}{Acc}{Accuracy}
\newacronym{msle}{MSLE}{Mean Squared Logarithmic Error}
\newacronym{mae}{MAE}{Mean Absolute Error}
\newacronym{rmse}{RMSE}{Root Mean Squared Error}

\newcommand{\ulint}[1]{%
    \underline{\tablenum[table-format=3]{#1}}%
}

\newcommand{\ulinttwo}[1]{%
    \phantom{0}\underline{\tablenum[table-format=2]{#1}}%
}

\usepackage{calc}

\newcommand{\bfmain}[2]{%
    {\makebox[\widthof{\tablenum[table-format=2.2]{#1}}][r]{\bfseries\boldmath\tablenum[table-format=2.2, reset-text-series=false, reset-math-version=false]{#1}}${}\pm{}$\tablenum[table-format=1.2]{#2}}%
}

\newcommand{\bfpadded}[2]{%
    {\phantom{0}\makebox[\widthof{\tablenum[table-format=1.2]{#1}}][r]{\bfseries\boldmath\tablenum[table-format=1.2, reset-text-series=false, reset-math-version=false]{#1}}${}\pm{}$\tablenum[table-format=1.2]{#2}}%
}

\newcommand{\bfmainsmall}[2]{%
    {\makebox[\widthof{\tablenum[table-format=1.2]{#1}}][r]{\bfseries\boldmath\tablenum[table-format=1.2, reset-text-series=false, reset-math-version=false]{#1}}${}\pm{}$\tablenum[table-format=1.2]{#2}}%
}

\newcommand{\ulmain}[2]{%
    {\underline{\tablenum[table-format=2.2]{#1}}${}\pm{}$\tablenum[table-format=1.2]{#2}}%
}

\newcommand{\ulpadded}[2]{%
    {\phantom{0}\underline{\tablenum[table-format=1.2]{#1}}${}\pm{}$\tablenum[table-format=1.2]{#2}}%
}

\newcommand{\ulmainsmall}[2]{%
    {\underline{\tablenum[table-format=1.2]{#1}}${}\pm{}$\tablenum[table-format=1.2]{#2}}%
}

\usepackage[accsupp]{axessibility}  

\usepackage{hyperref}

\usepackage{fancyhdr}

\fancypagestyle{firststyle}{%
\fancyhf{}%
\lfoot{%
\raisebox{0.35em}[0pt][0pt]{%
\parbox[b]{\textwidth}{
{\color{black!30}\rule{\textwidth}{0.4pt}}\par\vspace{-0.1em}%
\fontsize{7}{8.3}\selectfont\color{black!70}%
This preprint has not undergone peer review (when applicable) or any post-submission improvements or corrections. The Version of Record of this contribution is published in Computer Vision -- ECCV 2026, and is available online at \url{https://doi.org/10.1007/978-3-032-37372-4_4}.%
}%
}%
}
}

\usepackage{orcidlink}

\begin{document}

\title{Estimating Individual Tree Height and Species from UAV Imagery}


\author{
Jannik Endres\inst{1}\orcidlink{0009-0005-3995-8894} \quad
Etienne Laliberté\inst{2,1}\orcidlink{0000-0002-3167-2622} \quad \\
David Rolnick\inst{1,3}\orcidlink{0000-0002-2855-393X} \quad
Arthur Ouaknine\inst{1,3}\orcidlink{0000-0003-1090-6204}
}

\authorrunning{J.~Endres \etal}

\institute{
\textsuperscript{1}Mila – Quebec AI Institute \quad
\textsuperscript{2}Université de Montréal \quad
\textsuperscript{3}McGill University\\
\url{https://RolnickLab.github.io/DINOvTree}
}

\maketitle

\thispagestyle{firststyle}

\begin{abstract}
Accurate estimation of forest biomass, a major carbon sink, relies heavily on tree-level traits such as height and species.
\glspl{uav} capturing high-resolution imagery from a single RGB camera offer a cost-effective and scalable approach for mapping and measuring individual trees.
We introduce \benchmark, the first benchmark for individual tree height and species estimation from tree-centered UAV images, spanning three datasets: temperate forests, tropical forests, and boreal plantations.
We also present \method, a unified approach using a \gls{vfm} backbone with task-specific heads for simultaneous height and species prediction.
Through extensive evaluations on \benchmark, we compare \method against commonly used vision methods, including \glspl{vfm}, as well as biological allometric equations.
We find that \method achieves top overall results with accurate height predictions and competitive classification accuracy while using only $54\%$ to $58\%$ of the parameters of the second-best approach.


\end{abstract}
\section{Introduction}

Forests cover 4.14 billion hectares, $32\%$ of global land area \cite{fao_global_2025}, and play a critical role as biodiversity habitats and major carbon sinks \cite{pan2011large}. 
Estimating the biomass of individual tree species in forests is essential both for ecological monitoring and for quantifying the carbon stored in mature forests \cite{davies2021forestgeo} and restoration projects \cite{camarretta2020monitoring}.
Tree species and \gls{dbh}, a measure of trunk width, are the two most important variables for biomass estimation.
Both are traditionally measured by field workers, a process not scaling well.
While satellite imagery is widely used in remote sensing settings, it lacks the spatial resolution needed for individual tree trait estimation.
Drones offer a cost-effective alternative, capturing high-resolution imagery to monitor individual trees at large scales \cite{Laliberte2025.09.02.673753}.
However, \gls{dbh} is not directly observable from drones, and large-scale datasets for direct estimation remain unavailable.
While LiDAR yields direct tree height measurements, a valuable proxy for \gls{dbh} \cite{jucker2022tallo}, its cost limits deployment at scale, particularly in tropical regions where resources are constrained and forest knowledge is most incomplete \cite{phillips2023sensing}.
\begin{wrapfigure}[19]{r}{0.38\textwidth}
    \centering
    \includegraphics[width=0.92\linewidth]{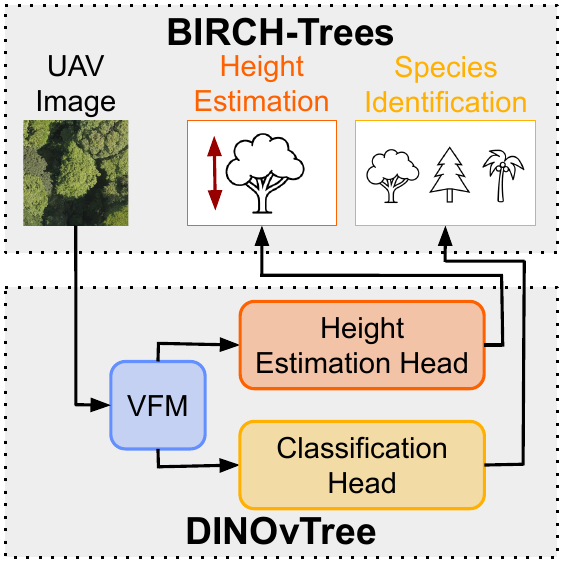}
    \caption{
    \textbf{Overview of our benchmark \benchmark and method \method.}
    \benchmark is a benchmark for joint individual tree height estimation and species identification from \gls{uav} images.
    \method includes a \gls{vfm} backbone with two task-specific heads.\label{fig:teaser}}
\end{wrapfigure}
Thus, here we estimate height from RGB drone imagery, a cost-effective alternative.

Recent work has successfully addressed generalized tree crown detection \cite{baudchon2025selvabox} and segmentation \cite{duguay2026selvamask} on high-resolution drone imagery, achieving robust localization even on out-of-distribution datasets.
However, the joint task of localizing trees and identifying their species remains a significant challenge due to persistent issues such as imbricated crowns and long-tailed class distributions \cite{teng2025bringing}.
Individual tree trait prediction isolates classification and height challenges (\eg, long-tailed distribution) while avoiding the complexity of delineating all trees in a forest.
Significant challenges remain: high inter-class similarity and high intra-class variability complicate species identification, while height estimation is fundamentally ill-posed without ground-level visibility.

To advance research in individual tree trait prediction, we introduce \benchmark (\textit{\underline{B}enchmark for \underline{I}ndividual \underline{R}ecognition of \underline{C}lass and \underline{H}eight of \underline{Trees}}), illustrated in \cref{fig:teaser}.
We focus on individual tree height estimation and species identification, formulated as regression and classification tasks, respectively.
To build our benchmark, we use RGB drone images of tree canopies from three environments: \emph{(i)} temperate forest, \emph{(ii)} tropical forest, and \emph{(iii)} boreal plantation.
We evaluate common \glspl{cnn}, \glspl{vit} and modern \glspl{vfm}.
Furthermore, we demonstrate the limitations of traditional in-domain allometric equations, which are commonly employed for height estimation in the field, by showing their inferior results compared to learned approaches.
To efficiently tackle both tasks, we propose \method, a novel multi-task method for tree species identification and height estimation (see \cref{fig:teaser}).
Given the scarcity of domain-specific data, \method leverages the representation of a fine-tuned \gls{vfm} as its shared backbone.
This shared architecture significantly reduces the parameter count and consequently the computational cost compared to separate task-specific models.
The separate heads in \method incorporate a cross-attention mechanism to effectively attend to the most task-relevant regions within the image.

Specifically, our main contributions are as follows:
\textbf{(1)} We construct \benchmark, a benchmark for individual tree height estimation and species identification based on single RGB UAV images from three existing datasets.
\textbf{(2)} We conduct an exhaustive evaluation of existing vision models on both tasks, demonstrating the superiority of modern \glspl{vfm} compared to commonly employed methods.
\textbf{(3)} Our method, \method, achieves state-of-the-art height estimation results on \benchmark and competitive classification results while using only $54\%$ to $58\%$ of the parameters of competing methods.

\section{Related Work}
\label{sec:related_work}

\paragraph{Tree height estimation.}
Tree height has traditionally been estimated from crown radius using species-specific allometric equations~\cite{jucker2022tallo}, but such approaches are frequently inaccurate, and therefore machine learning-based approaches have gained increasing attention.
Prior work on learned tree height estimation~\cite{lang2019country, lang2023high, tolan2024very, pauls2024estimating, pauls2025capturing, wagner2024sub, wagner2025high} has largely focused on predicting dense \glspl{chm} from satellite imagery.
Lang~\etal~\cite{lang2019country, lang2023high} predict national and global \glspl{chm} using \glspl{cnn}, while Tolan~\etal~\cite{tolan2024very} estimate a global \gls{chm} from RGB satellite images using a self-supervised \gls{vit}, specifically DINOv2~\cite{oquab2023dinov2}.
More recently, DINOv3 trained on satellite imagery has achieved superior \gls{chm} estimation, reaching \gls{sota} \cite{simeoni2025dinov3}.
However, the spatial resolution of these large-scale \gls{chm} predictions is insufficient for extracting accurate individual tree heights, inherently introducing absolute height errors~\cite{wagner2025high} (see~\cref{tab:comparative_results_qt,tab:comparative_results_bci,tab:comparative_results_qp}). 
Li~\etal~\cite{li2023deep} addressed this limitation by predicting a national \gls{chm} at \SI{0.4}{m} resolution from RGB-NIR aerial imagery using a U-Net architecture~\cite{ronneberger2015u}, then extracting individual heights via tree crown segmentation.
However, this approach is constrained by costly aerial images at limited resolution where small trees cannot be distinguished.
More closely related to our work, Hao~\etal~\cite{hao2021automated} and Fu~\etal~\cite{fu2024automatic} estimate individual tree heights from \gls{uav} imagery using Mask R-CNN~\cite{he2017mask} by discretizing height values into classification bins. Nevertheless, these studies are limited to single species, specific datasets, and in some cases require additional input modalities (\eg, LiDAR). 
To our knowledge, no existing work directly regresses a single height value from tree-centered RGB images.

\paragraph{Tree species identification.}
General image classification has evolved from \glspl{cnn} \cite{krizhevsky2012imagenet, he2016deep, liu2022convnet} to \glspl{vit}~\cite{dosovitskiy2020image} and recent Mamba-based methods~\cite{gu2024mamba, liu2024vmamba, hatamizadeh2025mambavision}.
For fine-grained classification, linear probing of \glspl{vfm} yields strong results using weakly-supervised~\cite{tschannen2025siglip, bolya2025perception} or self-supervised~\cite{oquab2023dinov2, simeoni2025dinov3} \glspl{vit}.
Similarly, recent plant identification methods~\cite{stevens2024bioclip, lefort2024cooperative, gu2025bioclip} rely only on natural images.
In remote sensing, foundation models such as AnySat~\cite{astruc2025anysat} and Galileo~\cite{tseng2025galileo} have demonstrated effectiveness across various downstream tasks, including tree species classification, but achieve limited accuracy on fine-grained classification due to the coarse resolution of the input data.
Early work on tree species identification from RGB UAV imagery~\cite{natesan2019resnet, sun2019deep, egli2020cnn, natesan2020individual, cloutier2024influence} primarily applied \glspl{cnn} to study-specific datasets, often without evaluating cross-dataset generalization. 
Pierdicca~\etal~\cite{pierdicca2023uav4tree} demonstrated the superior accuracy of \glspl{vit} compared to \gls{cnn}-based methods within a specific study area. 
Zhang~\etal~\cite{zhang2025rsvmamba} recently proposed a Mamba-based architecture for pixel-wise class prediction, whereas our work assumes pre-detected trees and focuses on image-level classification.
Another research direction involves fusing information from multiple modalities~\cite{nezami2020tree, qin2022individual, li2022ace, ecke2024towards, li2025tree}, such as LiDAR~\cite{qin2022individual}.
However, multi-sensor data acquisition is considerably more expensive than RGB-only pipelines.
Finally, extensive research combines tree species classification with object detection~\cite{beloiu2023individual}, segmentation~\cite{schiefer2020mapping, ramesh2024tree}, or both~\cite{teng2025bringing, que2026fm}.
To the best of our knowledge, there is no former work applying a \gls{vfm}-based method for single-tree species prediction from RGB \gls{uav} images.

\paragraph{Multi-task learning for remote sensing.}
\gls{mtl} in remote sensing has primarily focused on dense prediction tasks, such as \gls{dsm} prediction and semantic segmentation~\cite{srivastava2017joint, carvalho2019multitask, gao2023joint, shen2025learning}. 
More recently, \gls{mtl} has been applied to supervised pre-training of remote sensing foundation models, with both global and local tasks~\cite{bastani2023satlaspretrain, wang2024mtp}. 
While these methods employ a shared feature extractor with task-specific heads, similar to our approach, they rely on multi-modal satellite imagery combining RGB with additional spectral channels.
Recent forest monitoring works have combined detection and segmentation with classification~\cite{hao2021automated, fu2024automatic, teng2025bringing}.
Additionally, Xu~\etal~\cite{xu2022novel} used an \gls{mlp} to estimate total biomass and its individual components simultaneously from tabular \gls{dbh} values.
To our knowledge, we are the first to formulate single-tree height estimation and species classification, both essential for biomass estimation, as a \gls{mtl} problem from RGB imagery.

\section{{\benchmark}}
\label{sec:benchmark}
\begin{wrapfigure}[11]{R}{0.5\textwidth}
    \centering
    \vspace{-53pt}
    \begin{tikzpicture}
        
        \node[inner sep=0pt] (img2) at (0,0) {\includegraphics[width=0.32\linewidth]{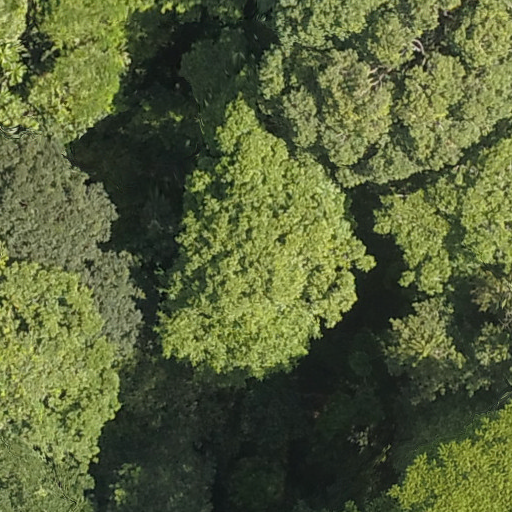}};
        
        \node[inner sep=0pt, anchor=east] (img1) at ([xshift=-1pt]img2.west) {\includegraphics[width=0.32\linewidth]{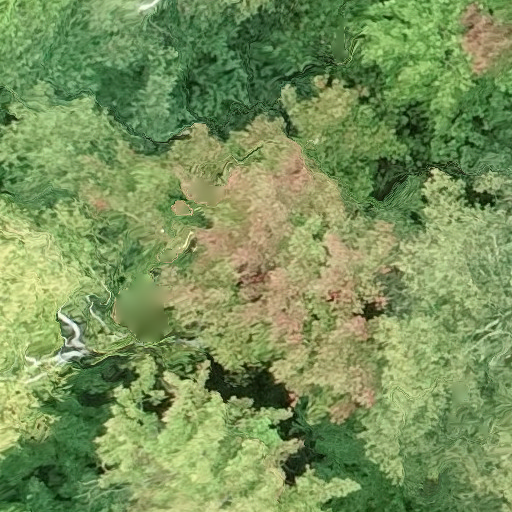}};
        
        \node[inner sep=0pt, anchor=west] (img3) at ([xshift=1pt]img2.east) {\includegraphics[width=0.32\linewidth]{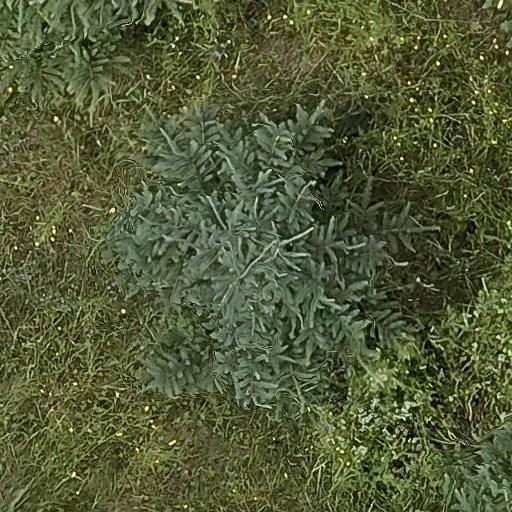}};

        \node[align=center, anchor=south, inner sep=0pt, yshift=5pt, font=\small] at (img1.north) {Temperate\\Forest};
        \node[align=center, anchor=south, inner sep=0pt, yshift=5pt, font=\small] at (img2.north) {Tropical\\Forest};
        \node[align=center, anchor=south, inner sep=0pt, yshift=5pt, font=\small] at (img3.north) {Boreal\\Plantation};

        \node[align=center, anchor=north, inner sep=0pt, yshift=-5pt, font=\scriptsize] at (img1.south) {Height: \SI{19.24}{m}\\Class:\\Acer rubrum};
        \node[align=center, anchor=north, inner sep=0pt, yshift=-5pt, font=\scriptsize] at (img2.south) {Height: \SI{26.04}{m}\\Class:\\Rubiaceae};
        \node[align=center, anchor=north, inner sep=0pt, yshift=-5pt, font=\scriptsize] at (img3.south) {Height: \SI{2.31}{m}\\Class:\\Picea glauca};

    \end{tikzpicture}
    \caption{
    \textbf{Examples from the \benchmark benchmark.}
    It consists of tree-centered RGB \gls{uav} images from three forest types with height and class label.
    \label{fig:benchmark}}
\end{wrapfigure}
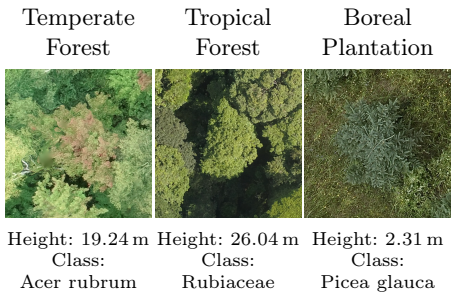
We introduce \benchmark, the first benchmark for individual tree species identification and height estimation from tree-centered RGB images. 
To construct \benchmark, we developed a data extraction pipeline (\cref{subsec:extraction}) and applied it to three diverse datasets: Quebec Trees~\cite{cloutier2024influence} (temperate forests), Barro Colorado Island (BCI)~\cite{Vasquez2023BCI} (tropical forests), and Quebec Plantations~\cite{Lefebvre:2024} (boreal plantations). 
We detail datasets in \cref{subsec:datasets} and examples of the benchmark in \cref{fig:benchmark}.

\subsection{Data Extraction Pipeline \label{subsec:extraction}}
\begin{wrapfigure}[14]{r}{0.296\textwidth}
    \centering
    \vspace{-48pt}
    \begin{tikzpicture}
    \node[anchor=south west, inner sep=0] (image) at (0,0) {\includegraphics[width=\linewidth]{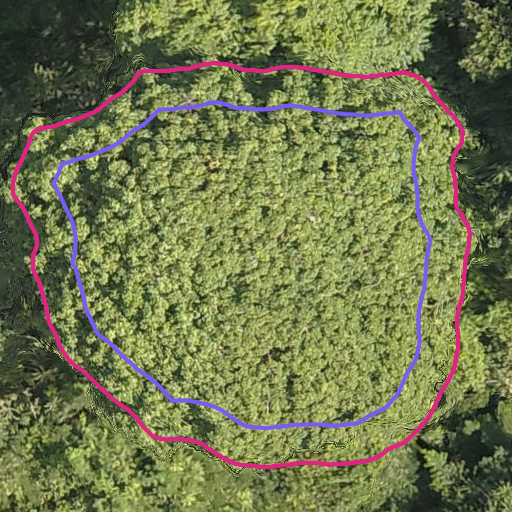}};
    \begin{scope}[x={(image.south east)}, y={(image.north west)}]
        \node[fill=gray!30, fill opacity=0.7, text opacity=1, inner sep=1pt] at (0.87, 0.53) {\textcolor{ourpink}{$\mathbf{S}$}};
        \node[fill=gray!30, fill opacity=0.7, text opacity=1, inner sep=1pt] at (0.3, 0.5) {\textcolor{ourpurple}{$\mathbf{S_{\textbf{buf}}}$}};
        \draw[<->, very thick, black] (0.5, 0.17) -- (0.5, 0.09);
        \node[fill=gray!30, fill opacity=0.7, text opacity=1, inner sep=1pt] at (0.65, 0.13) {\textcolor{black}{$\mathbf{0.1L}$}};
    \end{scope}
    \end{tikzpicture}
    \caption{
    \textbf{Segmentation boundary buffering.}
    The pixel bounded by the \textcolor{ourpink}{pink} and \textcolor{ourpurple}{purple} contours correspond to the original segmentation $S$ and the buffered one $S_\text{buf}$.
    \label{fig:buffer}}
\end{wrapfigure}
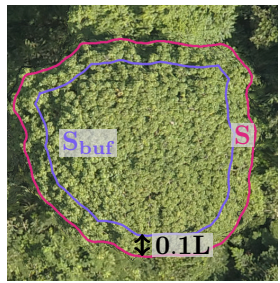
The source data comprises orthomosaics of multiple trees, each with individual segmentations and class labels.
To construct our benchmark, we use manually annotated segmentations of individual tree crowns as a preprocessing step.
While manual segmentations would not be available in field deployments, tree crown segmentation models achieving human-level accuracy have already been designed~\cite{duguay2026selvamask}, so our benchmark represents a realistic use case.
For each tree instance, we extract a 512$\times$512 tile centered at its centroid.

While species labels are available for all tree annotations, field measurements of height exist only for a subset of the Quebec Plantation dataset.
Consequently, we extract ground truth heights from LiDAR-derived \glspl{chm} collected concurrently with the imagery. 
LiDAR measurements serve as the standard ground truth for canopy height~\cite{lang2019country, wagner2024sub}.
Since individual tree segmentations lack pixel-level precision, height values near segmentation boundaries may correspond to neighboring trees.
To mitigate this issue, we create a buffered segmentation $S_{buf}$ from the original segmentation $S$ by excluding boundary pixels (see \cref{fig:buffer}). 
Let $\mathbf{x}, \mathbf{y} \in \{0, ..., W-1\} \times \{0, ..., H-1\}$ denote pixel coordinates within the \gls{chm} raster of dimensions $W \times H$. 
The buffered pixel set $S_{buf}$ is defined as:
\begin{equation}
    S_{\text{buf}}= \{ \mathbf{x} \in S \mid \min_{\mathbf{y} \notin S} \| \mathbf{x} - \mathbf{y} \|_2 > 0.1L \}. \label{eq:buffer}
\end{equation}
The characteristic length $L$ is defined as the square root of the segmentation area $|S|$, measured in pixels: $L = \sqrt{|S|}$.
To ensure robustness against outliers in the \gls{chm} raster $C_{x,y}$ with coordinates $x \in \{0, ..., W-1\}$ and $y \in \{0, ..., H-1\}$, we compute the tree height $h$ as the 99th percentile ($P_{99}$) of values within the buffered region:
\begin{equation}
    h = P_{99} ( \{ C_{x,y} \mid (x,y) \in S_{buf} \} ).
\end{equation}
We describe pipeline modifications for Quebec Plantations in App.~\ref{app:implementation_details}.

\subsection{Datasets \label{subsec:datasets}}
We use three datasets spanning two countries (Canada and Panama) and two forest types (mature forests and plantations). 
To mitigate spatial autocorrelation, we adopt the spatial splits defined by Teng~\etal~\cite{teng2025bringing}.
We summarize the properties of each dataset below, deferring a comprehensive analysis to App.~\ref{app:datasets}.

\begin{figure}[tb]
    \centering
    \begin{subfigure}[b]{0.325\linewidth}
        \centering
        \includegraphics[width=\linewidth]{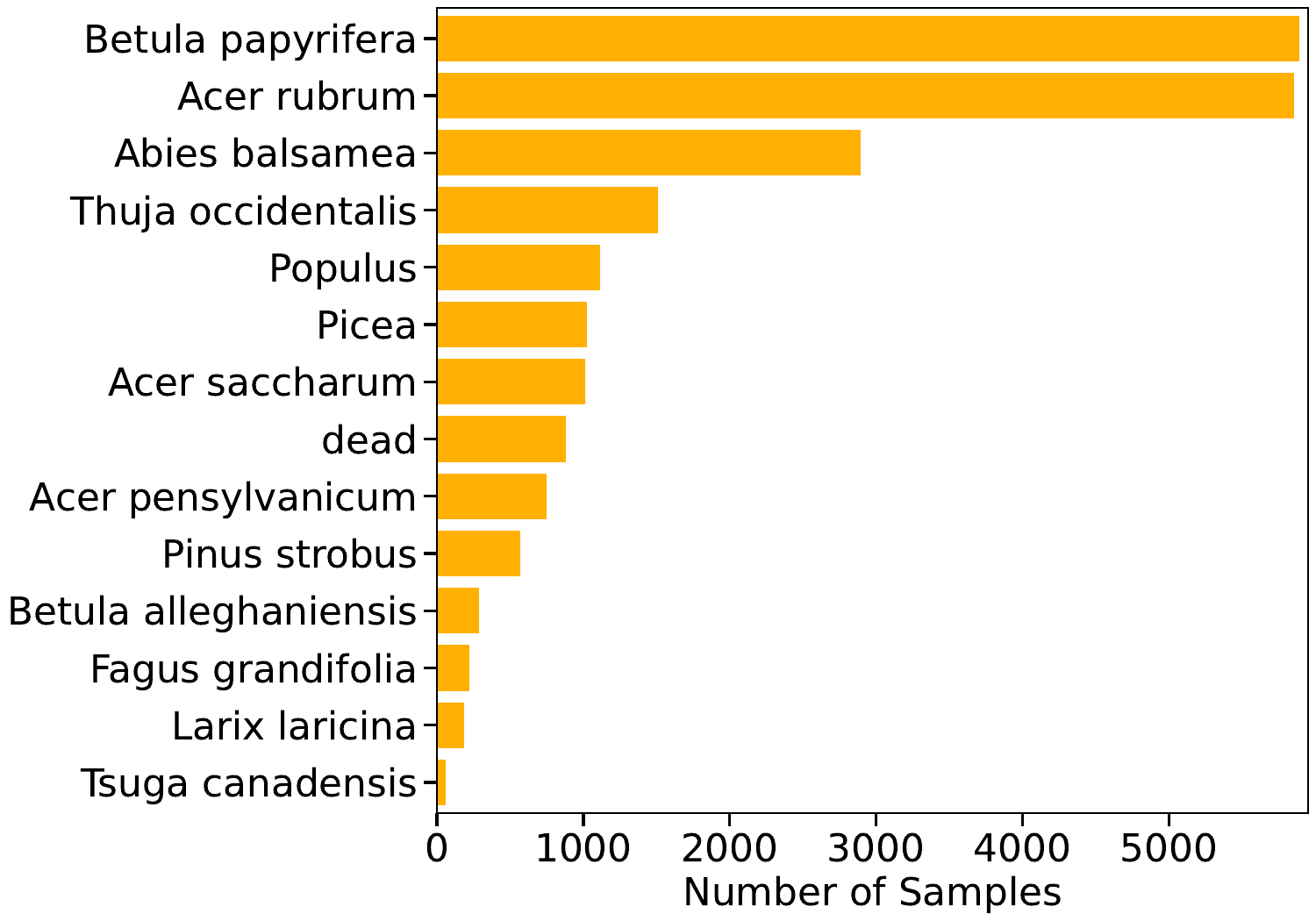}
        \caption{Quebec Trees}
        \label{fig:class_dis_qt}
    \end{subfigure}
    \hfill
    \begin{subfigure}[b]{0.325\linewidth}
        \centering
        \includegraphics[width=\linewidth]{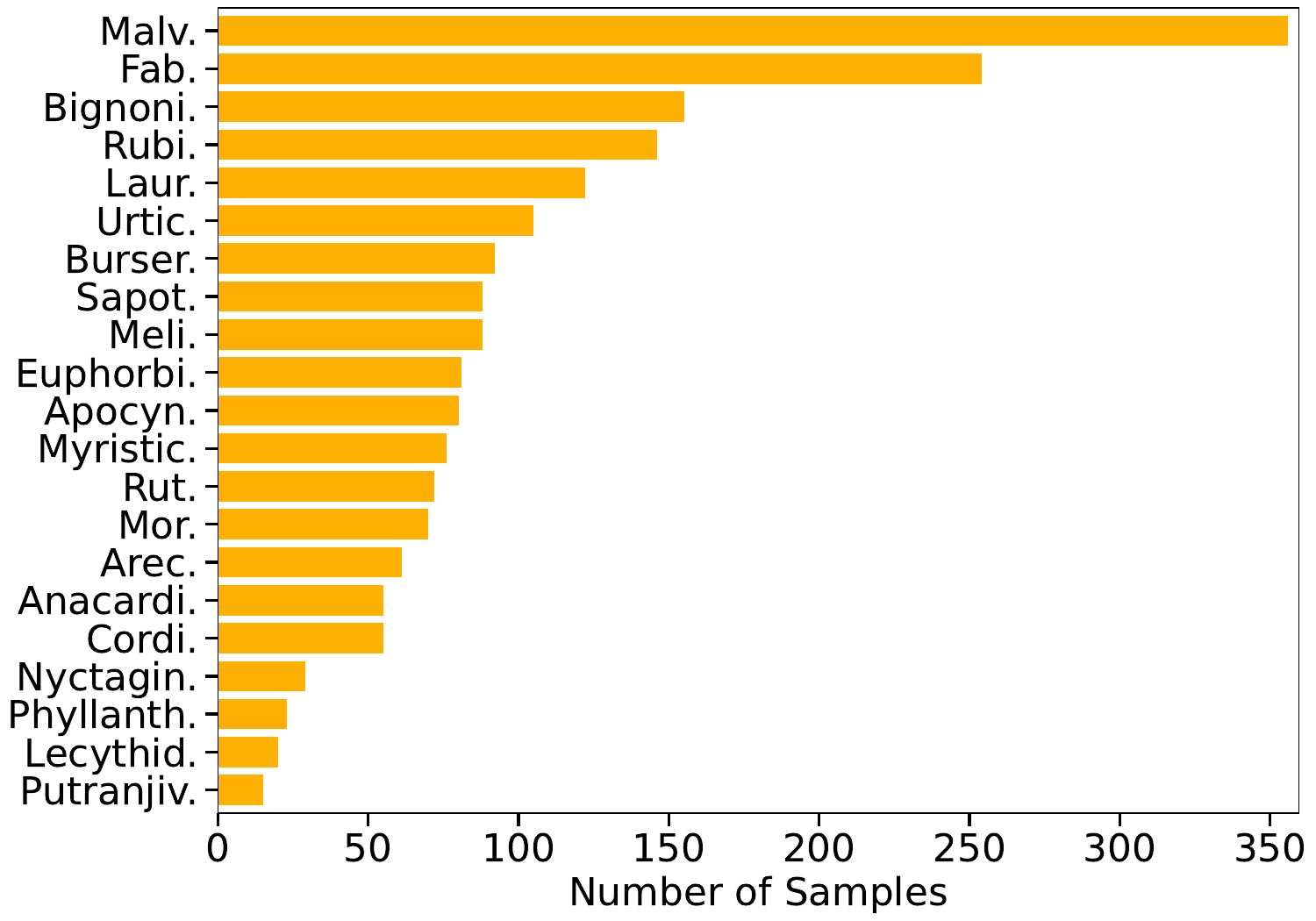}
        \caption{BCI}
        \label{fig:class_dis_bci}
    \end{subfigure}
    \hfill
    \begin{subfigure}[b]{0.325\linewidth}
        \centering
        \includegraphics[width=\linewidth]{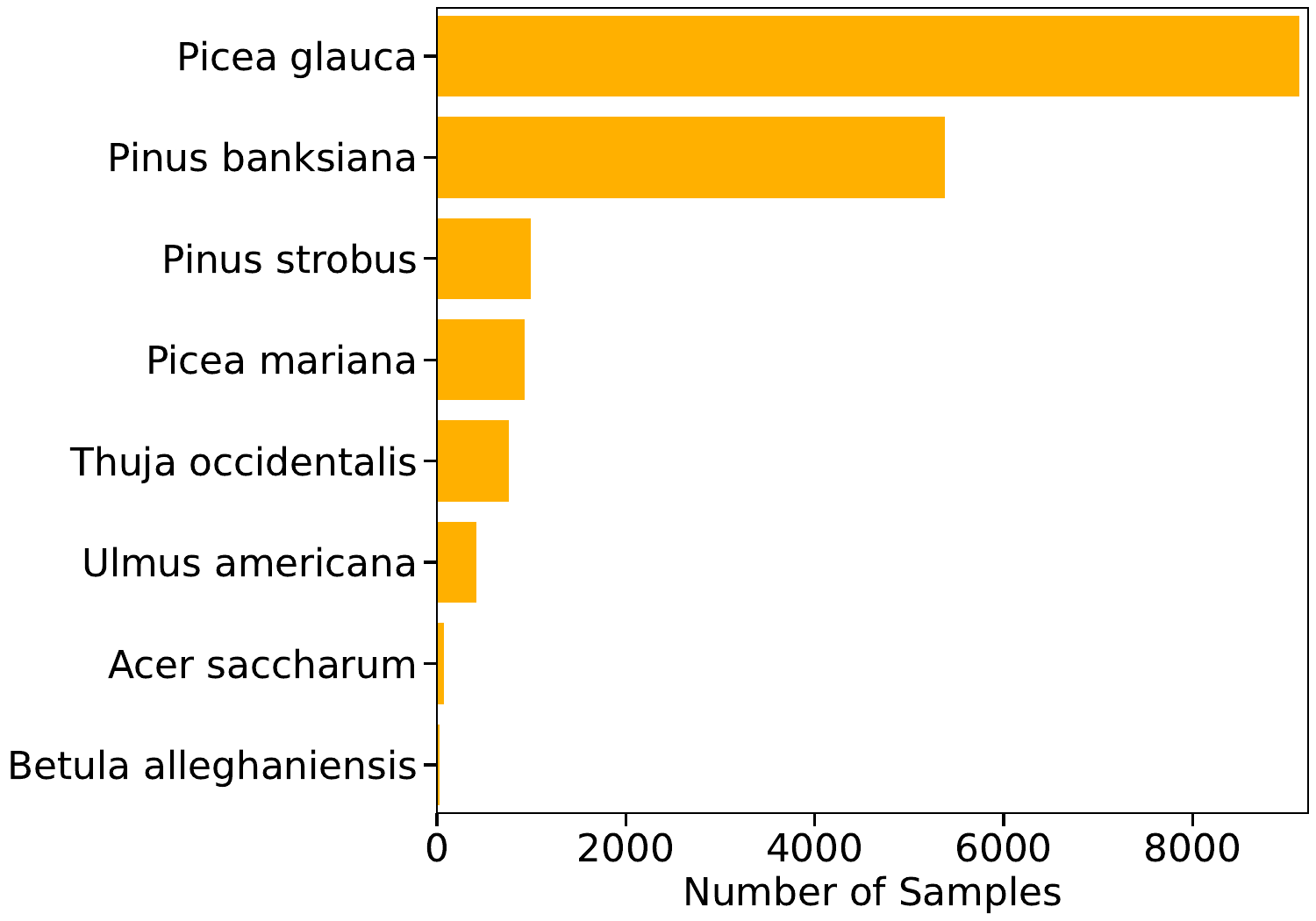}
        \caption{Quebec Plantations}
        \label{fig:class_dis_qp}
    \end{subfigure}
    \caption{\textbf{Class histograms} of the Quebec Trees, BCI and Quebec Plantation datasets. Distributions per split are in App.~\ref{app:class_distribution}.
    }
    \label{fig:class_dis}
\end{figure}

\begin{figure}[tb]
    \centering
    \begin{subfigure}[b]{0.325\linewidth}
        \centering
        \includegraphics[width=\linewidth]{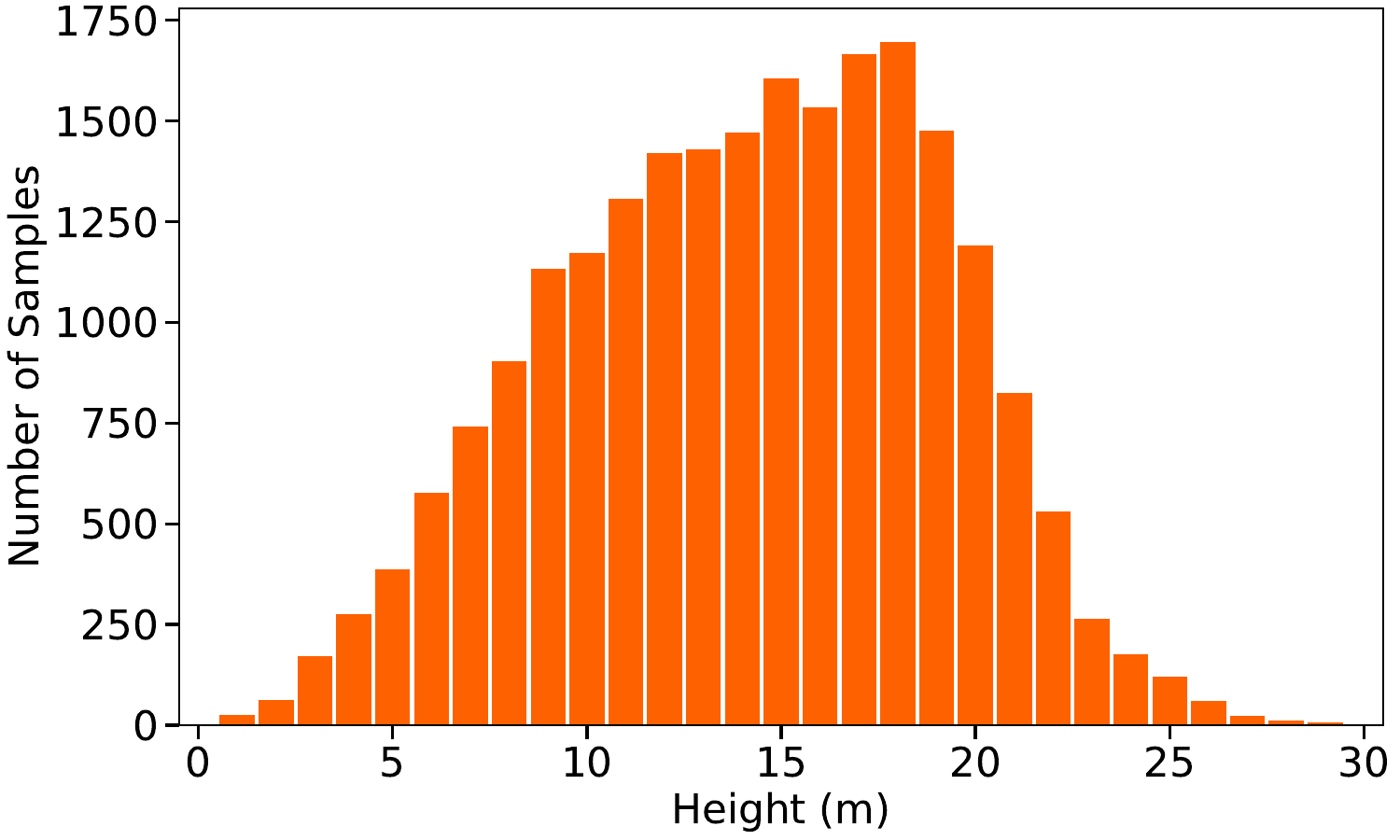}
        \caption{Quebec Trees}
        \label{fig:height_dis_qt}
    \end{subfigure}
    \hfill
    \begin{subfigure}[b]{0.325\linewidth}
        \centering
        \includegraphics[width=\linewidth]{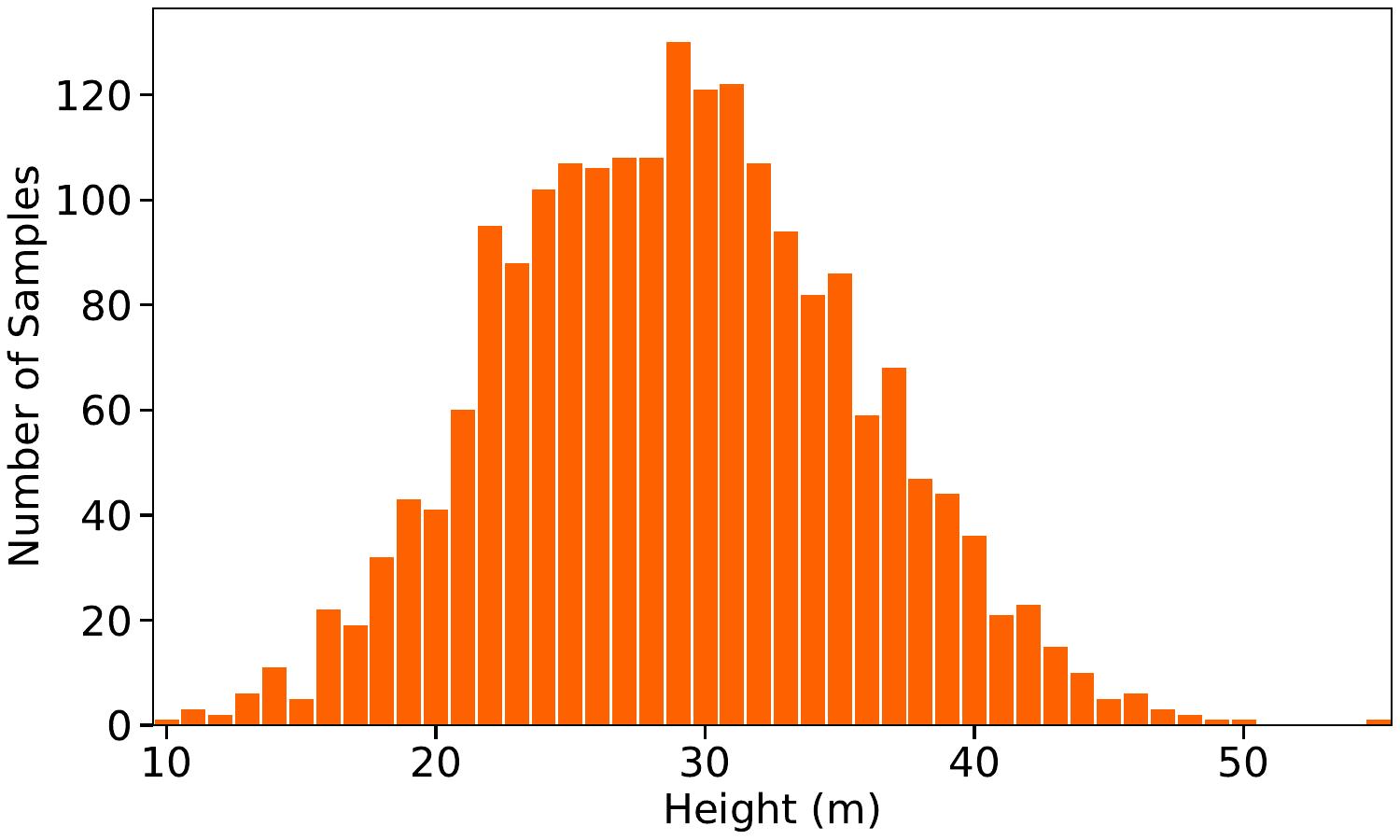}
        \caption{BCI}
        \label{fig:height_dis_bci}
    \end{subfigure}
    \hfill
    \begin{subfigure}[b]{0.325\linewidth}
        \centering
        \includegraphics[width=\linewidth]{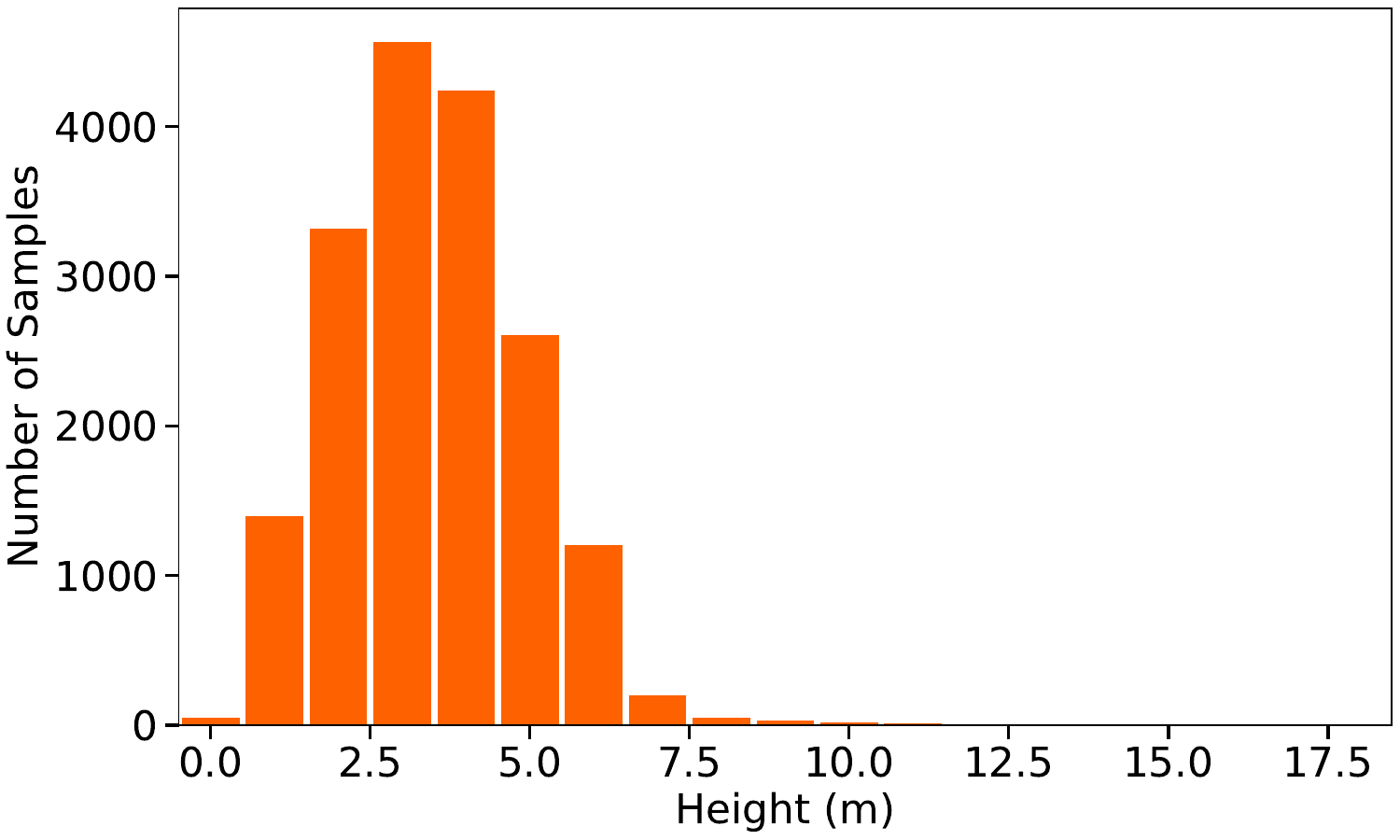}
        \caption{Quebec Plantations}
        \label{fig:height_dis_qp}
    \end{subfigure}
    \caption{\textbf{Height histograms} of the Quebec Trees, BCI and Quebec Plantation datasets with \SI{1}{m} intervals. Distributions per split are in App.~\ref{app:height_stats}.
    }
    \label{fig:height_dis}
\end{figure}

\paragraph{Quebec Trees dataset.}
The Quebec Trees dataset\cite{cloutier2024influence} contains 22.3K images (train: 13.3K, val: 3.6K, test: 5.4K) of temperate forest trees in Quebec, Canada, with a spatial resolution of \SI{1.9}{cm/pixel}.
We adopt the class definitions from Teng~\etal~\cite{teng2025bringing}, excluding supercategories (`Acer', `Betula', `Magnoliopsida', `Pinopsida'), which reflect annotator uncertainty.
After filtering these classes, the dataset comprises 14 classes\footnote{The classes `Populus' and `Picea' are genus-level and not species-level due to annotator uncertainty.} with a long-tailed distribution (\cref{fig:class_dis_qt}).
Quebec Trees have mean height \SI{14.22}{m} and standard deviation \SI{4.91}{m} (\cref{fig:height_dis_qt}).

\paragraph{BCI dataset.}
The BCI dataset~\cite{Vasquez2023BCI, forestgeo_smithsonian_2024}, captured in Panama, represents a complex tropical forest environment.
The dataset comprises 2.0K images (train: 1.4K, val: 0.3K, test: 0.3K) with a spatial resolution of \SI{4.0}{cm/pixel}.
Unlike Teng~\etal~\cite{teng2025bringing}, we exclude the `Other' class and any class with fewer than two samples in the validation or test sets to ensure reliable evaluation.
The resulting 21 classes\footnote{Classes correspond to families and not species on the BCI dataset.} exhibit a long-tailed distribution (\cref{fig:class_dis_bci}).
BCI has the widest height range, with the highest mean (\SI{29.05}{m}) and standard deviation (\SI{6.59}{m}) (\cref{fig:height_dis_bci}).

\paragraph{Quebec Plantations dataset.}
The Quebec Plantations dataset~\cite{Lefebvre:2024} comprises 17.7K images (train: 11.1K, val: 4.0K, test: 2.6K) from boreal plantations in Quebec, Canada, at \SI{0.5}{cm/pixel} spatial resolution, making it comparable in size to Quebec Trees. 
Plantations contain young trees in regular grids, so most images exclude neighboring trees.
We exclude the `Other' class aggregating diverse species without consistent visual characteristics. 
The eight remaining classes exhibit an imbalanced distribution (\cref{fig:class_dis_qp}), dominated by \textit{Picea glauca} and \textit{Pinus banksiana}.
Quebec Plantations has the lowest tree heights, with mean \SI{3.48}{m} and standard deviation \SI{1.47}{m} (\cref{fig:height_dis_qp}).
Minimum heights approach \SI{0}{m} due to the presence of young trees, the linear correction described in App.~\ref{app:implementation_details}, and small errors in the \gls{dtm} used to create the \gls{chm}~\cite{lefebvre2025suivi}.

\subsection{Evaluation Metrics}
\label{subsec:metrics}
We report macro-averaged \gls{f1} and \gls{acc} for species identification, and \gls{mae}, \gls{rmse}, \gls{msle}, and threshold accuracy $\delta_{1.25}$ for height estimation.
Standard metrics not detailed below are defined in App.~\ref{app:eval}.
The \gls{msle} accommodates the wide range of height values by computing squared errors in log-space.
For $N$ images, where $\hat{h}_i$ and $h_i$ represent the predicted and ground truth heights of the $i$-th image respectively, it is defined as:
\begin{equation}
    \text{MSLE} = \frac{1}{N} \sum_{i=1}^{N} \left( \ln(\hat{h}_i + 1) - \ln(h_i + 1) \right)^2.
\end{equation}
Following monocular depth estimation protocols~\cite{eigen2014depth}, we define threshold accuracy $\delta_{1.25}$ as the ratio of predictions within a relative error threshold:
\begin{equation}
    \delta_{1.25} = \frac{1}{N}\sum_{i=1}^{N} \mathbbm{1} \left( \max \left( \frac{\hat{h}_i}{h_i}, \frac{h_i}{\hat{h}_i} \right) < 1.25 \right),
\end{equation}
where $\mathbbm{1}(\cdot)$ is the indicator function.
A prediction is considered accurate if it lies within a multiplicative factor of $1.25$ from the ground truth. 
The key advantages of $\delta_{1.25}$ are robustness to outliers and tolerance for small deviations, which accounts for inherent noise in height measurements. 
Moreover, since $\delta_{1.25}$ shares the $[0,1]$ range with classification metrics, it enables averaging F1-Score and $\delta_{1.25}$ for checkpoint selection in \gls{mtl}.
\section{{\method} }
\label{sec:method}
\begin{figure*}[t]
\centering
\includegraphics[width=\textwidth]{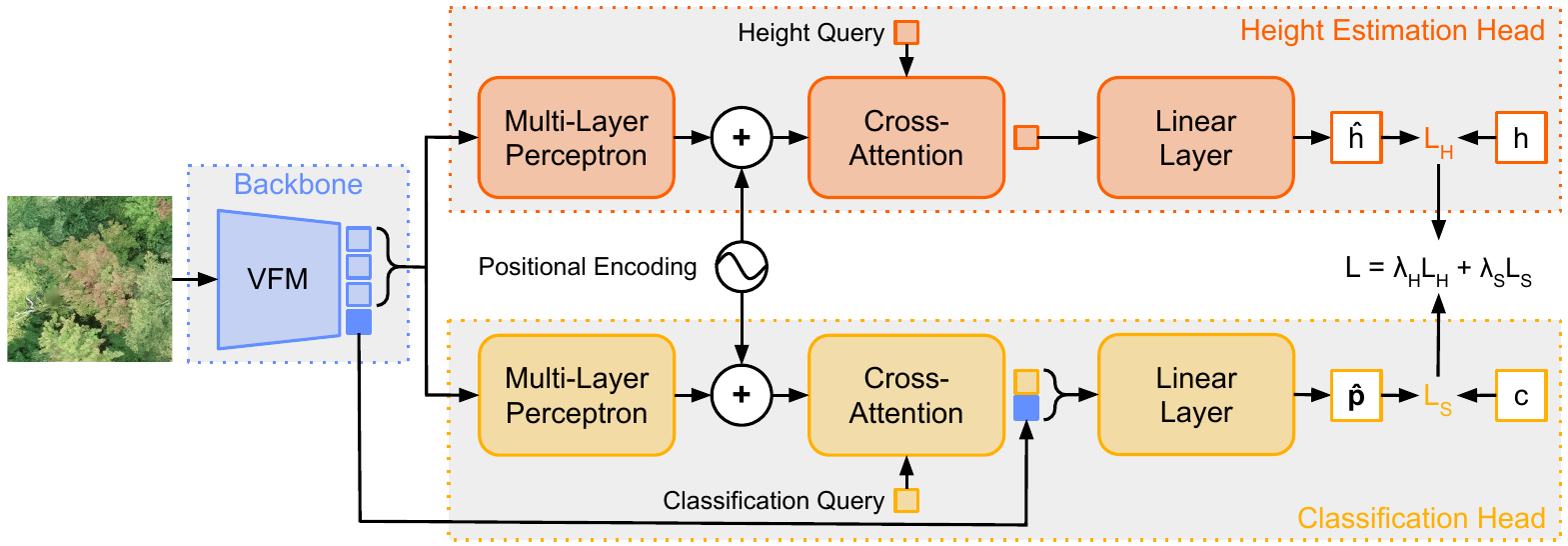}
\caption{
\textbf{Overview of {\method}.}
A shared \gls{vfm} (\textcolor{ourblue}{blue}) extracts features from an RGB image.
In the height estimation head (\textcolor{ourorange}{orange}), a learnable height query cross-attends to adapted patch tokens to predict $\hat{h}$.
In the classification head (\textcolor{ouryellow}{yellow}), a learnable classification query cross-attends to adapted patch tokens to obtain a classification token.
We then concatenate it with the \gls{vfm} [CLS] token to predict $\hat{\mathbf{p}}$.
The total loss $\mathcal{L}$ is the weighted sum of the height estimation loss $\mathcal{L}_H(h, \hat{h})$ and the classification loss $\mathcal{L}_S(\hat{\mathbf{p}}, c)$ supervised by ground truth height $h$ and class $c$.
\label{fig:architecture}}
\end{figure*}

We present \method, a multi-task model for tree height estimation and classification.
Given a single RGB image $I \in \mathbb{R}^{H \times W \times 3}$, our goal is to predict the height $h \in \mathbb{R}_{\geq0}$ and class $c \in \{1, \dots, C\}$ of the tree in the center of $I$ using a single model.
$C$ denotes the number of possible classes.
While \method is designed for cloud rather than on-device deployment, computational efficiency remains important due to the large number of tree tiles extracted from forest-scale surveys.
In \cref{subsec:architecture} we propose our multi-task architecture for tree height estimation and classification.
We develop suitable losses for both tasks as well as their combination in \cref{subsec:loss}.
\cref{fig:architecture} outlines our proposed method.

\subsection{Architecture \label{subsec:architecture}}
Our proposed framework comprises three main components: a shared \gls{vfm} backbone for feature extraction and two task-specific heads.

\paragraph{\gls{vfm} feature extraction.}
Given the scarcity of high-resolution forest drone imagery, we leverage a pre-trained \gls{vfm} to initialize our feature extractor.
Specifically, we use DINOv3~\cite{simeoni2025dinov3}, a \gls{vit}~\cite{dosovitskiy2020image} trained via self-supervised learning on over one billion web images, which achieves strong performance on species identification and \gls{chm} estimation even with a frozen backbone. 
DINOv3 provides robust initialization due to semantic correspondence between general object and tree species classification, and geometric alignment between \gls{chm} and individual tree height estimation (\cref{subsec:analysis_method}).
However, to address the distribution shift between web images and forest drone imagery, we fine-tune the \gls{vfm}. 
The resulting patch tokens and class token ([CLS]) are then fed to the task-specific heads.

\paragraph{Task-specific heads.}
Both tasks share a single backbone instance during training and inference for computational efficiency.
We keep the heads separate, as partial sharing degrades both tasks disproportionately to the minor parameter savings (App.~\ref{app:parameter_sharing}).
For \textbf{height estimation}, we aggregate spatial information using cross-attention.
First, an \gls{mlp} with two linear layers and GELU activation~\cite{hendrycks2016gaussian} projects patch tokens into a task-specific feature space.
A learnable height query token then cross-attends to these projected tokens to extract height-relevant features. 
Following Carion~\etal~\cite{carion2020end}, we add 2D sine positional encodings to the key tokens for spatially-aware attention.
Finally, a linear layer maps the extracted height token to a scalar, passed through ReLU~\cite{nair2010rectified} to ensure non-negative predictions ($\hat{h} \in \mathbb{R}_{\geq0}$).
The \textbf{classification} head uses a similar architecture but incorporates global context.
A separate learnable classification query aggregates local semantic features by cross-attending to adapted patch tokens.
To integrate local and global information efficiently, we concatenate this output with the backbone's [CLS] token and project to $C$ class logits $\hat{\mathbf{z}} \in \mathbb{R}^C$ via a linear layer.
Class probabilities $\hat{\mathbf{p}} \in [0,1]^C$ are obtained through the softmax function.
We detail the \method architecture further in App.~\ref{app:implementation_details}.

\subsection{Loss \label{subsec:loss}}
This section describes the loss functions for tree height regression and classification, followed by our loss weighting strategy.

\paragraph{Task-specific losses.}
We use smooth L1 loss~\cite{girshick2015fast} for \textbf{height regression}. 
Compared to L2 loss, it is less sensitive to outliers that may exist in our height ground truth (\cref{fig:height_dis}).
Additionally, small errors, potentially arising from measurement noise, incur smaller penalties than with standard L1 loss. 
Given predicted height $\hat{h}$ and ground truth height $h$, the loss is defined as:
\begin{equation}
    \mathcal{L}_H(\hat{h}, h) = 
    \begin{cases} 
        \frac{1}{2} \left( \hat{h} - h \right)^2, & \text{if } |\hat{h} - h| < 1 \\
        |\hat{h} - h| - \frac{1}{2}, & \text{otherwise}
    \end{cases}.
\end{equation}
We employ standard cross-entropy loss for \textbf{classification}. 
As detailed in App.~\ref{app:loss}, we evaluated specialized long-tail losses~\cite{cui2019class, lin2017focal} but observed no substantial improvement.
For predicted class probabilities $\hat{\mathbf{p}} \in [0,1]^C$ and ground truth class index $c$, the loss is defined as:
\begin{equation}
    \mathcal{L}_S(\hat{\mathbf{p}}, c) = - \sum_{i=1}^{C} \mathbbm{1}_{[i = c]} \log(\hat{p}_i).
\end{equation}

\paragraph{Loss weighting.}
Balancing task-specific objectives is a key challenge in \gls{mtl}. 
Manual tuning is inefficient and often yields suboptimal results. 
Kendall~\etal~\cite{kendall2018multi} proposed uncertainty-based weighting, while Yu~\etal~\cite{yu2020gradient} introduced PCGrad for gradient conflict resolution. 
Based on empirical comparisons (\cref{subsec:analysis_method}), we adopt \gls{dwa}~\cite{liu2019end}, which dynamically adjusts weights based on relative loss changes. 
The total loss at epoch $t$ is defined as:
\begin{equation}
    \mathcal{L}(t) = \lambda_H(t) \mathcal{L}_H(t) + \lambda_S(t) \mathcal{L}_S(t).
\end{equation}
The dynamic weights $\lambda_k(t)$ for task $k \in \{H, S \}$ are computed as:
\begin{equation}
    \lambda_k(t) = \frac{2 \exp(w_k(t-1) / T)}{\sum_{j \in \{H, S\}} \exp(w_j(t-1) / T)}, \quad \text{with } w_k(t-1) = \frac{\mathcal{L}_k(t-1)}{\mathcal{L}_k(t-2)}.
\end{equation}
We set the temperature $T=2$ and initialize weights as $\lambda_H=\lambda_S=1$ for $t<3$.

\section{Experiments}
In this section, we evaluate \method against competing methods on \benchmark.
We then analyze our design choices and the resulting height predictions.

\subsection{Baseline and Competing Methods}
\label{subsec:baselines}

\paragraph{Allometric equations as a baseline.}
Practitioners estimate tree height from easier measurable crown radius using allometric equations \cite{jucker2022tallo}. 
Specifically, these equations approximate tree height $h$ as a function of crown radius $r$ using parameters $a_c$ and $b_c$ for a given species $c$, via $\ln \hat{h} \approx a_c \ln r + b_c$.
This baseline assumes an oracle access to both ground truth species and segmentation masks.
We compute the length $l$ and width $w$ of the minimum rotated bounding rectangle enclosing the ground truth mask and derive the crown radius as $r \approx (l + w)/4$.
Alternatively, we tested deriving radius from mask area (assuming circular crowns), but this yielded worse height estimates.
We fit the allometric equation parameters\footnote{For non-species classes, we combine all species within the genus or family.} using the Tallo dataset~\cite{jucker2022tallo}.

\paragraph{Mask R-CNN.}
Prior work has adopted Mask R-CNN~\cite{he2017mask} for height estimation by discretizing continuous height values into classification bins~\cite{hao2021automated, fu2024automatic}.
Empirically, we found an interval size of \SI{0.2}{m} to yield the best results.
We detail the adaptation of this method to \benchmark in App.~\ref{app:implementation_details}.

\paragraph{\glspl{cnn}, \glspl{vit} and Mamba-based backbones.}
ResNet-50~\cite{he2016deep}, using residual connections, is widely adopted for tree species identification~\cite{natesan2019resnet, sun2019deep}.
We extend ResNet-50 by adding a photogrammetric \gls{dsm} as a fourth input channel, as \glspl{dsm} have proven beneficial in forest scene understanding~\cite{teng2025bringing}.
We also evaluate ConvNeXt-B~\cite{liu2022convnet}, a modern \gls{cnn} that incorporates \gls{vit} design principles into the ResNet architecture, and SwinV2-B~\cite{liu2022swin}, a hierarchical \gls{vit} that uses shifted windows to capture local and global dependencies.
Furthermore, we evaluate MambaVision~\cite{hatamizadeh2025mambavision}, a recent hybrid Mamba Transformer~\cite{vaswani2017attention} backbone, achieving \gls{sota} on ImageNet-1K~\cite{deng2009imagenet}.

\paragraph{Foundation models.}
We evaluate AnySat\cite{astruc2025anysat}, a remote sensing foundation model that includes aerial forest imagery in its pre-training, achieving \gls{sota} on tree species identification. 
We also consider PECore~\cite{bolya2025perception}, a weakly-supervised \gls{vfm} with strong generalization across tasks, especially on zero-shot fine-grained classification, which is closely related to species identification. 
Finally, we benchmark DINOv3~\cite{simeoni2025dinov3}, a state-of-the-art self-supervised \gls{vfm}. We fine-tune DINOv3-B and DINOv3-L and evaluate Large version with frozen backbone, as well as ViT-L Sat, which was pre-trained on satellite rather than web images.

\paragraph{Implementation details of competing methods.}
We adapt pre-trained models by replacing only the final projection layer. 
For classification, we use a linear layer mapping to $C$ class logits.
For height estimation, we use a single-output scalar from a linear layer with ReLU activation to ensure $\hat{h} \geq 0$.
For all competing methods, we use identical training hyperparameters to \method (App.~\ref{app:implementation_details}) but train separate models for height estimation and classification.

\begin{table}[tb]
    \caption{
        \textbf{Comparative results on the Quebec Trees test split.}
        We compare \method against competing methods in terms of parameter count, height estimation, and classification.
        We group models by parameter count: fewer than 300M (top) and more than 300M (bottom).
        For competing methods, `Param.' denotes combined parameters of two independently-trained task-specific models.
        We report the mean ± standard error over 5 seeds.
        \textbf{Bold} and \underline{underline} denote best and second-best.
        $\ast$ indicates an oracle with ground truth class and segmentation;
        $\dagger$ denotes the Mask R-CNN variant from~\cite{hao2021automated, fu2024automatic} (\cref{subsec:baselines});
        {\color{iceblue}\faSnowflake} indicates a frozen backbone.
    }
    \label{tab:comparative_results_qt}
    \centering
    \setlength{\tabcolsep}{3pt}
    \sisetup{table-number-alignment=center, separate-uncertainty=true, retain-zero-uncertainty=true}
    \renewcommand{\arraystretch}{1.2}
    \resizebox{\textwidth}{!}{%
        \begin{tabular}{
        l
        S[table-format=3]
        S[table-format=2.2(2), separate-uncertainty]
        S[table-format=2.2(2), separate-uncertainty]
        S[table-format=1.2(2), separate-uncertainty]
        S[table-format=1.2(2), separate-uncertainty]
        S[table-format=2.2(2), separate-uncertainty]
        S[table-format=2.2(2), separate-uncertainty]}
            \toprule
            & & \multicolumn{4}{c}{\textbf{Height Estimation}} & \multicolumn{2}{c}{\textbf{Classification}} \\
            \cmidrule(lr){3-6} \cmidrule(lr){7-8}
            \multirow{-2}{*}{\vspace{0.5em}\textbf{Method}} & {\multirow{-2}{*}{\textbf{\shortstack[c]{Param.\\(M)~$\downarrow$}}}} & {$\delta_{1.25}$~(\%)~$\uparrow$} & {MSLE~($10^{-2}$)~$\downarrow$} & {MAE~(m)~$\downarrow$} & {RMSE~(m)~$\downarrow$} & {F1~(\%)~$\uparrow$} & {Acc~(\%)~$\uparrow$} \\
            \midrule
            Allometric equations$^\ast$ & \bfseries 0 & 48.74 & 10.97 & 2.98 & 3.73 & {--} & {--} \\
            Mask R-CNN$^\dagger$~\cite{he2017mask} & 88 & 74.34(32) & 4.09(06) & 1.73(02) & 2.26(02) & 65.49(45) & 65.86(60) \\
            ResNet50~\cite{he2016deep} & \ulinttwo{47} & 68.16(13) & 5.45(09) & 1.96(01) & 2.48(01) & 76.67(38) & 74.01(50) \\
            ResNet50~\cite{he2016deep} w/ DSM & \ulinttwo{47} & 68.11(16) & 5.12(05) & 1.96(01) & 2.47(01) & 76.36(36) & 73.58(50) \\
            ConvNext-B~\cite{liu2022convnet} & 175 & 84.82(39) & 2.47(08) & 1.32(02) & 1.71(02) & 84.87(42) & 82.91(40) \\
            SwinV2-B~\cite{liu2022swin} & 174 & 77.39(87) & 3.38(12) & 1.73(04) & 2.17(05) & 85.78(45) & 83.77(56) \\
            MambaVision-B~\cite{hatamizadeh2025mambavision} & 193 & 68.47(70) & 60.19(4.47) & 2.36(06) & 3.42(10) & 85.37(28) & 83.42(40) \\
            AnySat~\cite{astruc2025anysat} & 250 & 78.03(28) & 3.35(05) & 1.57(01) & 2.04(01) & 69.99(61) & 68.98(44) \\
            PECore-B~\cite{bolya2025perception} & 187 & 84.85(1.03) & 2.31(14) & 1.29(04) & 1.68(05) & 86.82(33) & 84.57(53) \\
            DINOv3-B~\cite{simeoni2025dinov3} & 171 & \ulmain{86.35}{0.26} & \ulpadded{2.10}{0.04} & \ulmainsmall{1.25}{0.01} & \ulmainsmall{1.62}{0.01} & \bfmain{87.46}{0.53} & \bfmain{85.10}{0.80} \\
            \rowcolor{gray!25} \method-B (Ours) & 100 & \bfmain{88.29}{0.32} & \bfpadded{1.85}{0.05} & \bfmainsmall{1.17}{0.02} & \bfmainsmall{1.54}{0.02} & \ulmain{86.89}{0.31} & \ulmain{85.03}{0.45} \\
            \midrule
            DINOv3-L-Sat~\cite{simeoni2025dinov3} ({\color{iceblue}\faSnowflake}) & \ulint{606} & 53.07(04) & 10.71(02) & 2.81(00) & 3.40(00) & 27.43(12) & 25.03(11) \\
            DINOv3-L~\cite{simeoni2025dinov3} ({\color{iceblue}\faSnowflake}) & \ulint{606} & 58.28(06) & 8.42(01) & 2.48(00) & 3.06(00) & 28.76(07) & 26.64(06) \\
            DINOv3-L~\cite{simeoni2025dinov3} & \ulint{606} & \ulmain{88.03}{0.38} & \ulpadded{1.86}{0.04} & \ulmainsmall{1.17}{0.01} & \ulmainsmall{1.53}{0.01} & \bfmain{88.85}{0.37} & \ulmain{86.92}{0.57} \\
            \rowcolor{gray!25} \method-L (Ours) & \bfseries 328 & \bfmain{89.28}{0.28} & \bfpadded{1.71}{0.04} & \bfmainsmall{1.12}{0.02} & \bfmainsmall{1.46}{0.02} & \ulmain{88.38}{0.38} & \bfmain{87.25}{0.36} \\
            \bottomrule
        \end{tabular}
    }
\end{table}

\begin{table}[tb]
    \caption{
        \textbf{Comparative results on the BCI test split.}
        We compare \method against competing methods in terms of parameter count, height estimation and classification.
        We group models by parameter count: fewer than 300M (top) and more than 300M (bottom).
        For competing methods, `Param.' denotes combined parameters of two independently-trained task-specific models.
        We report mean ± standard error over 5 seeds.
        \textbf{Bold} and \underline{underline} denote best and second-best.
    }
    \label{tab:comparative_results_bci}
    \centering
    \setlength{\tabcolsep}{3pt}
    \sisetup{table-number-alignment=center, separate-uncertainty=true, retain-zero-uncertainty=true}
    \renewcommand{\arraystretch}{1.2}
    \resizebox{0.99\textwidth}{!}{%
        \begin{tabular}{
        l
        S[table-format=3]
        S[table-format=2.2(2), separate-uncertainty]
        S[table-format=1.2(2), separate-uncertainty]
        S[table-format=1.2(2), separate-uncertainty]
        S[table-format=1.2(2), separate-uncertainty]
        S[table-format=2.2(2), separate-uncertainty]
        S[table-format=2.2(2), separate-uncertainty]}
            \toprule
            & & \multicolumn{4}{c}{\textbf{Height Estimation}} & \multicolumn{2}{c}{\textbf{Classification}} \\
            \cmidrule(lr){3-6} \cmidrule(lr){7-8}
            \multirow{-2}{*}{\vspace{0.5em}\textbf{Method}} & {\multirow{-2}{*}{\textbf{\shortstack[c]{Param.\\(M)~$\downarrow$}}}} & {$\delta_{1.25}$~(\%)~$\uparrow$} & {MSLE~($10^{-2}$)~$\downarrow$} & {MAE~(m)~$\downarrow$} & {RMSE~(m)~$\downarrow$} & {F1~(\%)~$\uparrow$} & {Acc~(\%)~$\uparrow$} \\
            \midrule
            ConvNext-B~\cite{liu2022convnet} & 175 & 85.91(96) & 2.53(07) & 3.68(09) & 4.75(08) & 45.99(55) & \ulmain{48.16}{0.63} \\
            SwinV2-B~\cite{liu2022swin} & 174 & 86.38(69) & 2.53(09) & 3.51(07) & 4.64(07) & 45.73(1.21) & 46.94(1.42) \\
            PECore-B~\cite{bolya2025perception} & 187 & \ulmain{89.22}{0.49} & \ulmainsmall{2.18}{0.07} & \ulmainsmall{3.21}{0.04} & \ulmainsmall{4.24}{0.06} & 41.92(1.94) & 42.31(1.59) \\
            DINOv3-B~\cite{simeoni2025dinov3} & \ulint{171} & 85.74(69) & 2.61(09) & 3.61(07) & 4.69(07) & \ulmain{46.44}{1.32} & 46.44(1.32) \\
            \rowcolor{gray!25} \method-B (Ours) & \bfseries 100 & \bfmain{91.25}{0.40} & \bfmainsmall{1.95}{0.05} & \bfmainsmall{3.11}{0.08} & \bfmainsmall{4.14}{0.10} & \bfmain{48.96}{0.85} & \bfmain{50.22}{1.00} \\
            \midrule
            ConvNext-L~\cite{liu2022convnet} & \ulint{392} & 86.14(74) & 2.49(07) & 3.63(10) & 4.71(09) & 46.98(70) & 48.79(71) \\
            PECore-L~\cite{bolya2025perception} & 634 & \ulmain{91.88}{0.49} & \ulmainsmall{1.93}{0.03} & \ulmainsmall{3.01}{0.02} & \ulmainsmall{3.99}{0.03} & 51.93(1.17) & 52.97(59) \\
            DINOv3-L~\cite{simeoni2025dinov3} & 606 & 86.78(1.39) & 2.32(12) & 3.41(10) & 4.44(11) & \ulmain{53.49}{0.61} & \ulmain{53.87}{1.20} \\
            \rowcolor{gray!25} \method-L (Ours) & \bfseries 328 & \bfmain{94.14}{0.23} & \bfmainsmall{1.68}{0.01} & \bfmainsmall{2.75}{0.01} & \bfmainsmall{3.75}{0.02} & \bfmain{54.39}{0.67} & \bfmain{55.58}{0.63} \\
            \bottomrule
        \end{tabular}
    }
\end{table}

\begin{table}[!tb]
    \caption{
        \textbf{Comparative results on the Quebec Plantations test split.}
        We compare \method against competing methods in terms of parameter count, height estimation and classification.
        We group models by parameter count: fewer than 300M (top) and more than 300M (bottom).
        For competing methods, `Param.' denotes combined parameters of two independently-trained task-specific models.
        We report the mean ± standard error over 5 seeds.
        \textbf{Bold} and \underline{underline} denote best and second-best.
    }
    \label{tab:comparative_results_qp}
    \centering
    \setlength{\tabcolsep}{3pt}
    \sisetup{table-number-alignment=center, separate-uncertainty=true, retain-zero-uncertainty=true}
    \renewcommand{\arraystretch}{1.2}
    \resizebox{0.99\textwidth}{!}{%
        \begin{tabular}{
        l
        S[table-format=3]
        S[table-format=2.2(2), separate-uncertainty]
        S[table-format=1.2(2), separate-uncertainty]
        S[table-format=1.2(2), separate-uncertainty]
        S[table-format=1.2(2), separate-uncertainty]
        S[table-format=2.2(2), separate-uncertainty]
        S[table-format=2.2(2), separate-uncertainty]}
            \toprule
            & & \multicolumn{4}{c}{\textbf{Height Estimation}} & \multicolumn{2}{c}{\textbf{Classification}} \\
            \cmidrule(lr){3-6} \cmidrule(lr){7-8}
            \multirow{-2}{*}{\vspace{0.5em}\textbf{Method}} & {\multirow{-2}{*}{\textbf{\shortstack[c]{Param.\\(M)~$\downarrow$}}}} & {$\delta_{1.25}$~(\%)~$\uparrow$} & {MSLE~($10^{-2}$)~$\downarrow$} & {MAE~(m)~$\downarrow$} & {RMSE~(m)~$\downarrow$} & {F1~(\%)~$\uparrow$} & {Acc~(\%)~$\uparrow$} \\
            \midrule
            ConvNext-B~\cite{liu2022convnet} & 175 & 84.70(31) & 1.63(04) & 0.39(00) & 0.61(01) & \ulmain{83.78}{1.47} & \bfmain{85.95}{1.47} \\
            SwinV2-B~\cite{liu2022swin} & 174 & 80.37(58) & 2.16(11) & 0.47(01) & 0.71(02) & 79.33(2.34) & 80.21(2.22) \\
            PECore-B~\cite{bolya2025perception} & 187 & 84.57(31) & 1.56(04) & 0.39(01) & 0.58(01) & \bfmain{84.34}{1.36} & \ulmain{84.67}{1.92} \\
            DINOv3-B~\cite{simeoni2025dinov3} & \ulint{171} & \bfmain{86.71}{0.39} & \bfmainsmall{1.48}{0.04} & \bfmainsmall{0.37}{0.00} & \ulmainsmall{0.58}{0.01} & 82.56(1.86) & 84.00(1.76) \\
            \rowcolor{gray!25} \method-B (Ours) & \bfseries 100 & \ulmain{85.13}{0.40} & \ulmainsmall{1.54}{0.03} & \ulmainsmall{0.38}{0.00} & \bfmainsmall{0.57}{0.01} & 81.88(1.20) & 83.97(45) \\
            \midrule
            ConvNext-L~\cite{liu2022convnet} & \ulint{392} & 84.34(32) & 1.65(02) & 0.40(00) & 0.61(01) & 84.46(45) & 85.25(87) \\
            PECore-L~\cite{bolya2025perception} & 634 & \bfmain{86.74}{0.74} & \ulmainsmall{1.44}{0.04} & \bfmainsmall{0.36}{0.01} & \ulmainsmall{0.58}{0.01} & \ulmain{86.61}{0.70} & \ulmain{86.90}{1.35} \\
            DINOv3-L~\cite{simeoni2025dinov3} & 606 & \ulmain{86.26}{0.59} & \bfmainsmall{1.39}{0.03} & \bfmainsmall{0.36}{0.00} & \bfmainsmall{0.57}{0.01} & 85.94(37) & 85.80(38) \\
            \rowcolor{gray!25} \method-L (Ours) & \bfseries 328 & 85.09(26) & 1.47(02) & 0.37(00) & \ulmainsmall{0.58}{0.01} & \bfmain{87.80}{1.19} & \bfmain{87.63}{1.33} \\
            \bottomrule
        \end{tabular}
    }
\end{table}

\subsection{Comparative Results \label{subsec:comp}}
We compare our method to the competing methods from \cref{subsec:baselines}, using Quebec Trees as the primary development dataset (\cref{tab:comparative_results_qt}).
We extend our comparison to BCI (\cref{tab:comparative_results_bci}) and Quebec Plantations (\cref{tab:comparative_results_qp}) for the top-four methods. 
Additionally, we examine the scaling behavior by evaluating any method ranking first or second in any metric at Base scale with a Large backbone (\cref{tab:comparative_results_qt,tab:comparative_results_bci,tab:comparative_results_qp}).

\paragraph{Modern models surpass in-domain approaches.}
Learned models yield substantially more accurate height estimates than traditional allometric equations, even when the latter have access to ground truth species labels and crown segmentation. 
Mask R-CNN, used in prior work for individual tree height estimation~\cite{hao2021automated, fu2024automatic}, achieves a threshold accuracy of $\delta_{1.25} \approx 74 \%$, lagging behind modern methods ($\delta_{1.25} > 80 \%$).

\paragraph{Foundation models vs. others.}
On the Quebec Trees dataset (\cref{tab:comparative_results_qt}), incorporating \gls{dsm} as an additional input modality for ResNet-50 yields results comparable to RGB-only ResNet-50.
While ConvNeXt, SwinV2, and MambaVision achieve similar classification accuracy ($\text{F1} \approx 84.9 \%$, $85.8 \%$ and $85.4 \%$), ConvNeXt yields superior height estimation with significantly lower MAE (\SI{1.32}{m} vs. \SI{1.73}{m} and \SI{2.36}{m}).
Fine-tuned \glspl{vfm} (PECore, DINOv3) yield substantially better results than models with mainstream backbones and the remote sensing foundation model AnySat, suggesting AnySat's advantages may lie in temporal data or non-RGB spectral bands.

\paragraph{Fine-tuning matters for \glspl{vfm}.} 
Despite \glspl{vfm} being designed for generalization without fine-tuning \cite{simeoni2025dinov3}, frozen DINOv3 yields poor results. Interestingly, the model with the satellite-pretrained backbone is worse than the web-pretrained version across all metrics. 
Fine-tuning these \glspl{vfm} entirely, even on small annotated datasets (\cref{subsec:datasets}), remains feasible and achieves competitive results.

\paragraph{\method ranks among the best methods overall.}
\method achieves state-of-the-art height estimation and ranks first or second for classification (with comparable results within standard error) on both Quebec Trees (\cref{tab:comparative_results_qt}) and BCI (\cref{tab:comparative_results_bci}).
While DINOv3-B ranks first for classification on Quebec Trees, it requires training two separate models, whereas \method-B achieves similar accuracy with $58\%$ of the parameters using a single model. 
On Quebec Plantations (\cref{tab:comparative_results_qp}), DINOv3 yields better height estimation accuracy, though the margin falls within LiDAR ground truth measurement uncertainty~\cite{lefebvre2025suivi}.
On this dataset, Base-size PECore and ConvNeXt yield the highest classification accuracy, yet \method surpasses them when utilizing a Large backbone.

\paragraph{Scaling up backbones improves results.}
Scaling \gls{vit} backbones from Base to Large consistently improves results across the benchmark (\cref{tab:comparative_results_qt,tab:comparative_results_bci,tab:comparative_results_qp}), except for height estimation on Quebec Plantations, where height metrics remain within the same standard error (\cref{tab:comparative_results_qp}). 
\method-L ranks first or second overall while using only $54\%$ of DINOv3-L's parameters. 
On BCI (\cref{tab:comparative_results_bci}), \method-L achieves a threshold accuracy $\delta_{1.25}$ more than $7$ percentage points higher than DINOv3-L.
\method exhibits promising scaling behavior for classification, having an accuracy higher than all baselines when scaled to \gls{vit}-L.

\pagebreak
\subsection{Analyzing \method}
\label{subsec:analysis_method}

\paragraph{Component analysis.}
\begin{wraptable}[16]{r}{0.54\textwidth}
    \vspace{-50pt}
    \caption{
        \textbf{\method-B component analysis} on the Quebec Trees dataset.
        We add or remove components of the task-specific heads, using `(h)' and `(c)' to denote changes specific to height estimation and classification heads, respectively.
        We also evaluate the DINOv3 cls token `[CLS]' with a single linear layer per task.
        We report the mean ± standard error over 5 seeds.
        \textbf{Bold} and \underline{underline} denote best and second-best.
    }
    \label{tab:component_analysis}
    \centering
    \setlength{\tabcolsep}{3pt}
    \scriptsize
    \sisetup{table-number-alignment=center, separate-uncertainty=true, retain-zero-uncertainty=true}
    \renewcommand{\arraystretch}{1.1}
    \begin{tabular}{
        l
        S[table-format=2.2(2), separate-uncertainty]
        S[table-format=2.2(2), separate-uncertainty]}
        \toprule
        \textbf{Variant} & {$\delta_{1.25}$~(\%)~$\uparrow$} & {F1~(\%)~$\uparrow$} \\
        \midrule
        \rowcolor{gray!25} \method & \ulmain{88.29}{0.32}  & \bfmain{86.89}{0.31} \\
        \hspace{2mm} - w/o mlp & 87.80(30) & 85.00(32) \\
        \hspace{2mm} - w/o pos. encoding & 87.83(30) & 86.33(29) \\
        \hspace{2mm} - w/o task token (c) & 88.04(37) & 85.92(29) \\
        \hspace{2mm} - w/o [CLS] (c) & \bfmain{88.49}{0.26} & 85.76(18)  \\
        \hspace{2mm} - w/ [CLS] (h) & 87.72(26) & \ulmain{86.32}{0.71}  \\
        \hspace{2mm} - w/ sigmoid (h) & 87.75(34) & 86.03(24)  \\
        DINOv3 [CLS], linear & 85.26(11) & 83.88(57) \\
        \bottomrule
    \end{tabular}
\end{wraptable}
In \cref{tab:component_analysis}, we evaluate the contribution of different architectural components within \method-B.
Removing the \gls{mlp} or positional encoding prior to cross-attention causes marginal decline in height estimation but significantly degrades classification accuracy.
For the classification head, we concatenate the cross-attention output with the backbone's [CLS] token (\cref{subsec:architecture}).
Ablating this design reveals F1-score drops from $86.89\%$ (combined) to $85.92\%$ (without task token) and $85.76\%$ (without [CLS]), validating our concatenation strategy.
Concatenating the task token with [CLS] for height estimation yields minimally worse results.
For height estimation's final activation function, ReLU is better than Sigmoid scaled by maximum training/validation height, as Sigmoid biases predictions toward extreme height values.
Comparing our head design against a simple linear layer per task applied to the DINOv3 [CLS] token, our approach significantly improves threshold accuracy from $\delta_{1.25}=85.26\%$ to $88.29\%$ and F1-score from $83.88\%$ to $86.89\%$.
We analyze different backbone choices and two separate versions of \method for both tasks in App.~\ref{app:backbone} and App.~\ref{app:separate}.

\paragraph{Loss weighting analysis.}
\begin{table}[b]
    \begin{minipage}{0.61\textwidth}
        
    \caption{
    \textbf{Loss weighting analysis} for \method-B on Quebec Trees.
    We compare EW, UW, DWA, and PCGrad.
    We report the mean ± standard error over 5 seeds.
    \textbf{Bold} and \underline{underline} denote best and second-best.
    Gray corresponds to our design choice.
    }
    \label{tab:loss_mtl}
    \centering
    \setlength{\tabcolsep}{3pt}
    \scriptsize
    \sisetup{table-number-alignment=center, separate-uncertainty=true, retain-zero-uncertainty=true}
    \renewcommand{\arraystretch}{1.1}
    \begin{tabular}{
        l
        S[table-format=2.2(2), separate-uncertainty]
        S[table-format=1.2(2), separate-uncertainty]
        S[table-format=2.2(2), separate-uncertainty]}
        \toprule
        \textbf{Variant} & {$\delta_{1.25}$~(\%)~$\uparrow$} & {\text{MSLE~(m)~$\downarrow$}} & {F1~(\%)~$\uparrow$} \\
        \midrule
        EW & 87.53(0.47) & 2.00(10) & 85.95(56) \\
        UW & 87.45(86) & 1.97(10) & 86.42(39) \\
        \rowcolor{gray!25} DWA & \bfmain{88.29}{0.32} & \bfmainsmall{1.85}{0.05} & \bfmain{86.89}{0.31} \\
        DWA, PCGrad & \ulmain{88.09}{0.14} & \ulmainsmall{1.91}{0.02} & \ulmain{86.59}{0.53} \\
        \bottomrule
    \end{tabular}
    
    \end{minipage}
    \hfill
    \begin{minipage}{0.36\textwidth}
            \caption{
        \textbf{Bias in height predictions of \method-B.}
        We report the mean ± standard error of the mean signed difference (MSD) over 5 seeds.
    }
    \label{tab:height_bias}
    \centering
    \setlength{\tabcolsep}{3pt}
    \scriptsize
    \sisetup{table-number-alignment=center, separate-uncertainty=true, retain-zero-uncertainty=true}
    \renewcommand{\arraystretch}{1.1}
    \begin{tabular}{
        l
        S[table-format=+1.2(2), separate-uncertainty, retain-explicit-plus]}
        \toprule
        \textbf{Dataset} & \text{MSD $(m)$} \\
        \midrule
        Quebec Trees & +0.40(03) \\
        BCI & -0.91(15) \\
        Quebec Plantations & +0.06(02) \\
        \bottomrule
    \end{tabular}
    \end{minipage}
\end{table}
We investigate various \gls{mtl} loss balancing strategies for \method to optimize joint training of height estimation and classification.
\cref{tab:loss_mtl} compares equal weighting (EW), uncertainty weighting (UW), and dynamic weight average (DWA) (\cref{subsec:loss}).
Empirically, DWA yields the best results across all metrics.
Projecting conflicting gradients (PCGrad) does not yield further improvements in this setting.

\subsection{Analysis of Height Predictions}
\begin{wrapfigure}[14]{r}{0.47\textwidth}
    \vspace{-45pt}
    \centering
    \includegraphics[width=0.9\linewidth]{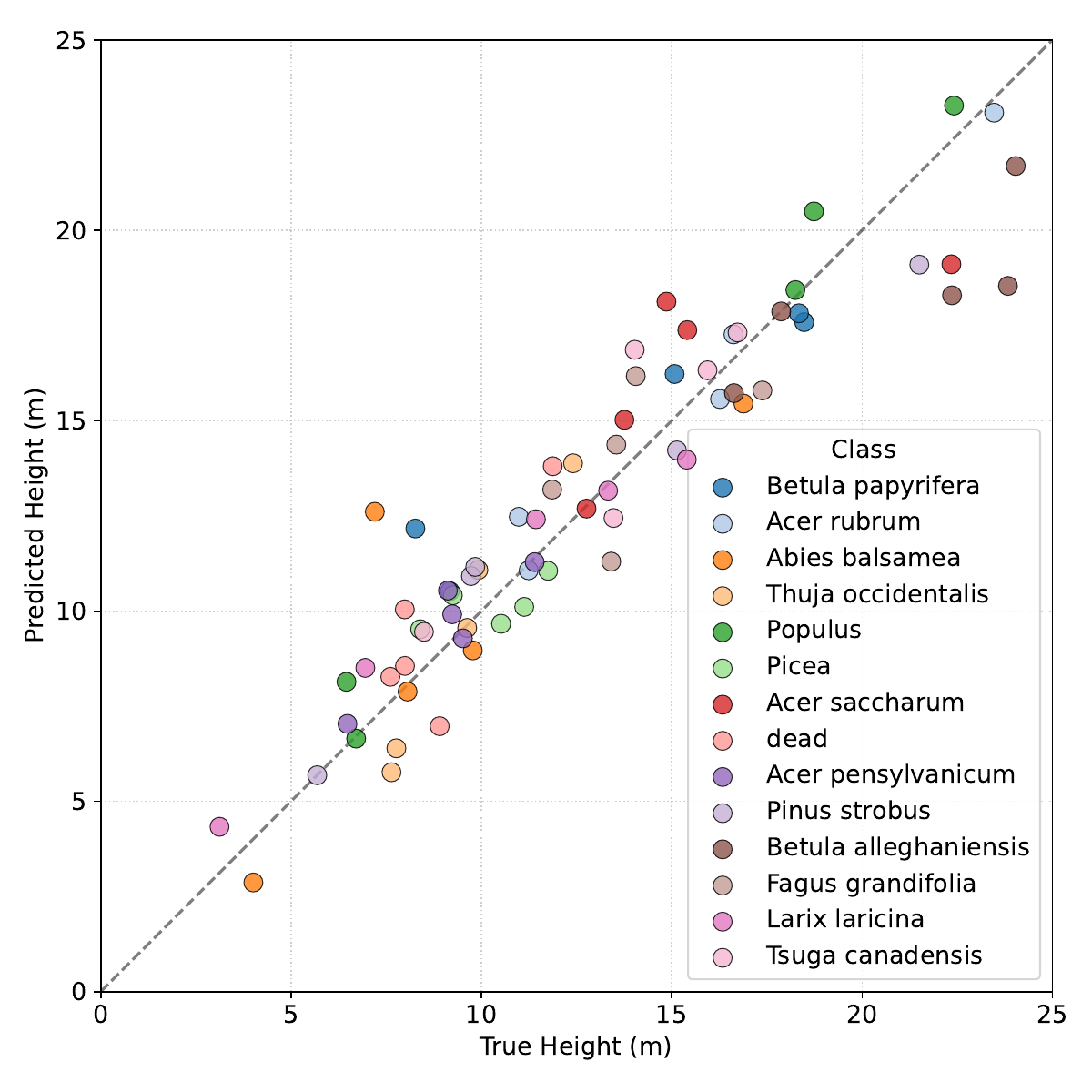}
    \caption{
    \textbf{Comparison of true vs. predicted height}
    using \method-B on the Quebec Trees test split. 
    We select the seed with highest $\delta_{1.25}$ and visualize 5 samples per class.
    The dashed diagonal line represents ideal predictions ($y=x$).
    \label{fig:true_predicted_height}}
\end{wrapfigure}
\cref{tab:height_bias} presents the mean signed difference ($\text{MSD} = \hat{h} - h$) for \method-B across \benchmark.
Height prediction bias is dataset-dependent and strongly correlates with the difference between training and test mean heights (App.~\ref{app:datasets}).
As illustrated in \cref{fig:true_predicted_height}, \method tends to overestimate smaller trees while underestimating larger ones.
Furthermore, class-specific biases emerge: \method overestimates \textit{Populus} on average, but underestimates \textit{Betula alleghaniensis}.
We present additional analyses in App.~\ref{app:results_analysis}.
\section{Conclusion}

We introduced \benchmark, a unified benchmark for individual tree height and species estimation from \gls{uav} imagery.
By evaluating methods across three datasets spanning diverse forest types, we demonstrate broad applicability of our findings.
Our comprehensive evaluation reveals that modern \glspl{vfm} substantially surpass conventional \glspl{cnn} in forest monitoring.
To address data scarcity, we proposed \method, which leverages a shared \gls{vfm} backbone across both tasks, ensuring computational efficiency during training and inference.
\method achieves state-of-the-art height estimation on two of three datasets while maintaining competitive classification performance.
Through extensive ablation studies, we validate our design choices and analyze systematic biases in height predictions.

Our benchmark has limited samples for rare species, resulting in high standard errors across seeds.
Large-scale data collection would enable more robust evaluation of these underrepresented classes.
The creation of a large biomass dataset would enable research on end-to-end biomass estimation from \gls{uav} imagery.
Additionally, as frozen \gls{vfm} backbones fail to achieve competitive results (\cref{tab:comparative_results_qt}), we encourage the community to use \benchmark as a challenging testbed for evaluating out-of-distribution generalization of future \glspl{vfm}.
Future research on semi-supervised learning techniques could exploit abundant unlabeled forest imagery, reducing dependence on limited labeled data.
Incorporating additional modalities such as \glspl{dsm} or photogrammetric point clouds could further improve predictions beyond single RGB images.
Finally, \method could be evaluated on estimating height and class of other objects in drone imagery, such as buildings.

Our work grounds real-world deployment of individual tree height and species estimation.
We hope \benchmark and \method inspire future research in computer vision for automated, large-scale forest monitoring.


\section*{Acknowledgements}
This research was enabled in part by compute resources provided by the Digital Research Alliance of Canada (alliancecan.ca) and Mila (mila.quebec).
We thank Francis Pelletier for his technical support.
We are grateful to funding from IVADO (R3AI, Postdoc Entrepreneur), the Canada CIFAR AI Chairs program, the Canada Research Chair program, the Chaire de recherche Angèle St-Pierre et Hugo Larochelle en IA appliquée à l'environnement, NSERC Discovery, and the Global Center on AI and Biodiversity Change (NSERC 585136).

%
%
\bibliographystyle{splncs04}
\bibliography{main}

\clearpage
\suppressfloats[t]
\appendix

\setcounter{section}{0}
\setcounter{page}{1}

\pagenumbering{roman}

\renewcommand{\theHsection}{Appendix.\Alph{section}}
\renewcommand{\thesection}{\Alph{section}}
\renewcommand{\thesubsection}{\thesection.\arabic{subsection}}

\begin{center}
\Large
\textbf{Estimating Individual Tree Height and Species from UAV Imagery}\\
Supplementary Material
\end{center}

\vspace{12pt}

In this appendix, we further detail the \benchmark datasets (App.~\ref{app:datasets}) and define evaluation metrics (App.~\ref{app:eval}) not covered in \cref{subsec:metrics}.
Additionally, App.~\ref{app:implementation_details} outlines our implementation details.
Finally, we present an analysis of \method's design choices and predictions in App.~\ref{app:additional_experiments}.

\section{Datasets}
\label[appendix]{app:datasets}
In this section, we provide examples of each dataset and further details about the class distribution and height statistics of the datasets.

\subsection{Examples}
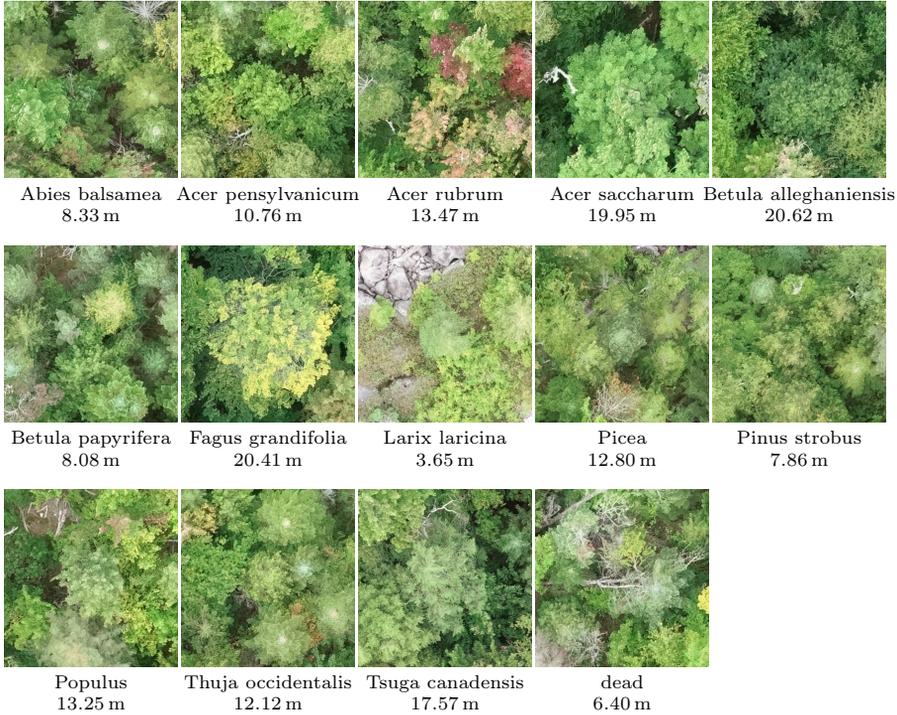
\begin{figure}[!b]
    \centering
    \begin{tikzpicture}
        
        \node[inner sep=0pt] (img2) at (0,0) {\includegraphics[width=0.19\linewidth]{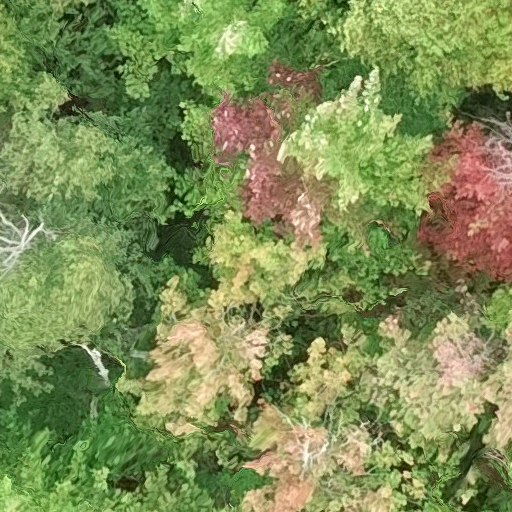}};
        
        \node[inner sep=0pt, anchor=east] (img1) at ([xshift=-1pt]img2.west) {\includegraphics[width=0.19\linewidth]{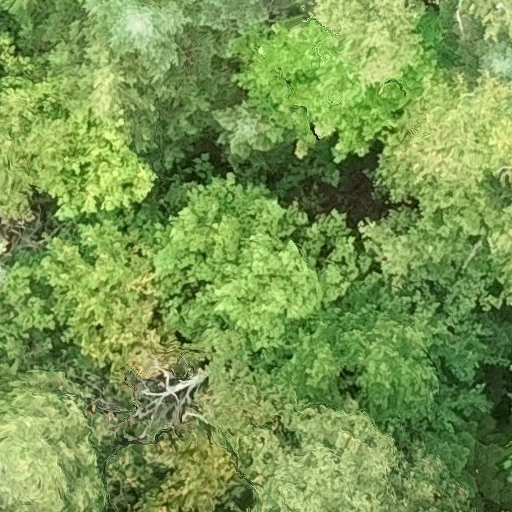}};
        
        \node[inner sep=0pt, anchor=east] (img0) at ([xshift=-1pt]img1.west) {\includegraphics[width=0.19\linewidth]{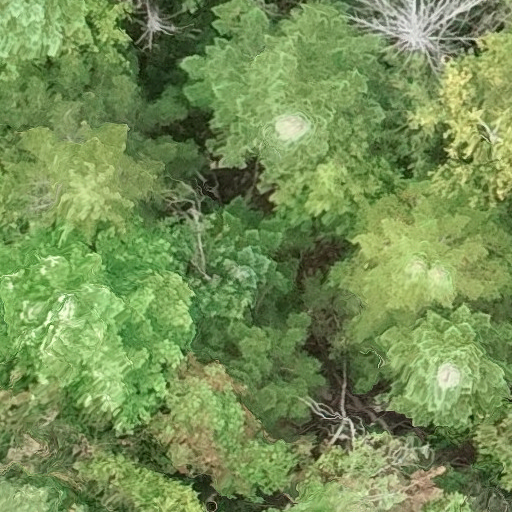}};
        
        \node[inner sep=0pt, anchor=west] (img3) at ([xshift=1pt]img2.east) {\includegraphics[width=0.19\linewidth]{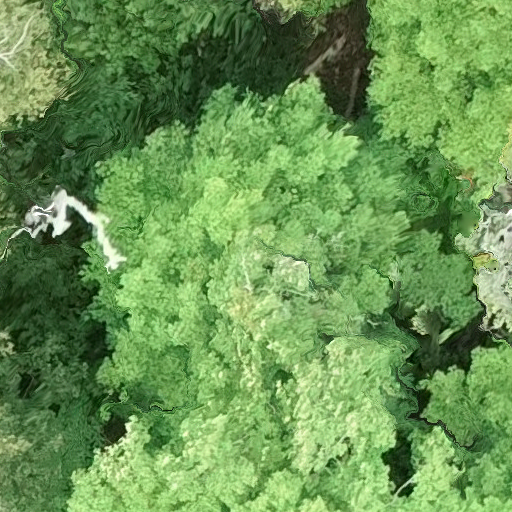}};
        
        \node[inner sep=0pt, anchor=west] (img4) at ([xshift=1pt]img3.east) {\includegraphics[width=0.19\linewidth]{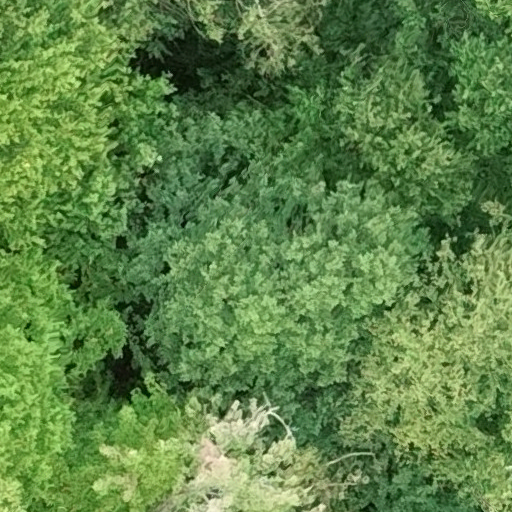}};
        
        \node[inner sep=0pt, anchor=north] (img5) at ([yshift=-25pt]img0.south) {\includegraphics[width=0.19\linewidth]{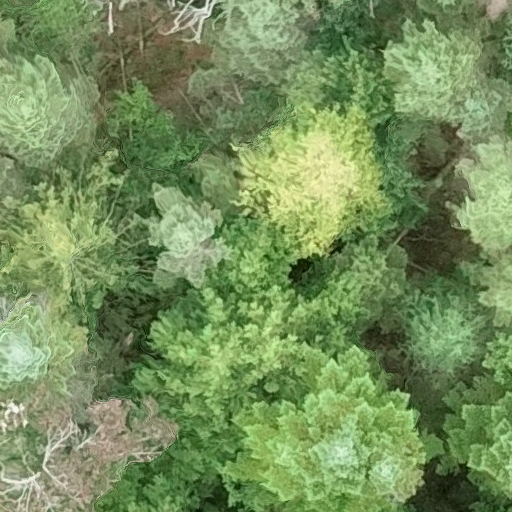}};
        
        \node[inner sep=0pt, anchor=north] (img6) at ([yshift=-25pt]img1.south) {\includegraphics[width=0.19\linewidth]{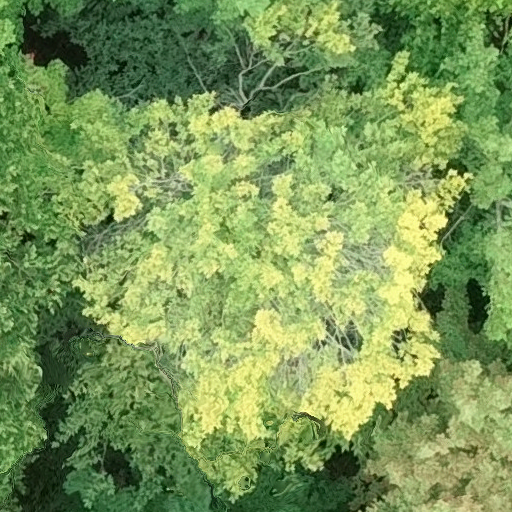}};
        
        \node[inner sep=0pt, anchor=north] (img7) at ([yshift=-25pt]img2.south) {\includegraphics[width=0.19\linewidth]{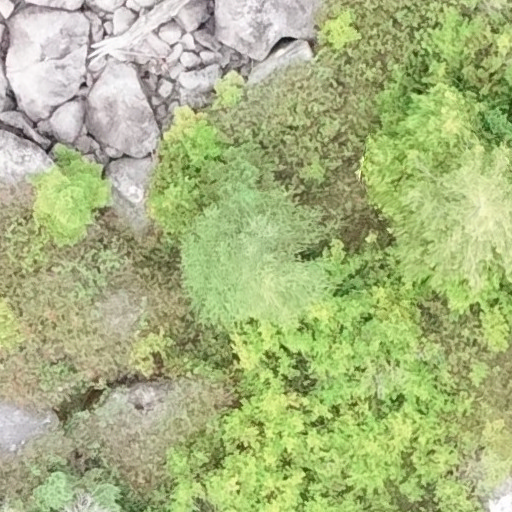}};
        
        \node[inner sep=0pt, anchor=north] (img8) at ([yshift=-25pt]img3.south) {\includegraphics[width=0.19\linewidth]{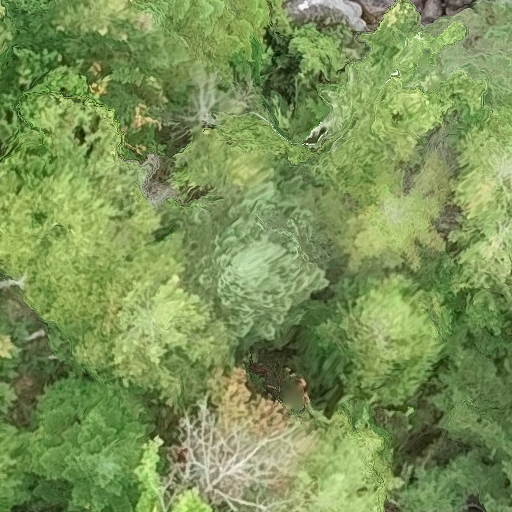}};
        
        \node[inner sep=0pt, anchor=north] (img9) at ([yshift=-25pt]img4.south) {\includegraphics[width=0.19\linewidth]{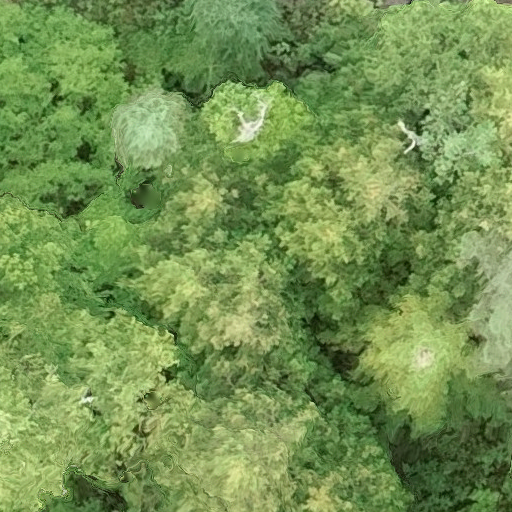}};
        
        \node[inner sep=0pt, anchor=north] (img10) at ([yshift=-25pt]img5.south) {\includegraphics[width=0.19\linewidth]{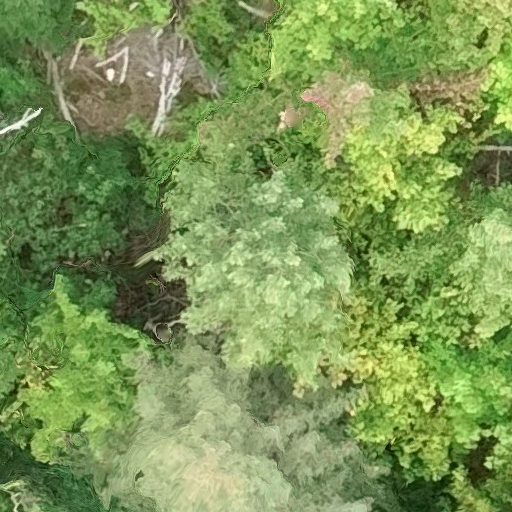}};
        
        \node[inner sep=0pt, anchor=north] (img11) at ([yshift=-25pt]img6.south) {\includegraphics[width=0.19\linewidth]{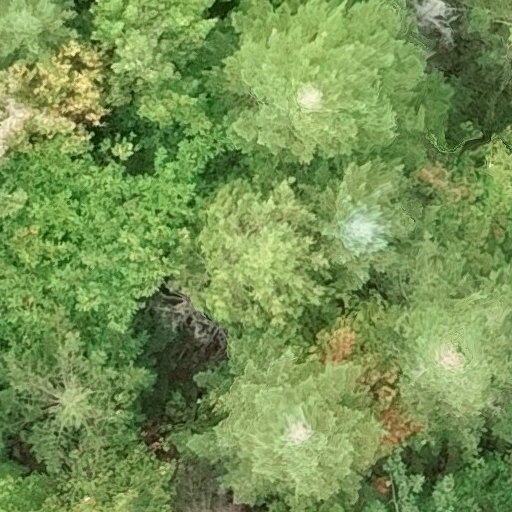}};
        
        \node[inner sep=0pt, anchor=north] (img12) at ([yshift=-25pt]img7.south) {\includegraphics[width=0.19\linewidth]{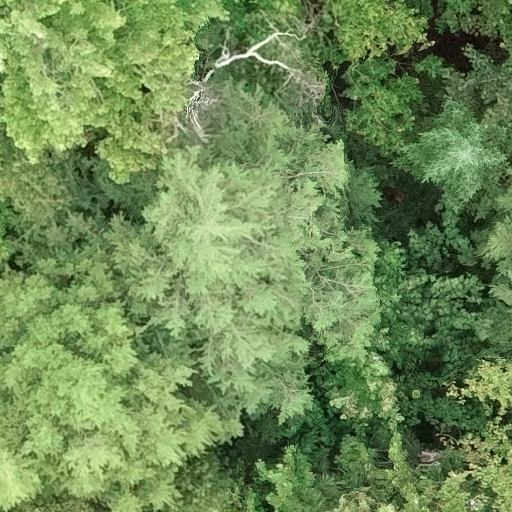}};
        
        \node[inner sep=0pt, anchor=north] (img13) at ([yshift=-25pt]img8.south) {\includegraphics[width=0.19\linewidth]{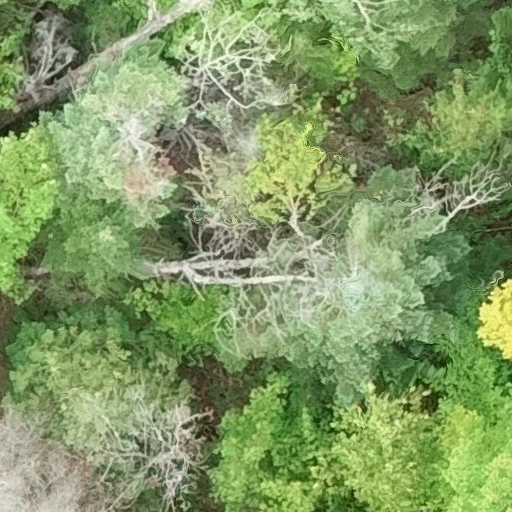}};

        \node[align=center, anchor=north, inner sep=0pt, yshift=-3pt, font=\scriptsize] at (img0.south)
        {Abies balsamea\\\SI{8.33}{m}};
        
        \node[align=center, anchor=north, inner sep=0pt, yshift=-3pt, font=\scriptsize] at (img1.south)
        {Acer pensylvanicum\\\SI{10.76}{m}};
        
        \node[align=center, anchor=north, inner sep=0pt, yshift=-3pt, font=\scriptsize] at (img2.south)
        {Acer rubrum\\\SI{13.47}{m}};
        
        \node[align=center, anchor=north, inner sep=0pt, yshift=-3pt, font=\scriptsize] at (img3.south)
        {Acer saccharum\\\SI{19.95}{m}};
        
        \node[align=center, anchor=north, inner sep=0pt, yshift=-3pt, font=\scriptsize] at (img4.south)
        {Betula alleghaniensis\\\SI{20.62}{m}};
        
        \node[align=center, anchor=north, inner sep=0pt, yshift=-3pt, font=\scriptsize] at (img5.south)
        {Betula papyrifera\\\SI{8.08}{m}};
        
        \node[align=center, anchor=north, inner sep=0pt, yshift=-3pt, font=\scriptsize] at (img6.south)
        {Fagus grandifolia\\\SI{20.41}{m}};
        
        \node[align=center, anchor=north, inner sep=0pt, yshift=-3pt, font=\scriptsize] at (img7.south)
        {Larix laricina\\\SI{3.65}{m}};
        
        \node[align=center, anchor=north, inner sep=0pt, yshift=-3pt, font=\scriptsize] at (img8.south)
        {Picea\\\SI{12.80}{m}};
        
        \node[align=center, anchor=north, inner sep=0pt, yshift=-3pt, font=\scriptsize] at (img9.south)
        {Pinus strobus\\\SI{7.86}{m}};
        
        \node[align=center, anchor=north, inner sep=0pt, yshift=-3pt, font=\scriptsize] at (img10.south)
        {Populus\\\SI{13.25}{m}};
        
        \node[align=center, anchor=north, inner sep=0pt, yshift=-3pt, font=\scriptsize] at (img11.south)
        {Thuja occidentalis\\\SI{12.12}{m}};
        
        \node[align=center, anchor=north, inner sep=0pt, yshift=-3pt, font=\scriptsize] at (img12.south)
        {Tsuga canadensis\\\SI{17.57}{m}};
        
        \node[align=center, anchor=north, inner sep=0pt, yshift=-3pt, font=\scriptsize] at (img13.south)
        {dead\\\SI{6.40}{m}};

    \end{tikzpicture}
    \caption{
    \textbf{Qualitative examples from the Quebec Trees dataset.}
    Each panel displays a representative image from the test split, illustrating one example per class alongside its corresponding ground truth class and height.
    }
    \label{fig:examples_qt}
\end{figure}

For each dataset in \benchmark, we visualize one randomly selected image per class alongside its height value.
\cref{fig:examples_qt} illustrates the 14 classes present in the Quebec Trees dataset.
In \cref{fig:examples_bci}, we display the 21 family-level classes of the BCI dataset, where we observe a notably high inter-class similarity.
Finally, \cref{fig:examples_qp} highlights the Quebec Plantation dataset, demonstrating that most images feature isolated trees without neighboring trees.
\begin{figure}[!tb]
    \centering
    \begin{tikzpicture}
        
        \node[inner sep=0pt] (img2) at (0,0) {\includegraphics[width=0.19\linewidth]{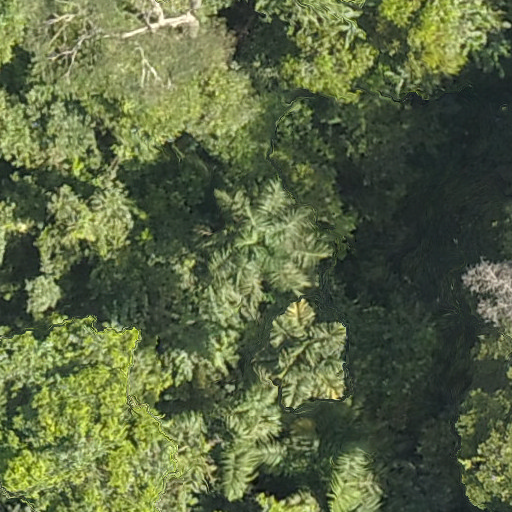}};
        
        \node[inner sep=0pt, anchor=east] (img1) at ([xshift=-1pt]img2.west) {\includegraphics[width=0.19\linewidth]{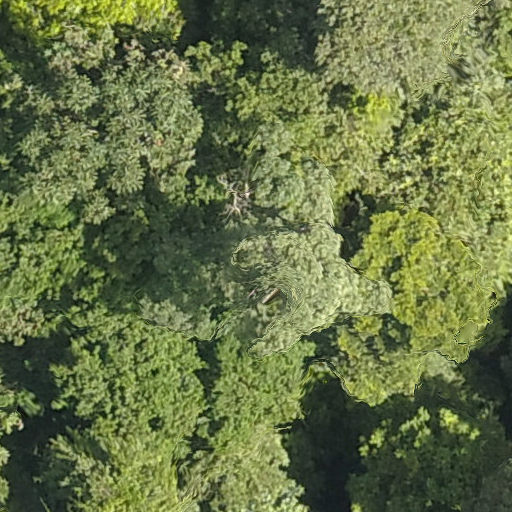}};
        
        \node[inner sep=0pt, anchor=east] (img0) at ([xshift=-1pt]img1.west) {\includegraphics[width=0.19\linewidth]{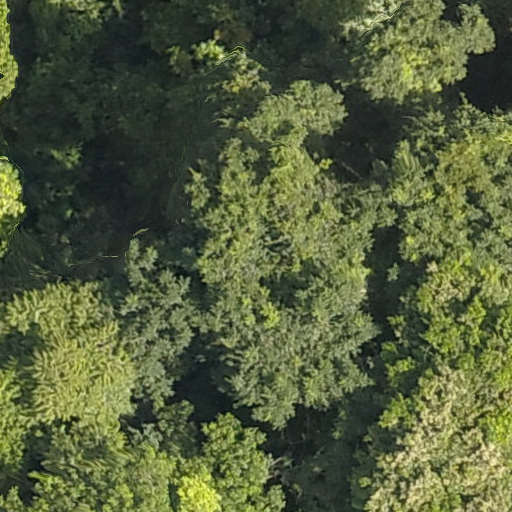}};
        
        \node[inner sep=0pt, anchor=west] (img3) at ([xshift=1pt]img2.east) {\includegraphics[width=0.19\linewidth]{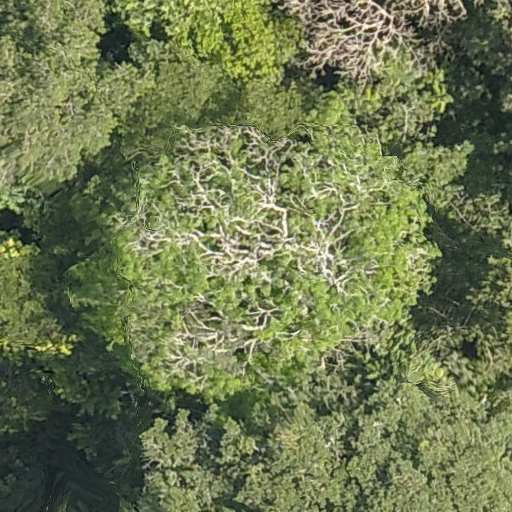}};
        
        \node[inner sep=0pt, anchor=west] (img4) at ([xshift=1pt]img3.east) {\includegraphics[width=0.19\linewidth]{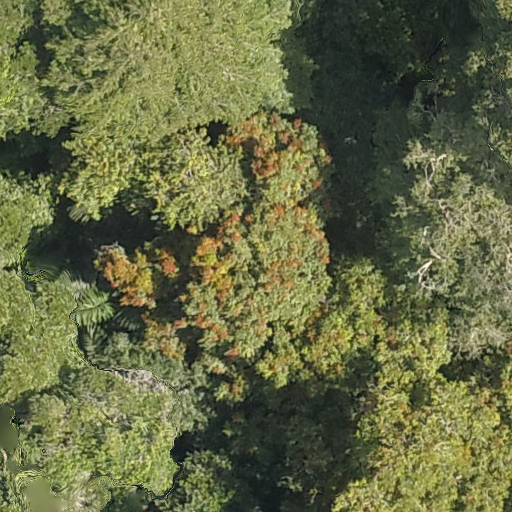}};
        
        \node[inner sep=0pt, anchor=north] (img5) at ([yshift=-25pt]img0.south) {\includegraphics[width=0.19\linewidth]{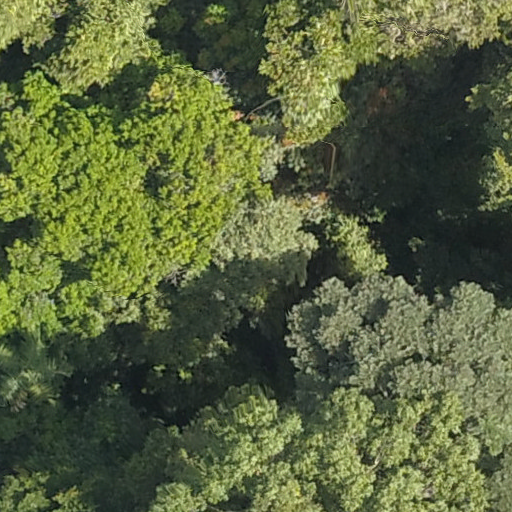}};
        
        \node[inner sep=0pt, anchor=north] (img6) at ([yshift=-25pt]img1.south) {\includegraphics[width=0.19\linewidth]{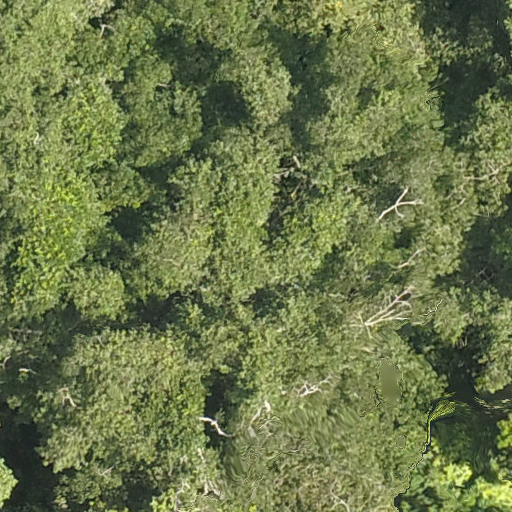}};
        
        \node[inner sep=0pt, anchor=north] (img7) at ([yshift=-25pt]img2.south) {\includegraphics[width=0.19\linewidth]{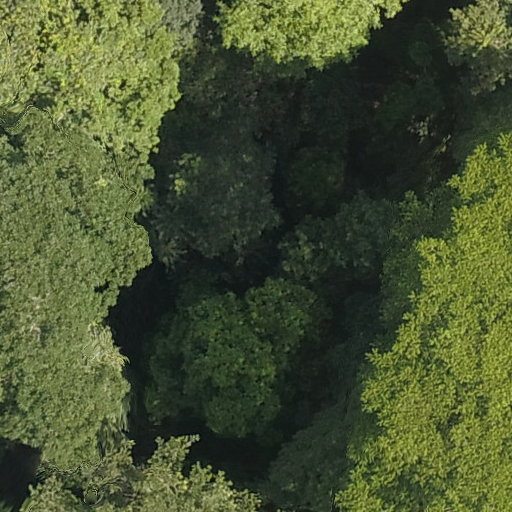}};
        
        \node[inner sep=0pt, anchor=north] (img8) at ([yshift=-25pt]img3.south) {\includegraphics[width=0.19\linewidth]{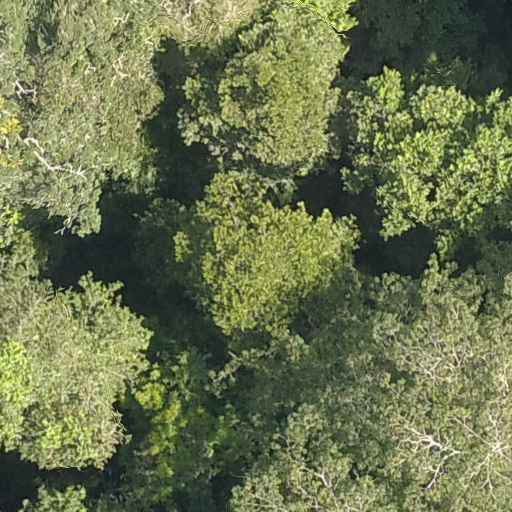}};
        
        \node[inner sep=0pt, anchor=north] (img9) at ([yshift=-25pt]img4.south) {\includegraphics[width=0.19\linewidth]{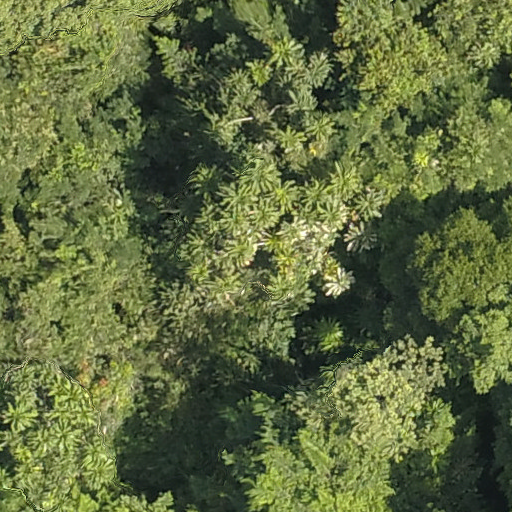}};
        
        \node[inner sep=0pt, anchor=north] (img10) at ([yshift=-25pt]img5.south) {\includegraphics[width=0.19\linewidth]{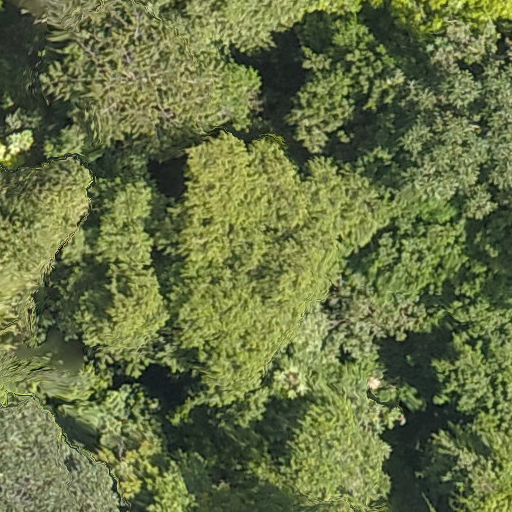}};
        
        \node[inner sep=0pt, anchor=north] (img11) at ([yshift=-25pt]img6.south) {\includegraphics[width=0.19\linewidth]{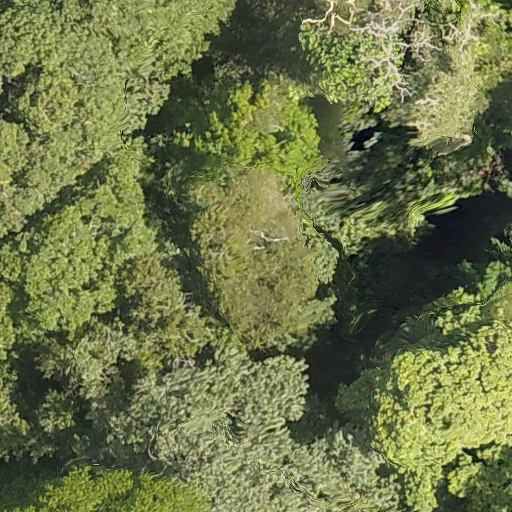}};
        
        \node[inner sep=0pt, anchor=north] (img12) at ([yshift=-25pt]img7.south) {\includegraphics[width=0.19\linewidth]{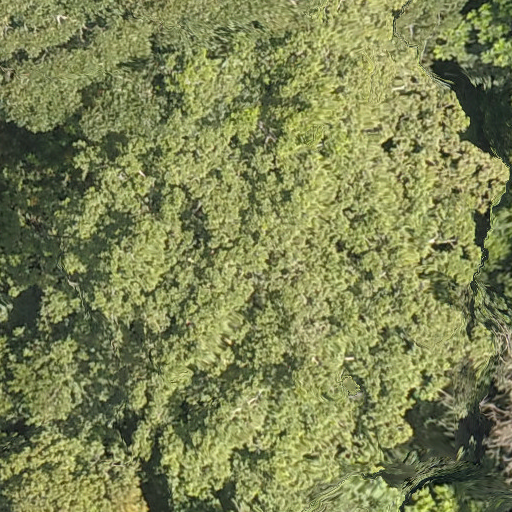}};
        
        \node[inner sep=0pt, anchor=north] (img13) at ([yshift=-25pt]img8.south) {\includegraphics[width=0.19\linewidth]{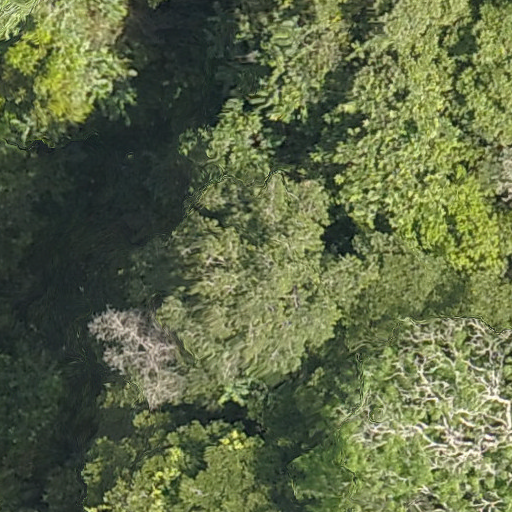}};
        
        \node[inner sep=0pt, anchor=north] (img14) at ([yshift=-25pt]img9.south) {\includegraphics[width=0.19\linewidth]{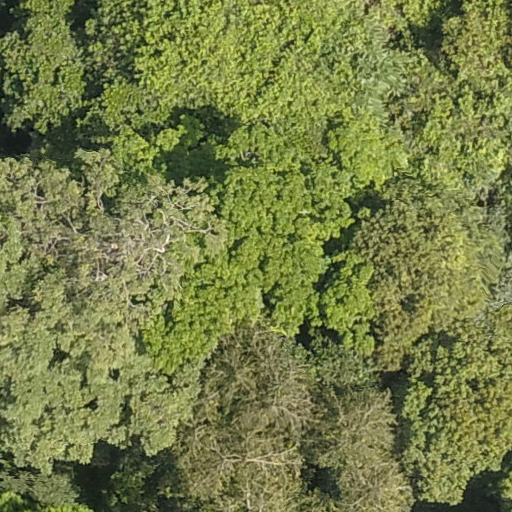}};
        
        \node[inner sep=0pt, anchor=north] (img15) at ([yshift=-25pt]img10.south) {\includegraphics[width=0.19\linewidth]{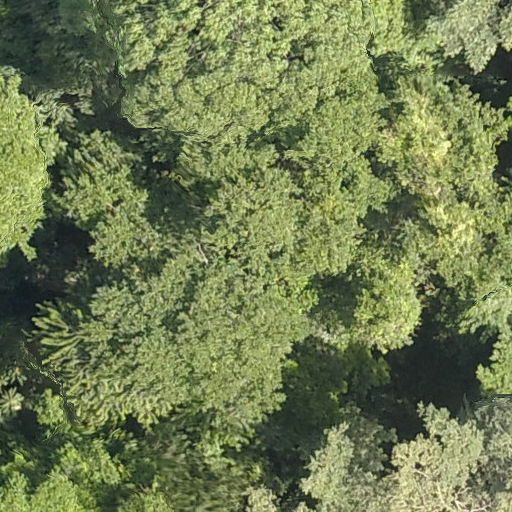}};
        
        \node[inner sep=0pt, anchor=north] (img16) at ([yshift=-25pt]img11.south) {\includegraphics[width=0.19\linewidth]{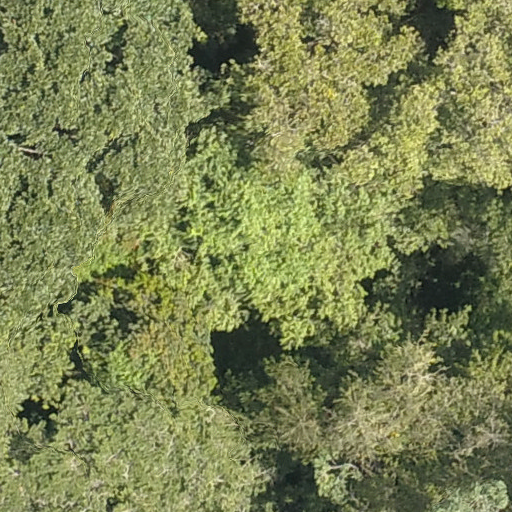}};
        
        \node[inner sep=0pt, anchor=north] (img17) at ([yshift=-25pt]img12.south) {\includegraphics[width=0.19\linewidth]{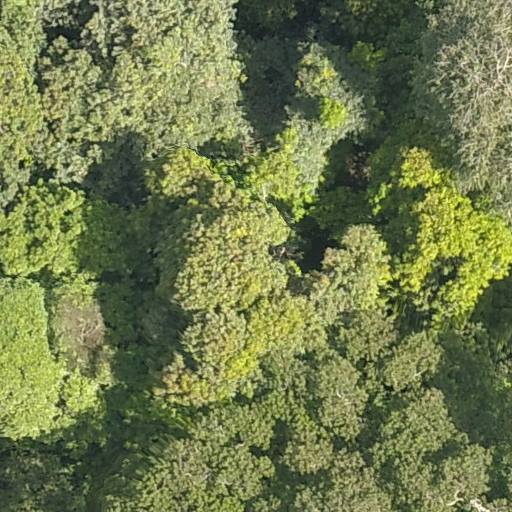}};
        
        \node[inner sep=0pt, anchor=north] (img18) at ([yshift=-25pt]img13.south) {\includegraphics[width=0.19\linewidth]{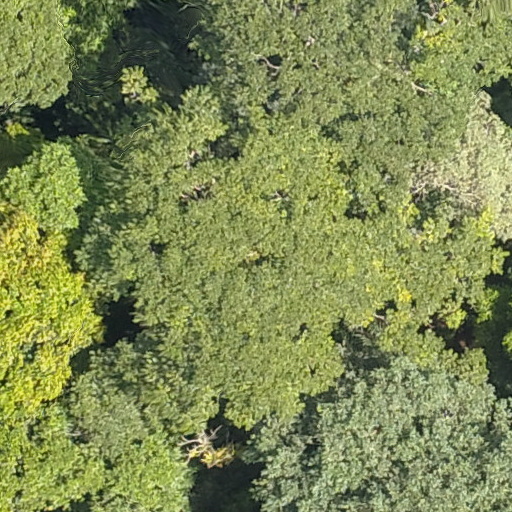}};
        
        \node[inner sep=0pt, anchor=north] (img19) at ([yshift=-25pt]img14.south) {\includegraphics[width=0.19\linewidth]{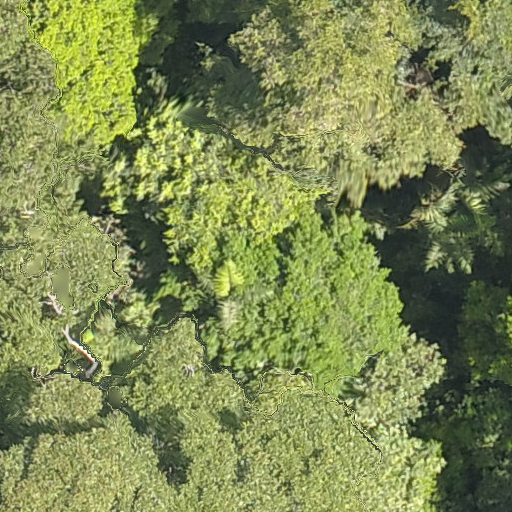}};
        
        \node[inner sep=0pt, anchor=north] (img20) at ([yshift=-25pt]img15.south) {\includegraphics[width=0.19\linewidth]{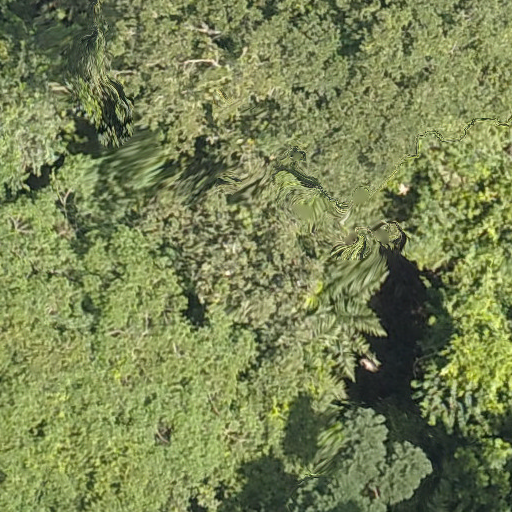}};

        \node[align=center, anchor=north, inner sep=0pt, yshift=-3pt, font=\scriptsize] at (img0.south)
        {Anacardiaceae\\\SI{28.04}{m}};
        
        \node[align=center, anchor=north, inner sep=0pt, yshift=-3pt, font=\scriptsize] at (img1.south)
        {Apocynaceae\\\SI{37.32}{m}};
        
        \node[align=center, anchor=north, inner sep=0pt, yshift=-3pt, font=\scriptsize] at (img2.south)
        {Arecaceae\\\SI{15.48}{m}};
        
        \node[align=center, anchor=north, inner sep=0pt, yshift=-3pt, font=\scriptsize] at (img3.south)
        {Bignoniaceae\\\SI{41.18}{m}};
        
        \node[align=center, anchor=north, inner sep=0pt, yshift=-3pt, font=\scriptsize] at (img4.south)
        {Burseraceae\\\SI{25.99}{m}};
        
        \node[align=center, anchor=north, inner sep=0pt, yshift=-3pt, font=\scriptsize] at (img5.south)
        {Cordiaceae\\\SI{24.39}{m}};
        
        \node[align=center, anchor=north, inner sep=0pt, yshift=-3pt, font=\scriptsize] at (img6.south)
        {Euphorbiaceae\\\SI{36.89}{m}};
        
        \node[align=center, anchor=north, inner sep=0pt, yshift=-3pt, font=\scriptsize] at (img7.south)
        {Fabaceae\\\SI{22.54}{m}};
        
        \node[align=center, anchor=north, inner sep=0pt, yshift=-3pt, font=\scriptsize] at (img8.south)
        {Lauraceae\\\SI{28.80}{m}};
        
        \node[align=center, anchor=north, inner sep=0pt, yshift=-3pt, font=\scriptsize] at (img9.south)
        {Lecythidaceae\\\SI{17.42}{m}};
        
        \node[align=center, anchor=north, inner sep=0pt, yshift=-3pt, font=\scriptsize] at (img10.south)
        {Malvaceae\\\SI{24.61}{m}};
        
        \node[align=center, anchor=north, inner sep=0pt, yshift=-3pt, font=\scriptsize] at (img11.south)
        {Meliaceae\\\SI{29.38}{m}};
        
        \node[align=center, anchor=north, inner sep=0pt, yshift=-3pt, font=\scriptsize] at (img12.south)
        {Moraceae\\\SI{42.40}{m}};
        
        \node[align=center, anchor=north, inner sep=0pt, yshift=-3pt, font=\scriptsize] at (img13.south)
        {Myristicaceae\\\SI{34.43}{m}};
        
        \node[align=center, anchor=north, inner sep=0pt, yshift=-3pt, font=\scriptsize] at (img14.south)
        {Nyctaginaceae\\\SI{28.32}{m}};
        
        \node[align=center, anchor=north, inner sep=0pt, yshift=-3pt, font=\scriptsize] at (img15.south)
        {Phyllanthaceae\\\SI{30.09}{m}};
        
        \node[align=center, anchor=north, inner sep=0pt, yshift=-3pt, font=\scriptsize] at (img16.south)
        {Putranjivaceae\\\SI{30.38}{m}};
        
        \node[align=center, anchor=north, inner sep=0pt, yshift=-3pt, font=\scriptsize] at (img17.south)
        {Rubiaceae\\\SI{32.47}{m}};
        
        \node[align=center, anchor=north, inner sep=0pt, yshift=-3pt, font=\scriptsize] at (img18.south)
        {Rutaceae\\\SI{30.40}{m}};
        
        \node[align=center, anchor=north, inner sep=0pt, yshift=-3pt, font=\scriptsize] at (img19.south)
        {Sapotaceae\\\SI{27.28}{m}};
        
        \node[align=center, anchor=north, inner sep=0pt, yshift=-3pt, font=\scriptsize] at (img20.south)
        {Urticaceae\\\SI{41.36}{m}};

    \end{tikzpicture}
    \caption{
    \textbf{Qualitative examples from the BCI dataset.}
    Each panel displays a representative image from the test split, illustrating one example per class alongside its corresponding ground truth class and height.
    }
    \label{fig:examples_bci}
\end{figure}
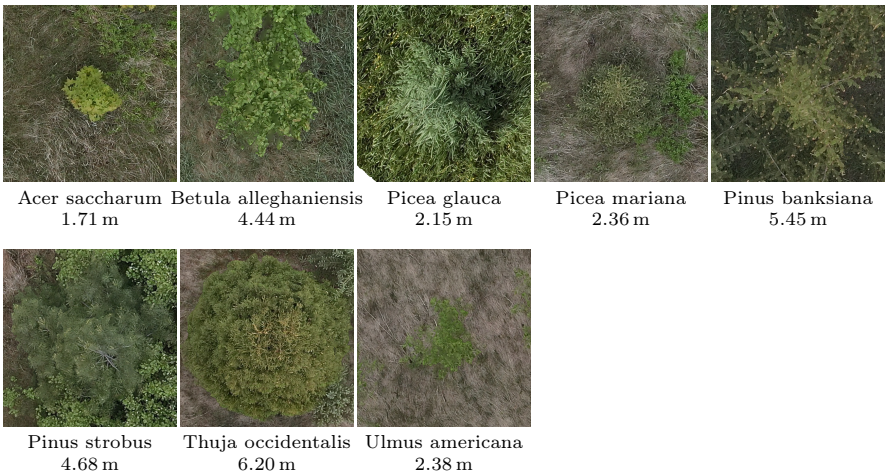
\begin{figure}[!tb]
    \centering
    \begin{tikzpicture}
        
        \node[inner sep=0pt] (img2) at (0,0) {\includegraphics[width=0.19\linewidth]{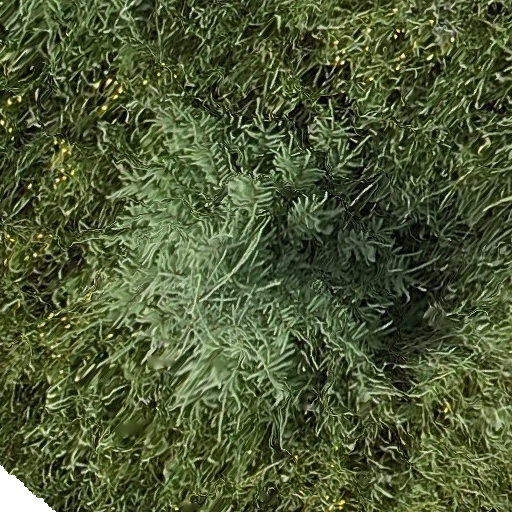}};
        
        \node[inner sep=0pt, anchor=east] (img1) at ([xshift=-1pt]img2.west) {\includegraphics[width=0.19\linewidth]{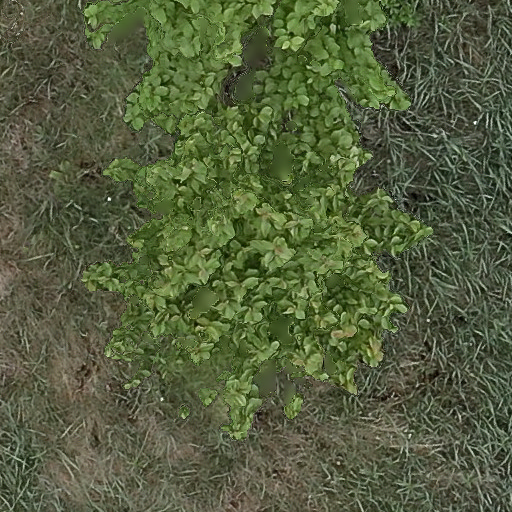}};
        
        \node[inner sep=0pt, anchor=east] (img0) at ([xshift=-1pt]img1.west) {\includegraphics[width=0.19\linewidth]{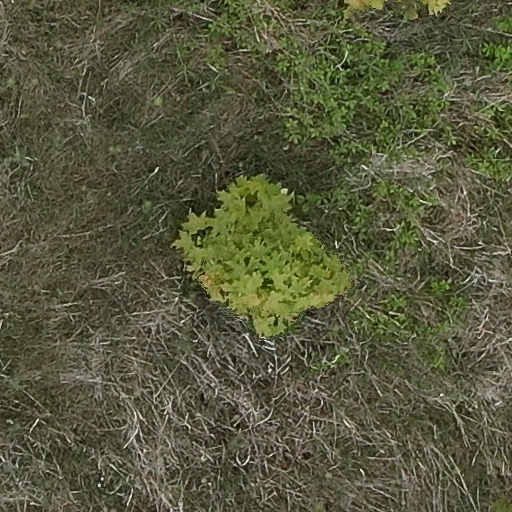}};
        
        \node[inner sep=0pt, anchor=west] (img3) at ([xshift=1pt]img2.east) {\includegraphics[width=0.19\linewidth]{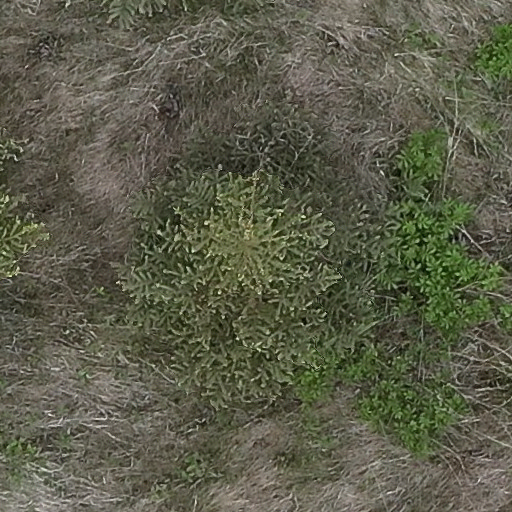}};
        
        \node[inner sep=0pt, anchor=west] (img4) at ([xshift=1pt]img3.east) {\includegraphics[width=0.19\linewidth]{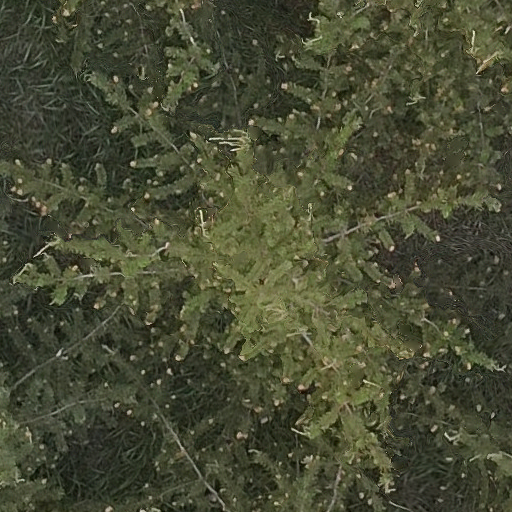}};
        
        \node[inner sep=0pt, anchor=north] (img5) at ([yshift=-25pt]img0.south) {\includegraphics[width=0.19\linewidth]{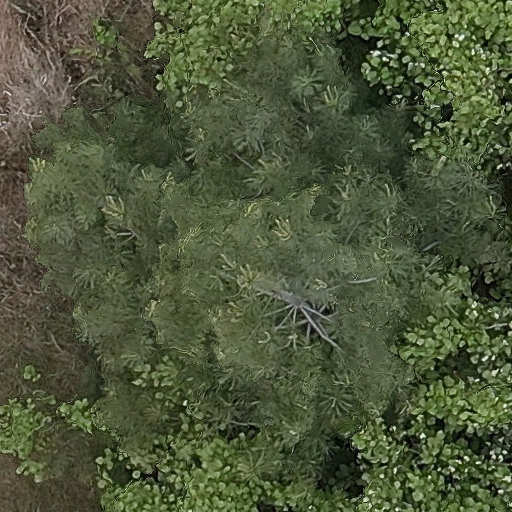}};
        
        \node[inner sep=0pt, anchor=north] (img6) at ([yshift=-25pt]img1.south) {\includegraphics[width=0.19\linewidth]{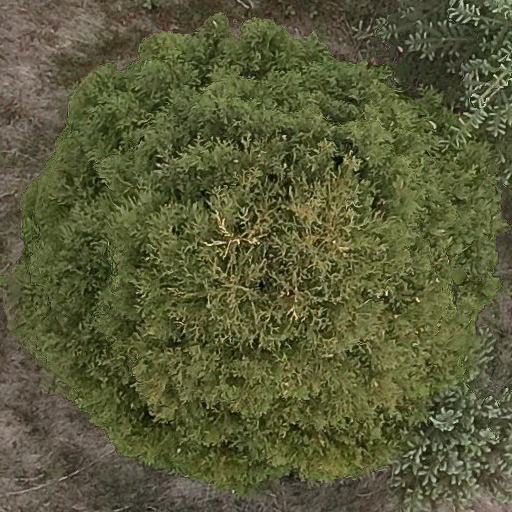}};
        
        \node[inner sep=0pt, anchor=north] (img7) at ([yshift=-25pt]img2.south) {\includegraphics[width=0.19\linewidth]{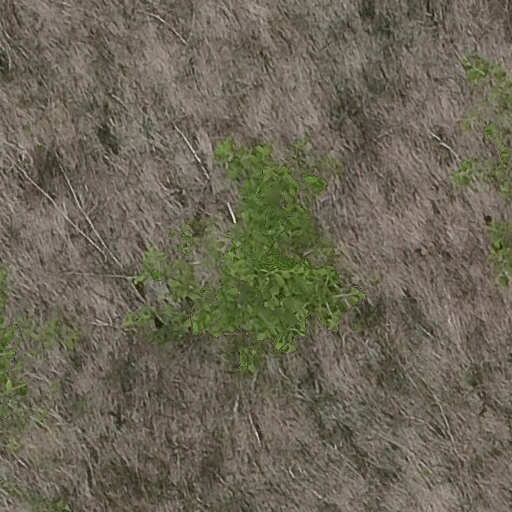}};

        \node[align=center, anchor=north, inner sep=0pt, yshift=-3pt, font=\scriptsize] at (img0.south)
        {Acer saccharum\\\SI{1.71}{m}};
        
        \node[align=center, anchor=north, inner sep=0pt, yshift=-3pt, font=\scriptsize] at (img1.south)
        {Betula alleghaniensis\\\SI{4.44}{m}};
        
        \node[align=center, anchor=north, inner sep=0pt, yshift=-3pt, font=\scriptsize] at (img2.south)
        {Picea glauca\\\SI{2.15}{m}};
        
        \node[align=center, anchor=north, inner sep=0pt, yshift=-3pt, font=\scriptsize] at (img3.south)
        {Picea mariana\\\SI{2.36}{m}};
        
        \node[align=center, anchor=north, inner sep=0pt, yshift=-3pt, font=\scriptsize] at (img4.south)
        {Pinus banksiana\\\SI{5.45}{m}};
        
        \node[align=center, anchor=north, inner sep=0pt, yshift=-3pt, font=\scriptsize] at (img5.south)
        {Pinus strobus\\\SI{4.68}{m}};
        
        \node[align=center, anchor=north, inner sep=0pt, yshift=-3pt, font=\scriptsize] at (img6.south)
        {Thuja occidentalis\\\SI{6.20}{m}};
        
        \node[align=center, anchor=north, inner sep=0pt, yshift=-3pt, font=\scriptsize] at (img7.south)
        {Ulmus americana\\\SI{2.38}{m}};

    \end{tikzpicture}
    \caption{
    \textbf{Qualitative examples from the Quebec Plantations dataset.}
    Each panel displays a representative image from the test split, illustrating one example per class alongside its corresponding ground truth class and height.
    }
    \label{fig:examples_qp}
\end{figure}
\clearpage

\subsection{Class Distribution}
\begin{figure}[!b]
    \centering
    \begin{subfigure}[b]{0.325\linewidth}
        \centering
        \includegraphics[width=\linewidth]{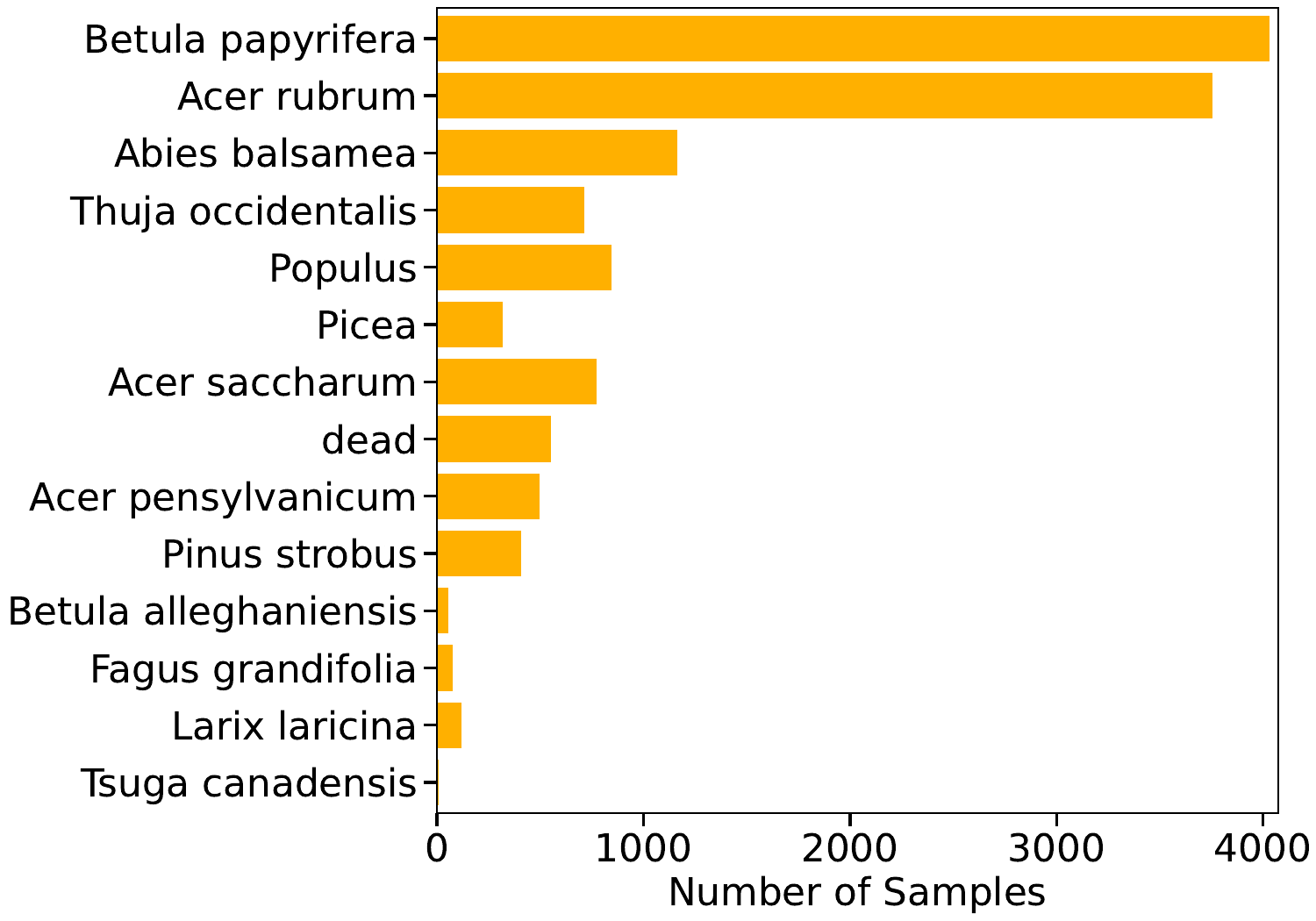}
        \caption{Training split}
        \label{fig:class_dis_qt_train}
    \end{subfigure}
    \hfill
    \begin{subfigure}[b]{0.325\linewidth}
        \centering
        \includegraphics[width=\linewidth]{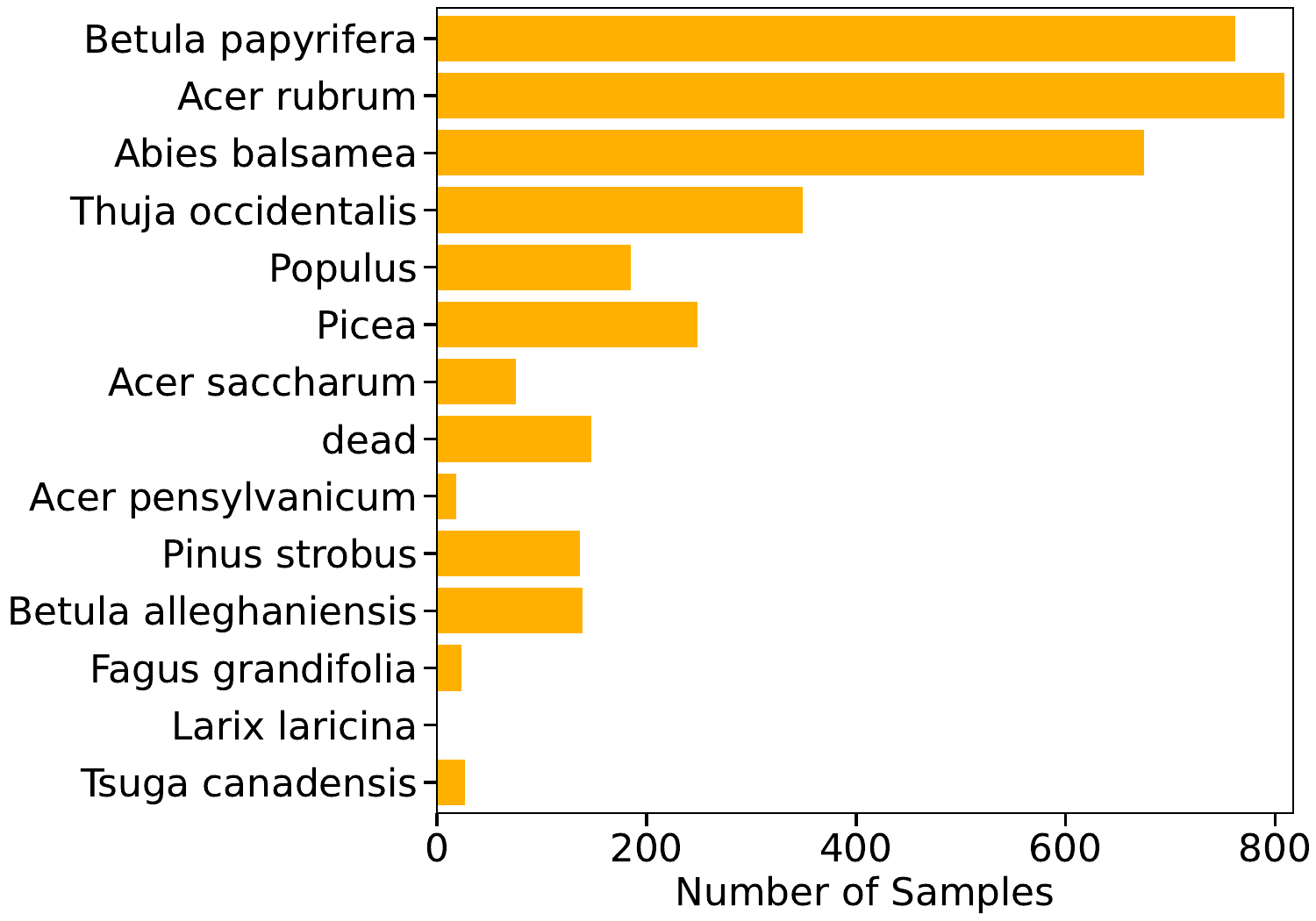}
        \caption{Validation split}
        \label{fig:class_dis_qt_val}
    \end{subfigure}
    \hfill
    \begin{subfigure}[b]{0.325\linewidth}
        \centering
        \includegraphics[width=\linewidth]{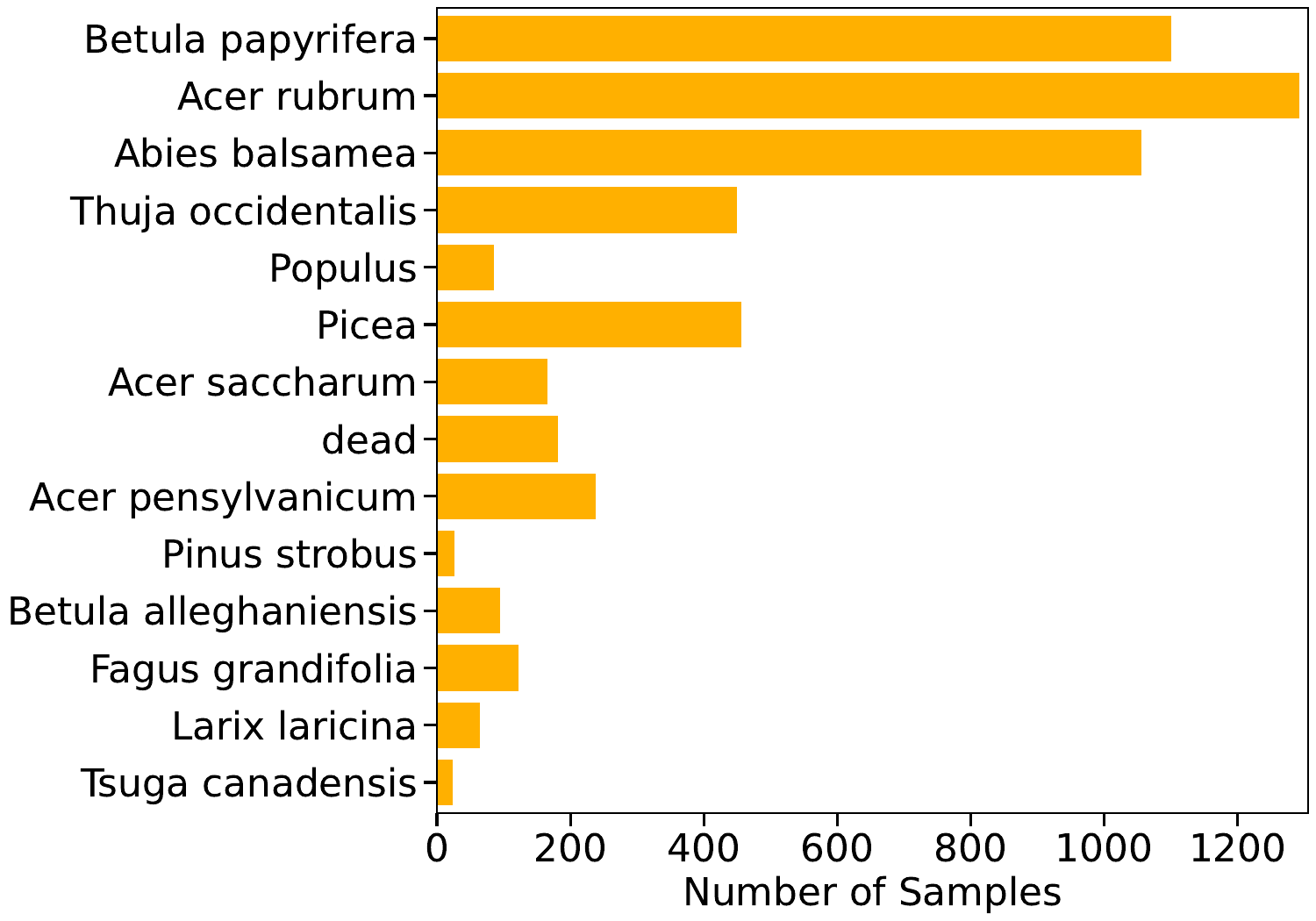}
        \caption{Test split}
        \label{fig:class_dis_qt_test}
    \end{subfigure}
    \caption{
    \textbf{Class distributions across splits for the Quebec Trees dataset.}
    Separate histograms illustrate the number of samples allocated to each class within the (a) training, (b) validation, and (c) test splits.
    }
    \label{fig:class_dis_qt_splits}
\end{figure}

\begin{figure}[t!]
    \centering
    \begin{subfigure}[b]{0.325\linewidth}
        \centering
        \includegraphics[width=\linewidth]{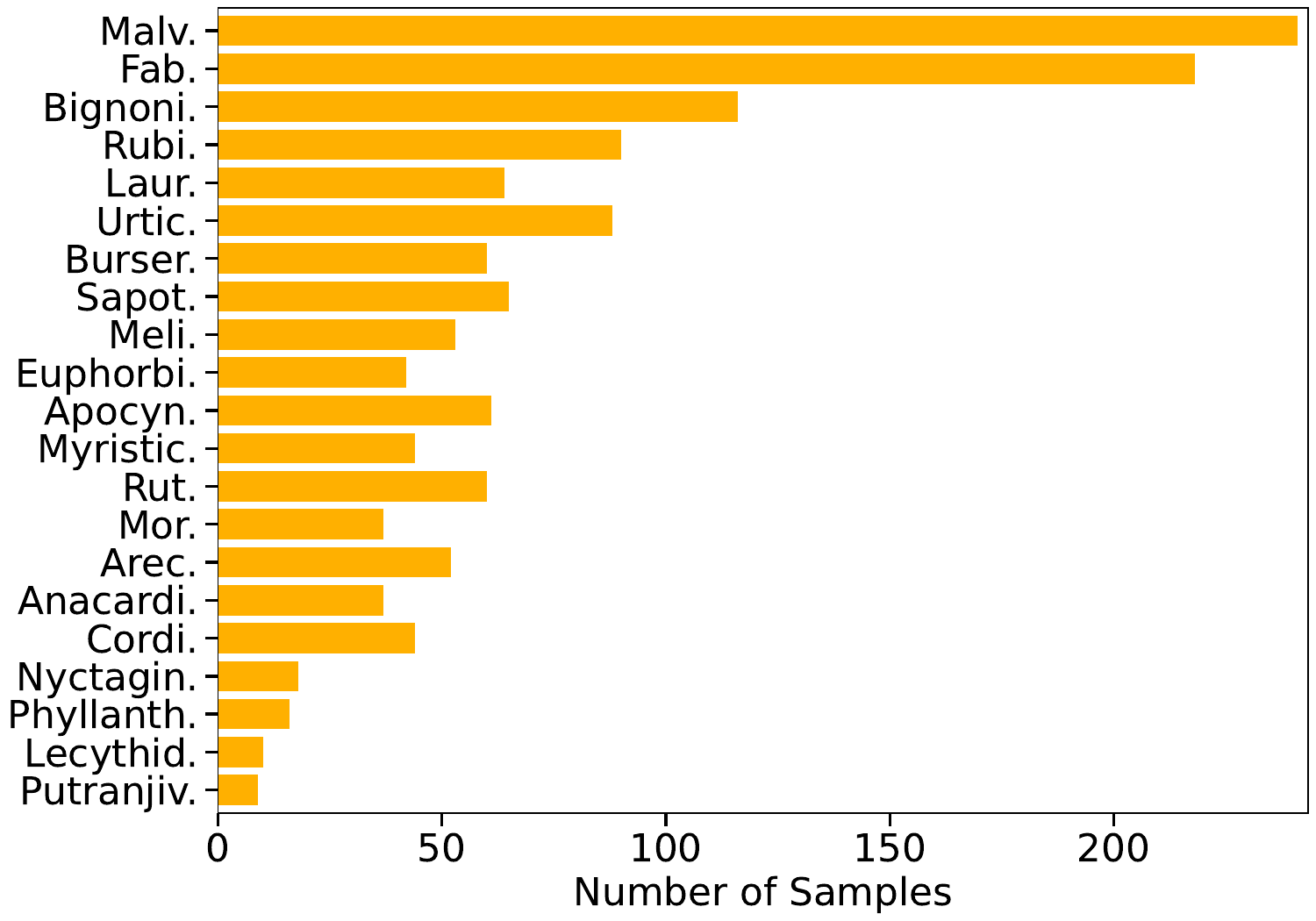}
        \caption{Training split}
        \label{fig:class_dis_bci_train}
    \end{subfigure}
    \hfill
    \begin{subfigure}[b]{0.325\linewidth}
        \centering
        \includegraphics[width=\linewidth]{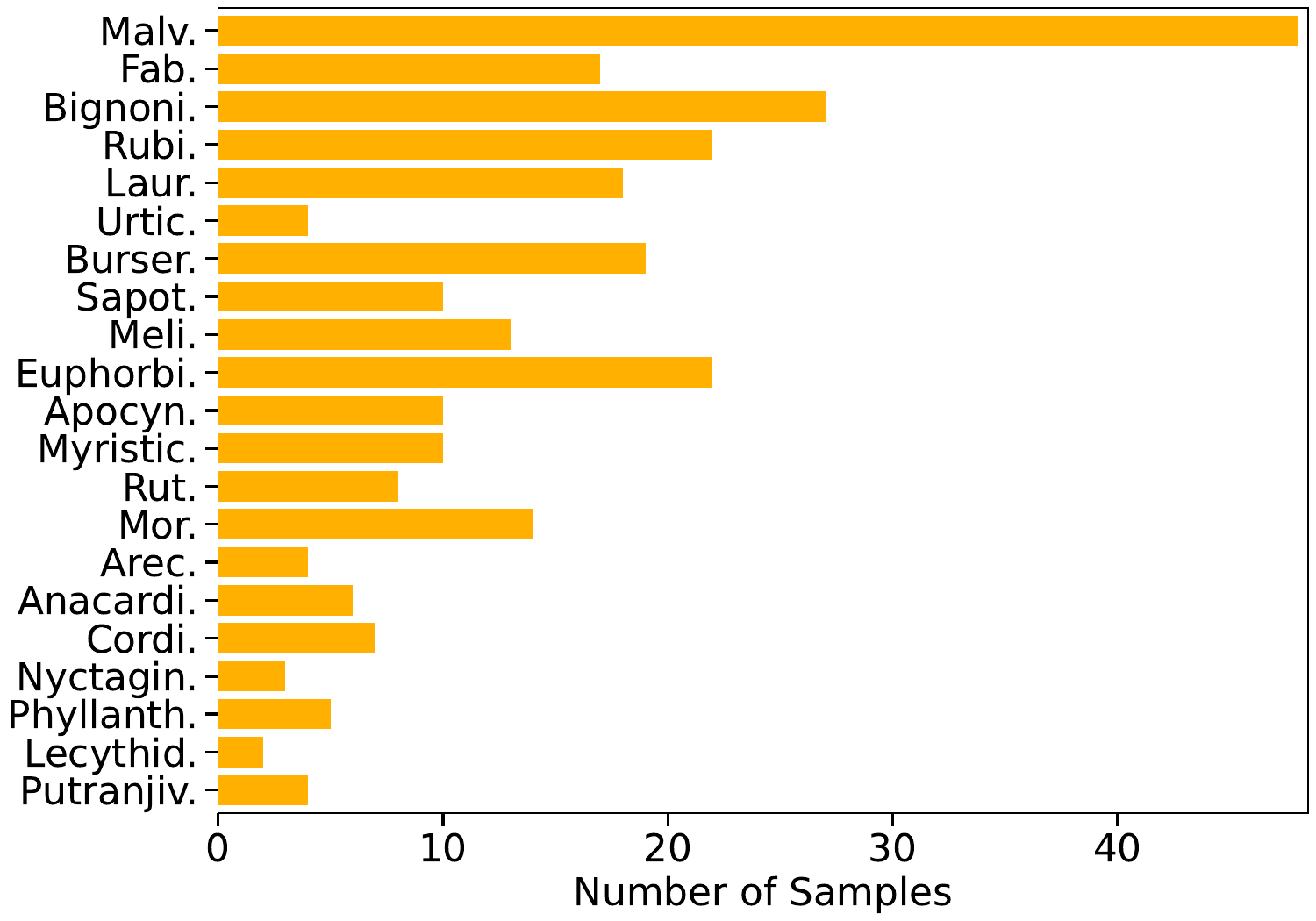}
        \caption{Validation split}
        \label{fig:class_dis_bci_val}
    \end{subfigure}
    \hfill
    \begin{subfigure}[b]{0.325\linewidth}
        \centering
        \includegraphics[width=\linewidth]{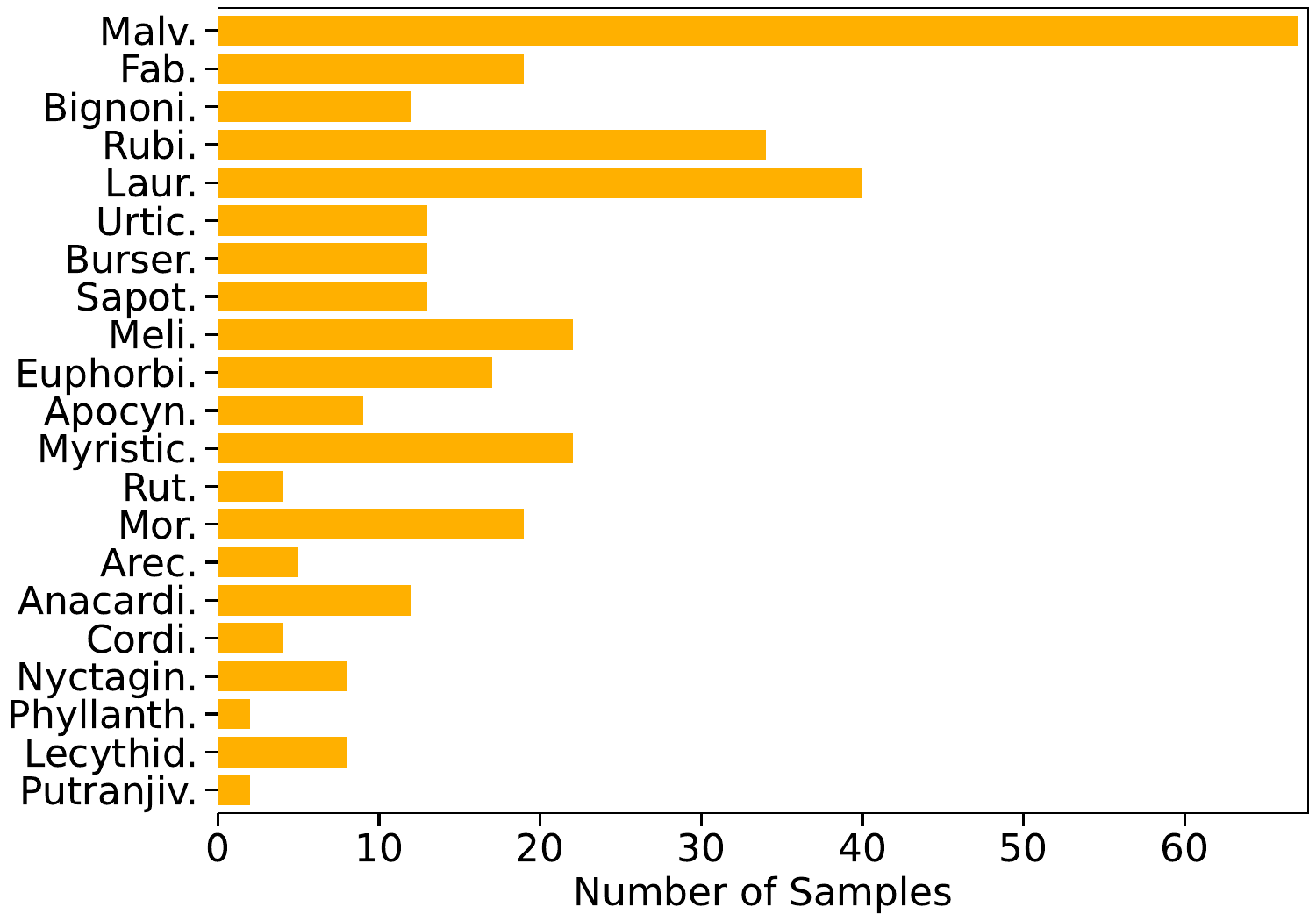}
        \caption{Test split}
        \label{fig:class_dis_bci_test}
    \end{subfigure}
    \caption{
    \textbf{Class distributions across splits for the BCI dataset.}
    Separate histograms illustrate the number of samples allocated to each class within the (a) training, (b) validation, and (c) test splits.
    }
    \label{fig:class_dis_bci_splits}
\end{figure}

\begin{figure}[t]
    \centering
    \begin{subfigure}[b]{0.325\linewidth}
        \centering
        \includegraphics[width=\linewidth]{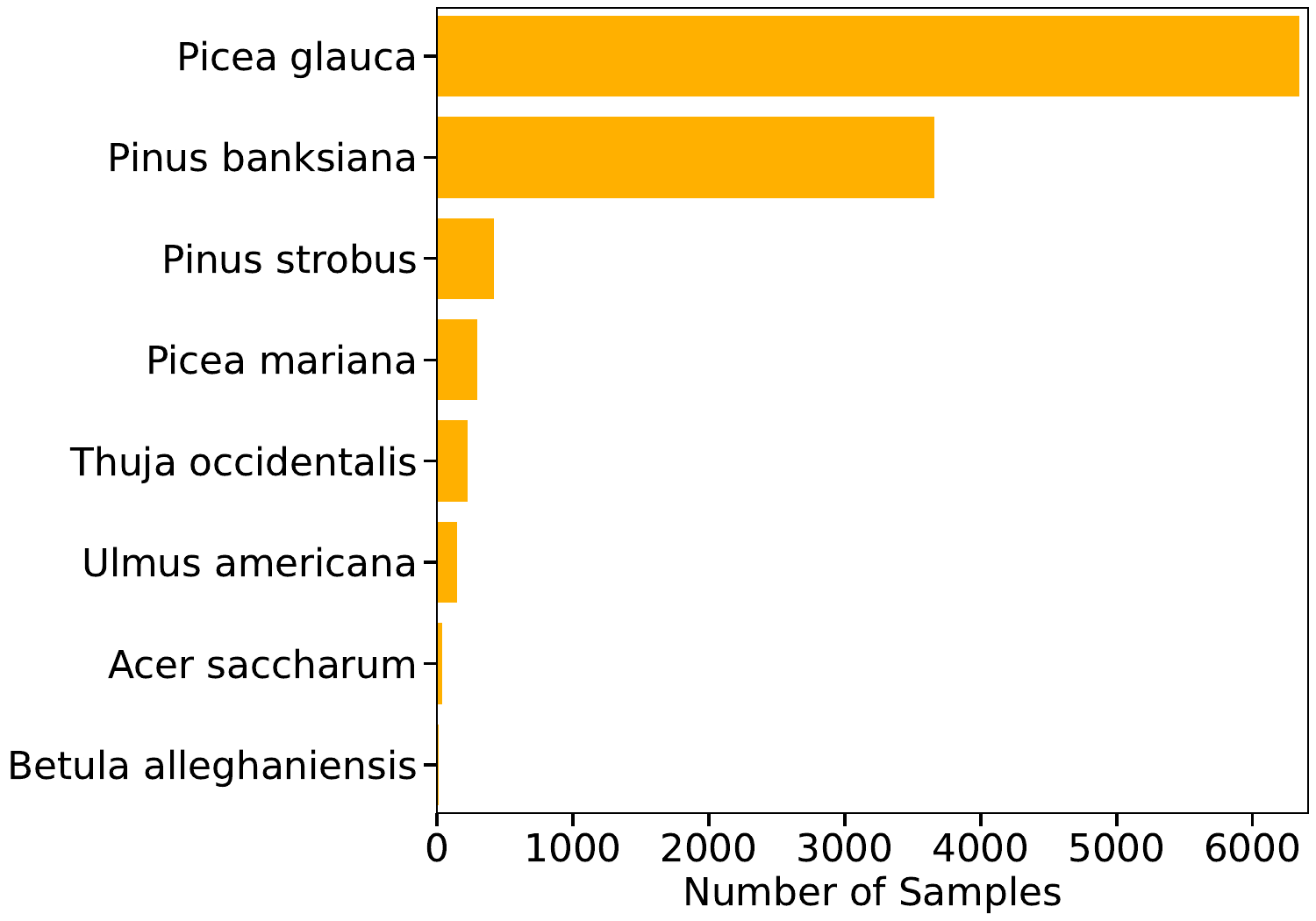}
        \caption{Training split}
        \label{fig:class_dis_qp_train}
    \end{subfigure}
    \hfill
    \begin{subfigure}[b]{0.325\linewidth}
        \centering
        \includegraphics[width=\linewidth]{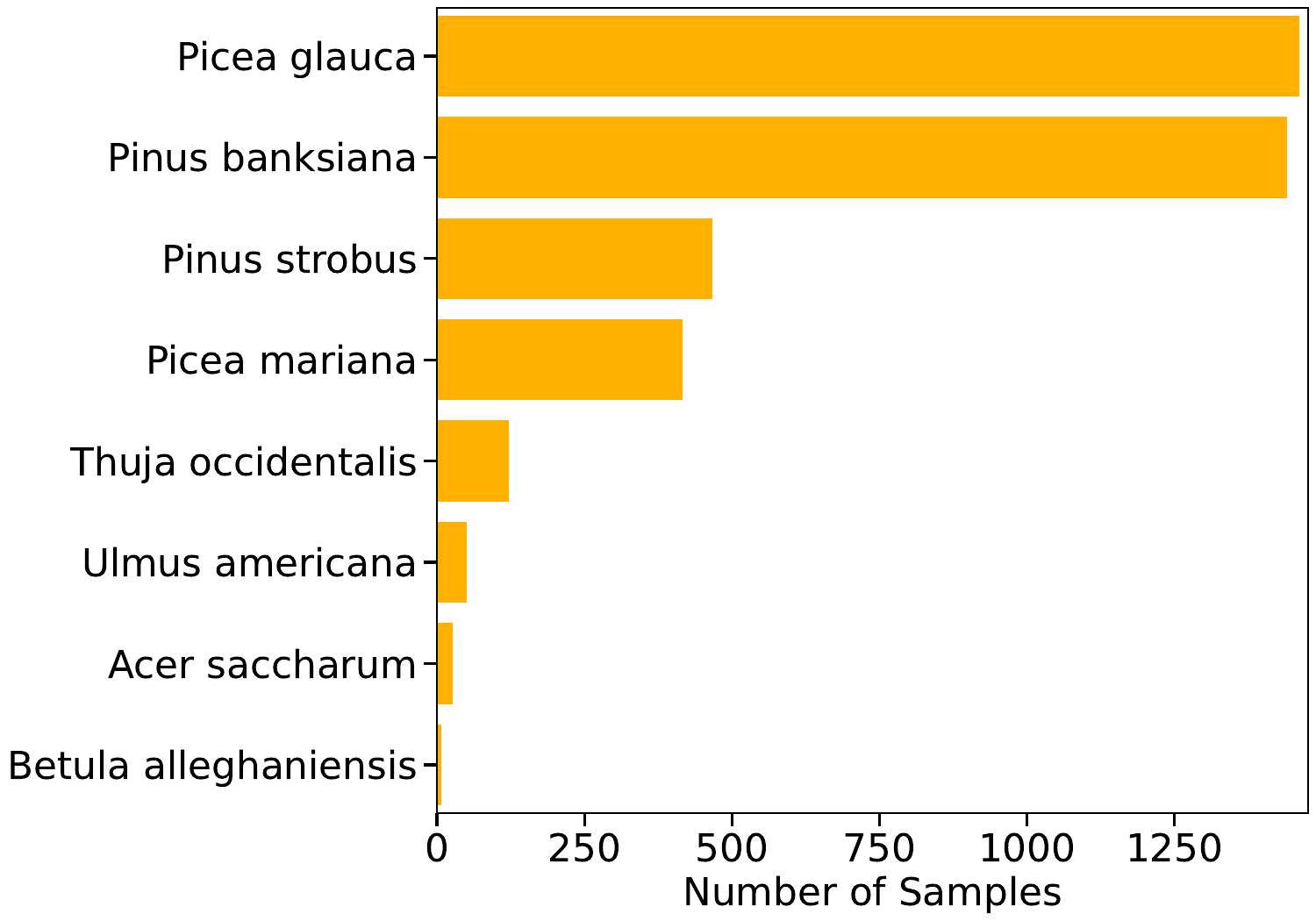}
        \caption{Validation split}
        \label{fig:class_dis_qp_val}
    \end{subfigure}
    \hfill
    \begin{subfigure}[b]{0.325\linewidth}
        \centering
        \includegraphics[width=\linewidth]{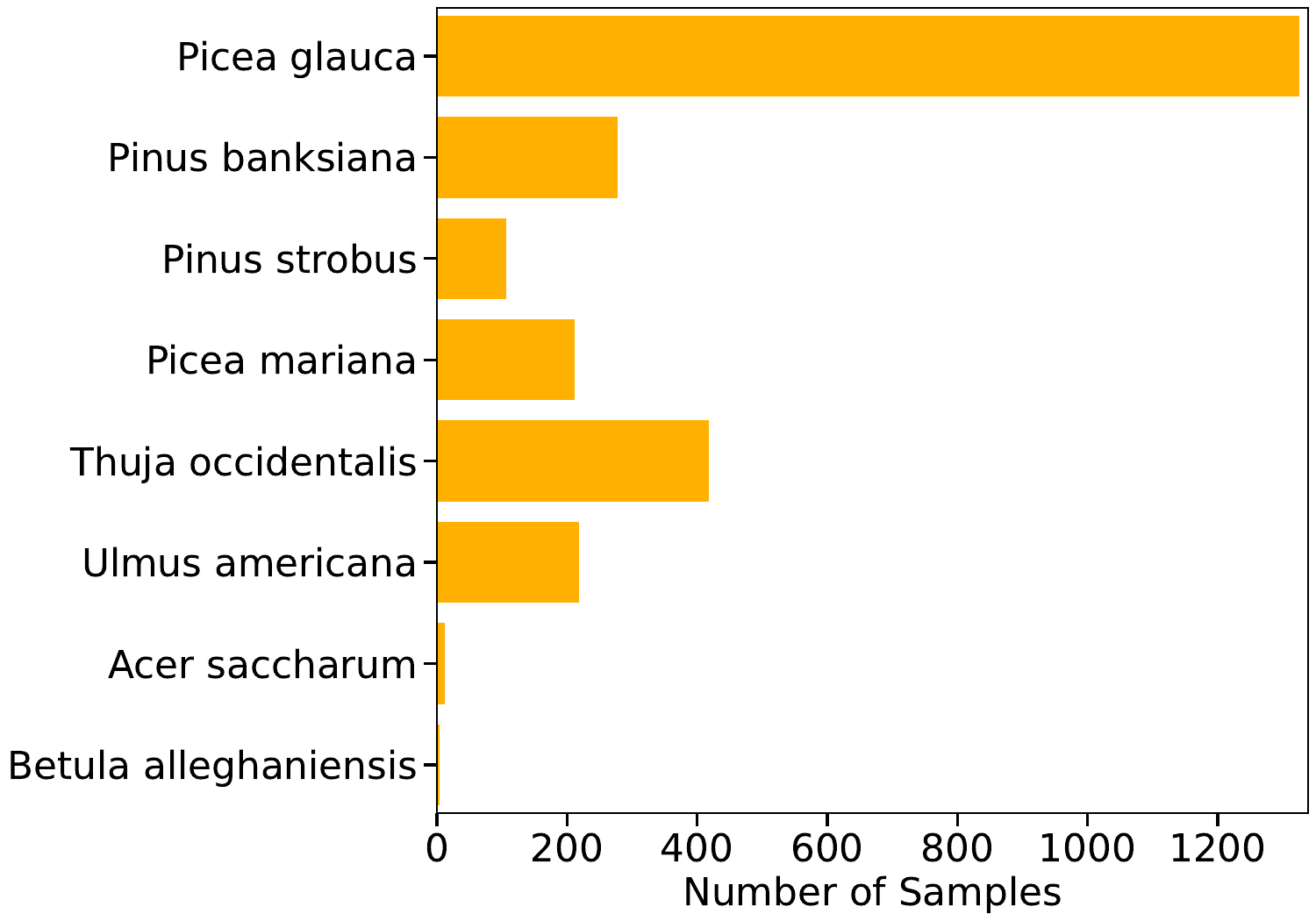}
        \caption{Test split}
        \label{fig:class_dis_qp_test}
    \end{subfigure}
    \caption{
    \textbf{Class distributions across splits for the Quebec Plantations dataset.}
    Separate histograms illustrate the number of samples allocated to each class within the (a) training, (b) validation, and (c) test splits.
    }
    \label{fig:class_dis_qp_splits}
\end{figure}

\label[appendix]{app:class_distribution}
As shown in \cref{fig:class_dis_qt_splits}, the class distribution of the Quebec Trees training split is heavily imbalanced, dominated by the species \textit{Betula papyrifera} and \textit{Acer rubrum}.
Conversely, the species \textit{Tsuga canadensis} is represented by only 9 samples (see \cref{tab:stats_class_qt}).
For the BCI dataset, the number of training samples is constrained to fewer than 250 per class (see \cref{fig:class_dis_bci_splits}).
As detailed in \cref{tab:stats_class_bci}, four families (`Putranjivaceae', `Lecythidaceae', `Phyllanthaceae', and `Nyctaginaceae') contain fewer than 20 training images each.
Finally, \cref{fig:class_dis_qp_splits} reveals that the Quebec Plantations training split is heavily dominated by \textit{Picea glauca} and \textit{Pinus banksiana}.
In contrast, \textit{Acer saccharum} and \textit{Betula alleghaniensis} are represented by just 39 and 11 training samples (see \cref{tab:stats_class_qp}), corresponding to only $0.43\%$ and $0.12\%$ of the training split, respectively.
\begin{table}[!b]
    \caption{
    \textbf{Sample count per class for the Quebec Trees dataset.}
    The table details the number of instances allocated to the training, validation, and test splits, alongside the overall total for each class.
    }
    \label{tab:stats_class_qt}
    \centering
    \setlength{\tabcolsep}{3pt}
    \scriptsize
    \sisetup{table-number-alignment=center}
    \renewcommand{\arraystretch}{1.1}
    \begin{tabular}{
        l
        S[table-format=4.0]
        S[table-format=4.0]
        S[table-format=4.0]
        S[table-format=4.0]
        }
        \toprule
        \text{Class} & \text{Training} & \text{Validation} & \text{Test} & \text{Overall} \\
        \midrule
        Betula papyrifera & 4032 & 762 & 1100 & 5894 \\
        Acer rubrum & 3755 & 809 & 1293 & 5857 \\
        Abies balsamea & 1164 & 675 & 1056 & 2895 \\
        Thuja occidentalis & 712 & 349 & 449 & 1510 \\
        Populus & 845 & 185 & 86 & 1116 \\
        Picea & 317 & 249 & 456 & 1022 \\
        Acer saccharum & 774 & 75 & 165 & 1014 \\
        dead & 550 & 147 & 181 & 878 \\
        Acer pensylvanicum & 495 & 18 & 238 & 751 \\
        Pinus strobus & 407 & 136 & 26 & 569 \\
        Betula alleghaniensis & 57 & 139 & 94 & 290 \\
        Fagus grandifolia & 77 & 23 & 122 & 222 \\
        Larix laricina & 120 & 1 & 64 & 185 \\
        Tsuga canadensis & 9 & 27 & 23 & 59 \\
        \bottomrule
    \end{tabular}
\end{table}

\begin{table}[!tb]
    \caption{
    \textbf{Sample count per class for the BCI dataset.}
    The table details the number of instances allocated to the training, validation, and test splits, alongside the overall total for each class.
    }
    \label{tab:stats_class_bci}
    \centering
    \setlength{\tabcolsep}{3pt}
    \scriptsize
    \sisetup{table-number-alignment=center}
    \renewcommand{\arraystretch}{1.1}
    \begin{tabular}{
        l
        S[table-format=4.0]
        S[table-format=4.0]
        S[table-format=4.0]
        S[table-format=4.0]
        }
        \toprule
        \text{Class} & \text{Training} & \text{Validation} & \text{Test} & \text{Overall} \\
        \midrule
        Malvaceae & 241 & 48 & 67 & 356 \\
        Fabaceae & 218 & 17 & 19 & 254 \\
        Bignoniaceae & 116 & 27 & 12 & 155 \\
        Rubiaceae & 90 & 22 & 34 & 146 \\
        Lauraceae & 64 & 18 & 40 & 122 \\
        Urticaceae & 88 & 4 & 13 & 105 \\
        Burseraceae & 60 & 19 & 13 & 92 \\
        Sapotaceae & 65 & 10 & 13 & 88 \\
        Meliaceae & 53 & 13 & 22 & 88 \\
        Euphorbiaceae & 42 & 22 & 17 & 81 \\
        Apocynaceae & 61 & 10 & 9 & 80 \\
        Myristicaceae & 44 & 10 & 22 & 76 \\
        Rutaceae & 60 & 8 & 4 & 72 \\
        Moraceae & 37 & 14 & 19 & 70 \\
        Arecaceae & 52 & 4 & 5 & 61 \\
        Anacardiaceae & 37 & 6 & 12 & 55 \\
        Cordiaceae & 44 & 7 & 4 & 55 \\
        Nyctaginaceae & 18 & 3 & 8 & 29 \\
        Phyllanthaceae & 16 & 5 & 2 & 23 \\
        Lecythidaceae & 10 & 2 & 8 & 20 \\
        Putranjivaceae & 9 & 4 & 2 & 15 \\
        \bottomrule
    \end{tabular}
\end{table}

\clearpage

\begin{table}[h]
    \caption{
    \textbf{Sample count per class for the Quebec Plantations dataset.}
    The table details the number of instances allocated to the training, validation, and test splits, alongside the overall total for each class.
    }
    \label{tab:stats_class_qp}
    \centering
    \setlength{\tabcolsep}{3pt}
    \scriptsize
    \sisetup{table-number-alignment=center}
    \renewcommand{\arraystretch}{1.1}
    \begin{tabular}{
        l
        S[table-format=4.0]
        S[table-format=4.0]
        S[table-format=4.0]
        S[table-format=4.0]
        }
        \toprule
        \text{Class} & \text{Training} & \text{Validation} & \text{Test} & \text{Overall} \\
        \midrule
        Picea glauca & 6346 & 1462 & 1326 & 9134 \\
        Pinus banksiana & 3661 & 1440 & 278 & 5379 \\
        Pinus strobus & 418 & 467 & 107 & 992 \\
        Picea mariana & 297 & 416 & 212 & 925 \\
        Thuja occidentalis & 223 & 122 & 418 & 763 \\
        Ulmus americana & 150 & 51 & 218 & 419 \\
        Acer saccharum & 39 & 27 & 12 & 78 \\
        Betula alleghaniensis & 11 & 8 & 4 & 23 \\
        \bottomrule
    \end{tabular}
\end{table}

\clearpage

\subsection{Height Statistics}
\label[appendix]{app:height_stats}

\begin{figure}[b!]
    \centering
    \begin{subfigure}[b]{0.325\linewidth}
        \centering
        \includegraphics[width=\linewidth]{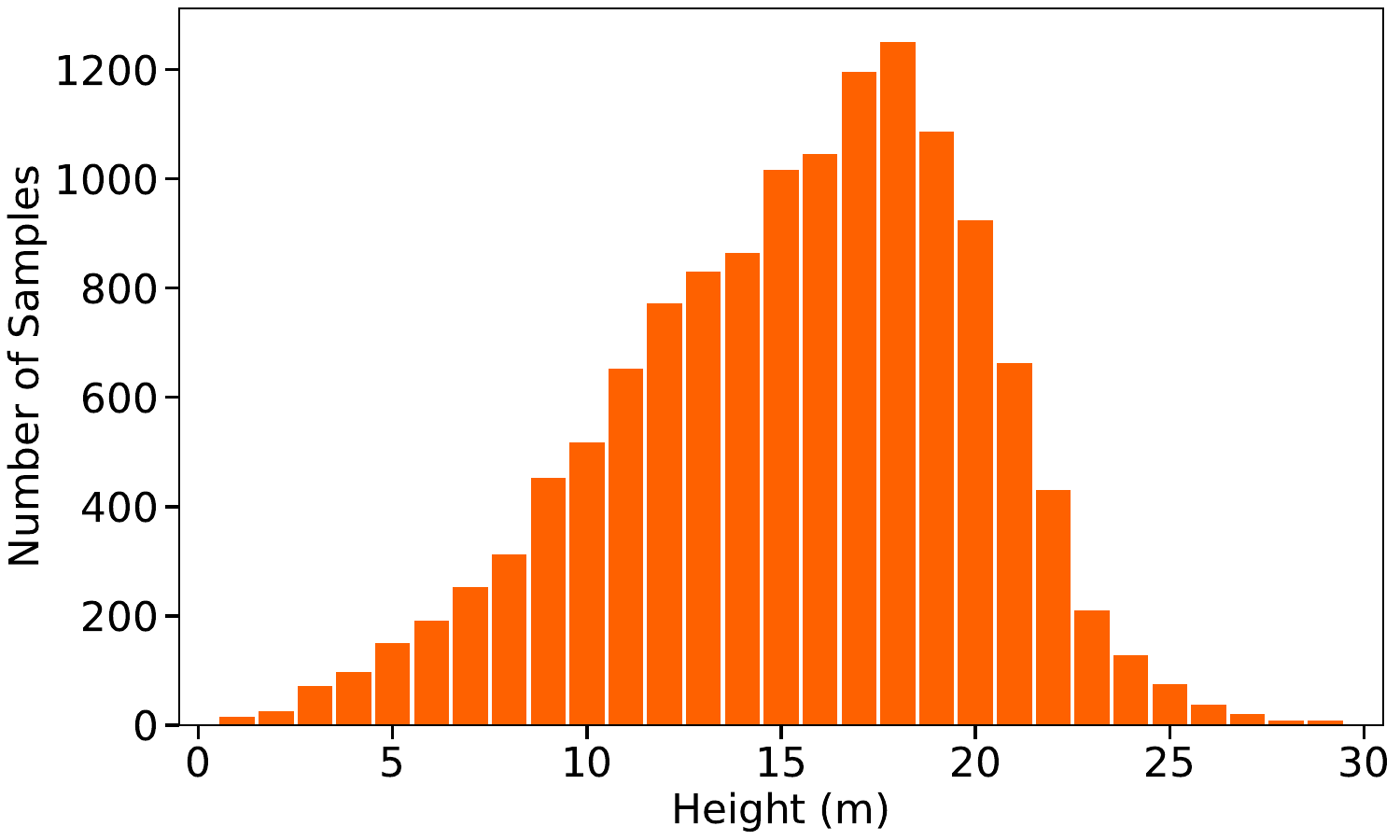}
        \caption{Training split}
        \label{fig:height_dis_qt_train}
    \end{subfigure}
    \hfill
    \begin{subfigure}[b]{0.325\linewidth}
        \centering
        \includegraphics[width=\linewidth]{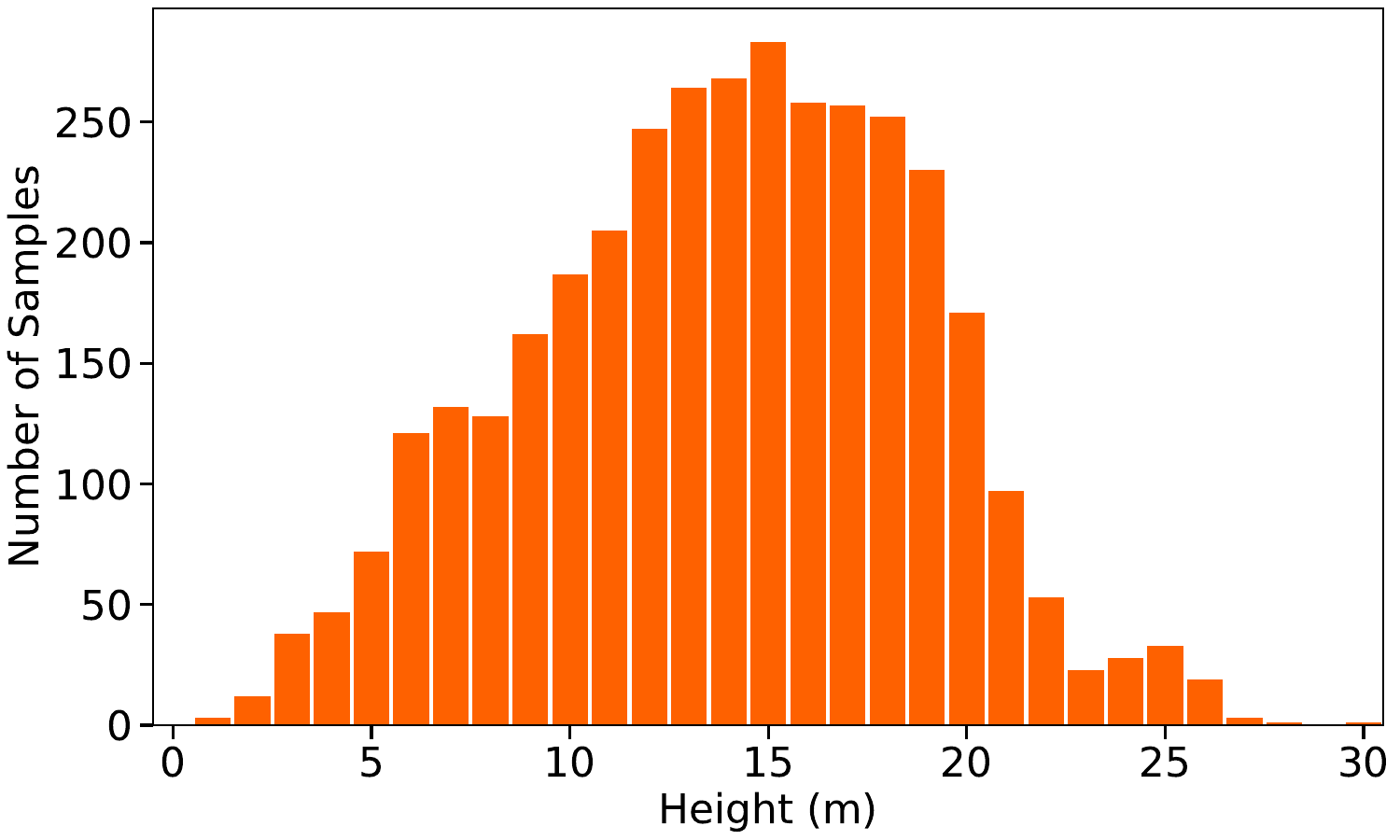}
        \caption{Validation split}
        \label{fig:height_dis_qt_val}
    \end{subfigure}
    \hfill
    \begin{subfigure}[b]{0.325\linewidth}
        \centering
        \includegraphics[width=\linewidth]{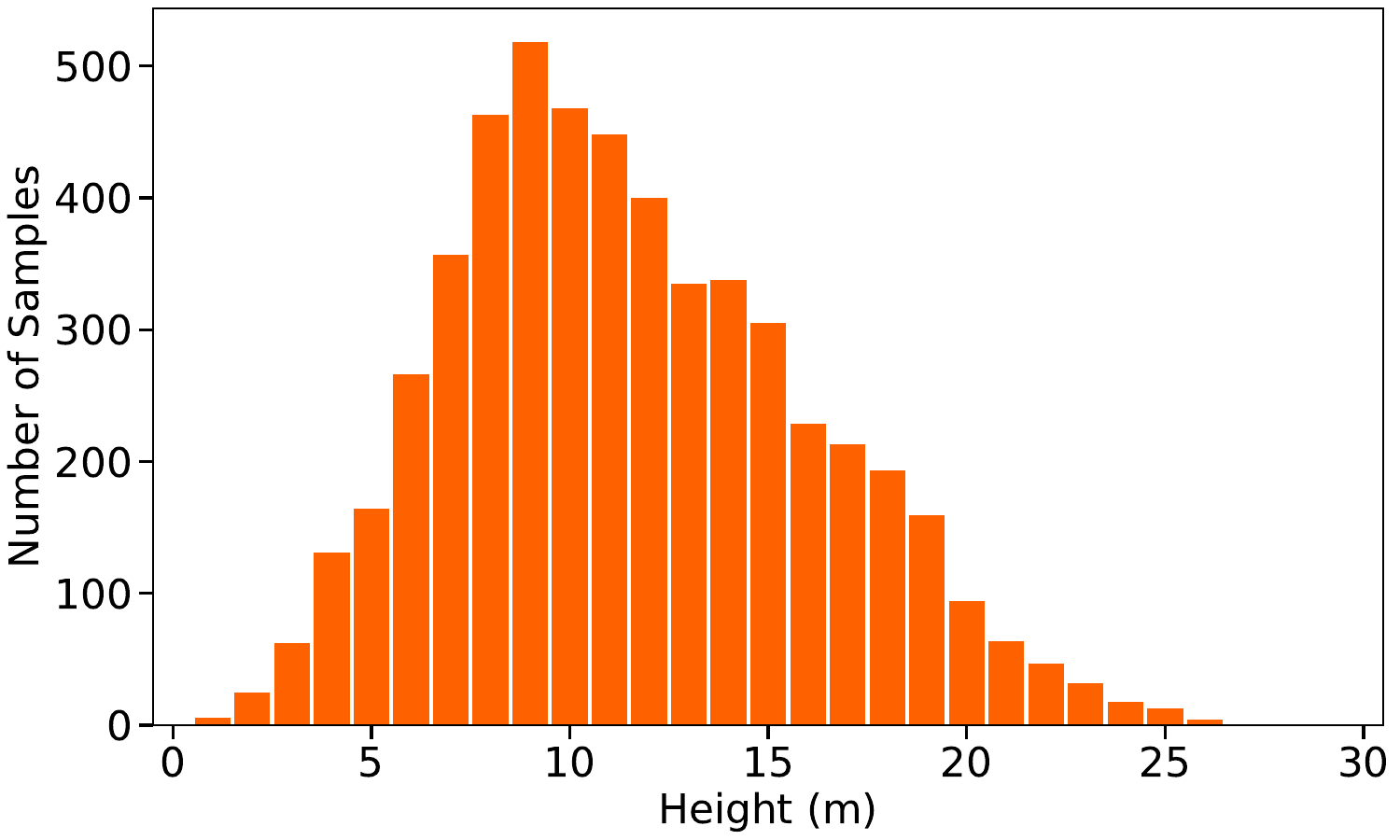}
        \caption{Test split}
        \label{fig:height_dis_qt_test}
    \end{subfigure}
    \caption{
    \textbf{Height distributions across splits for the Quebec Trees dataset.}
    Separate histograms illustrate the number of samples binned into \SI{1}{m} height intervals within the (a) training, (b) validation, and (c) test splits.
    }
    \label{fig:height_dis_qt_splits}
\end{figure}

\begin{figure}[!b]
    \centering
    \begin{subfigure}[b]{0.325\linewidth}
        \centering
        \includegraphics[width=\linewidth]{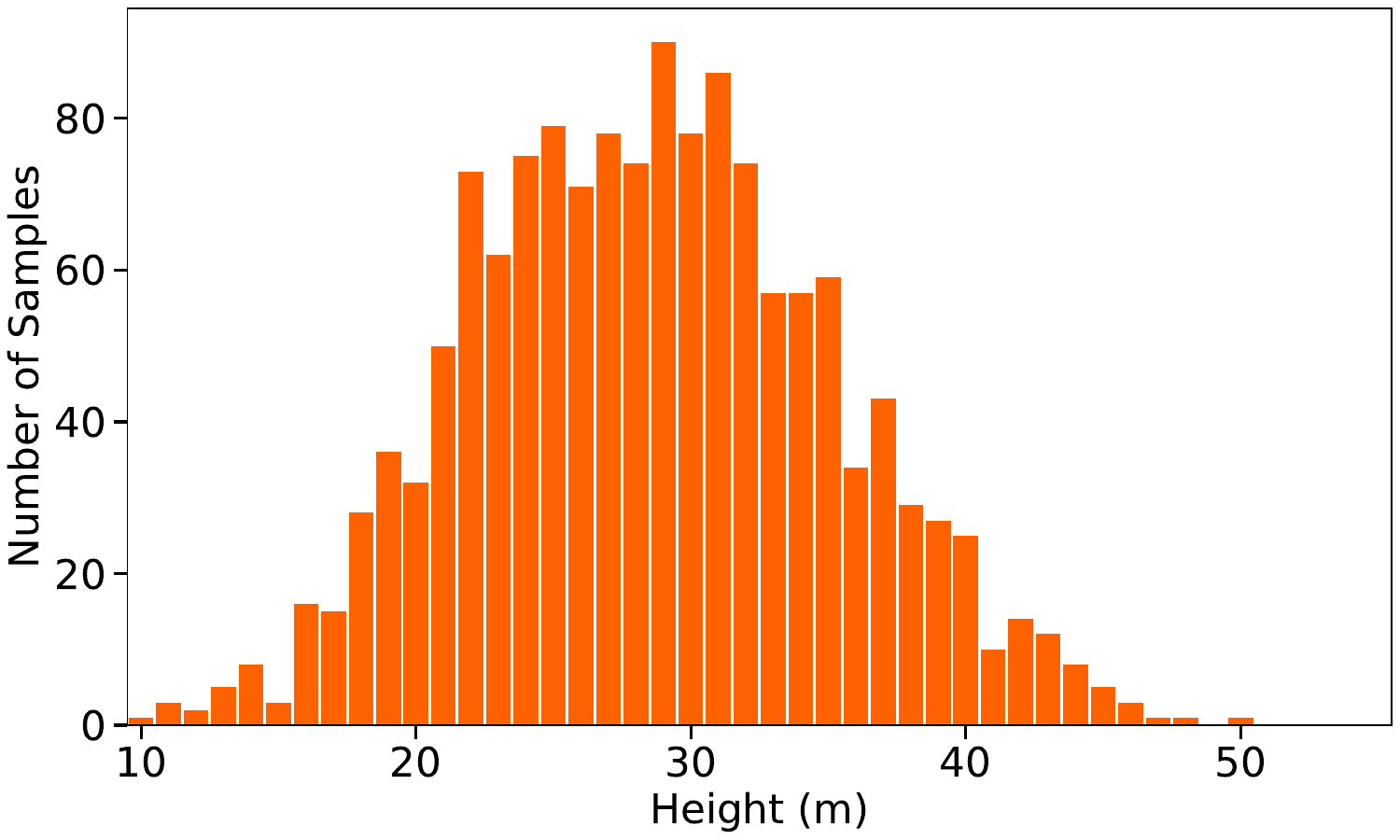}
        \caption{Training split}
        \label{fig:height_dis_bci_train}
    \end{subfigure}
    \hfill
    \begin{subfigure}[b]{0.325\linewidth}
        \centering
        \includegraphics[width=\linewidth]{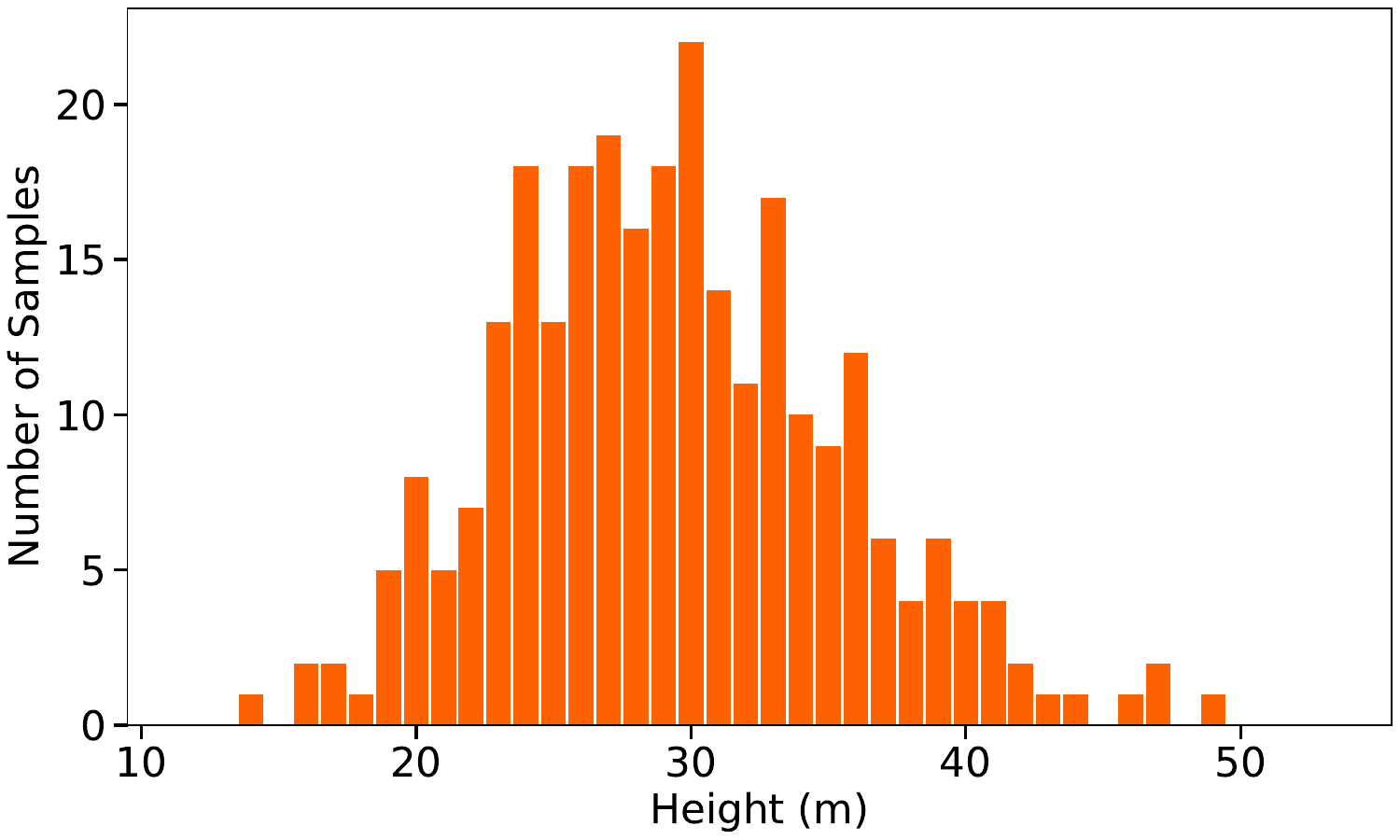}
        \caption{Validation split}
        \label{fig:height_dis_bci_val}
    \end{subfigure}
    \hfill
    \begin{subfigure}[b]{0.325\linewidth}
        \centering
        \includegraphics[width=\linewidth]{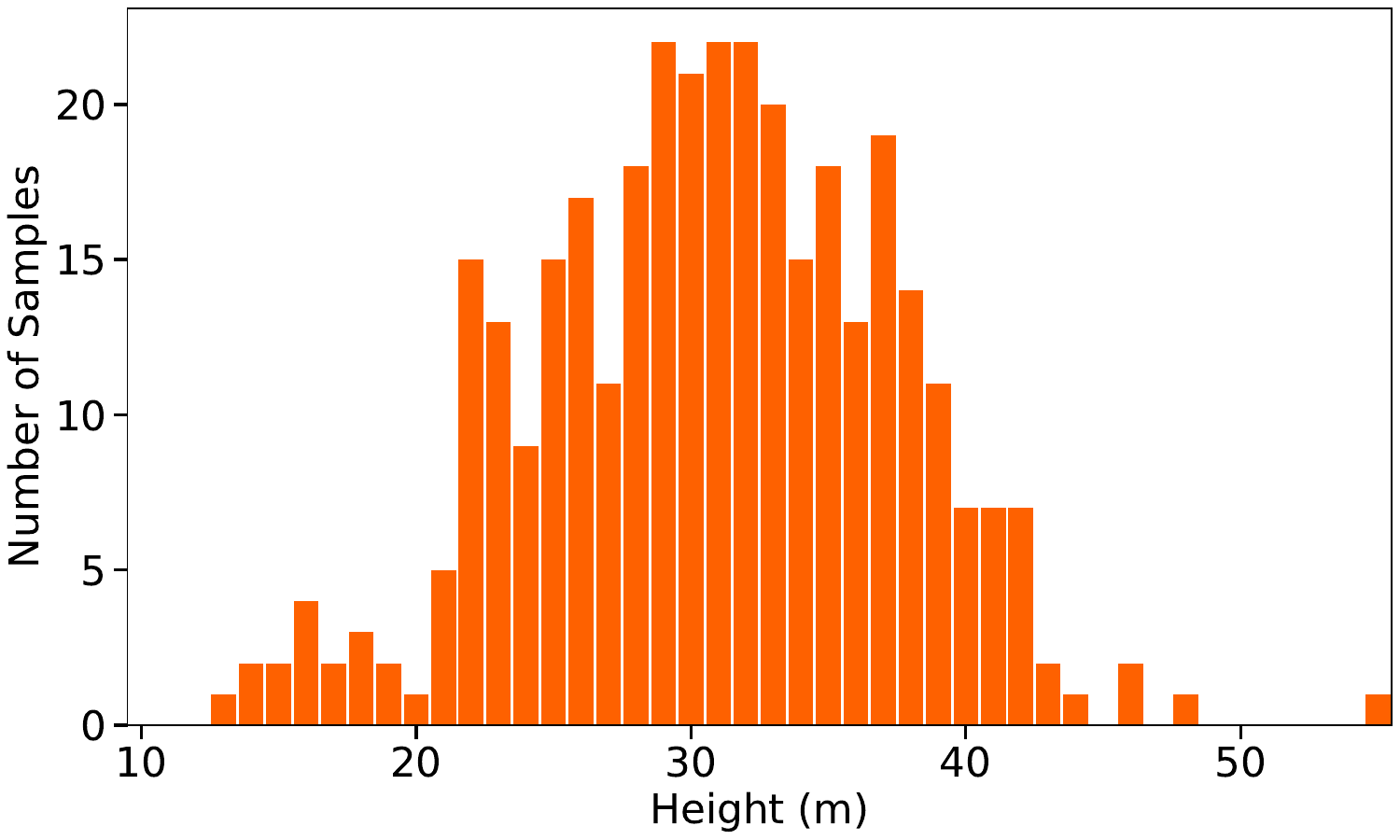}
        \caption{Test split}
        \label{fig:height_dis_bci_test}
    \end{subfigure}
    \caption{
    \textbf{Height distributions across splits for the BCI dataset.}
    Separate histograms illustrate the number of samples binned into \SI{1}{m} height intervals within the (a) training, (b) validation, and (c) test splits.
    }
    \label{fig:height_dis_bci_splits}
\end{figure}

\begin{figure}[!tb]
    \centering
    \begin{subfigure}[b]{0.325\linewidth}
        \centering
        \includegraphics[width=\linewidth]{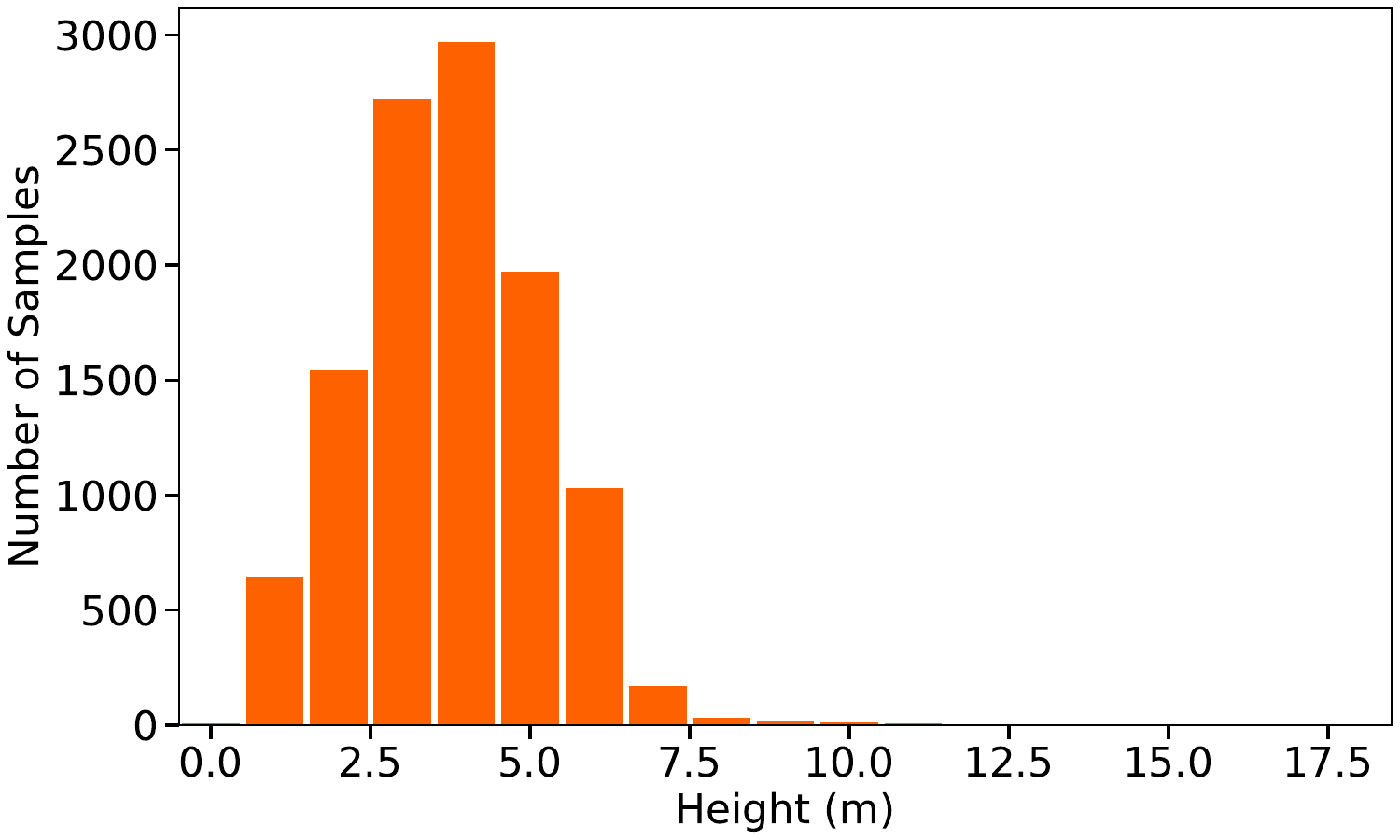}
        \caption{Training split}
        \label{fig:height_dis_qp_train}
    \end{subfigure}
    \hfill
    \begin{subfigure}[b]{0.325\linewidth}
        \centering
        \includegraphics[width=\linewidth]{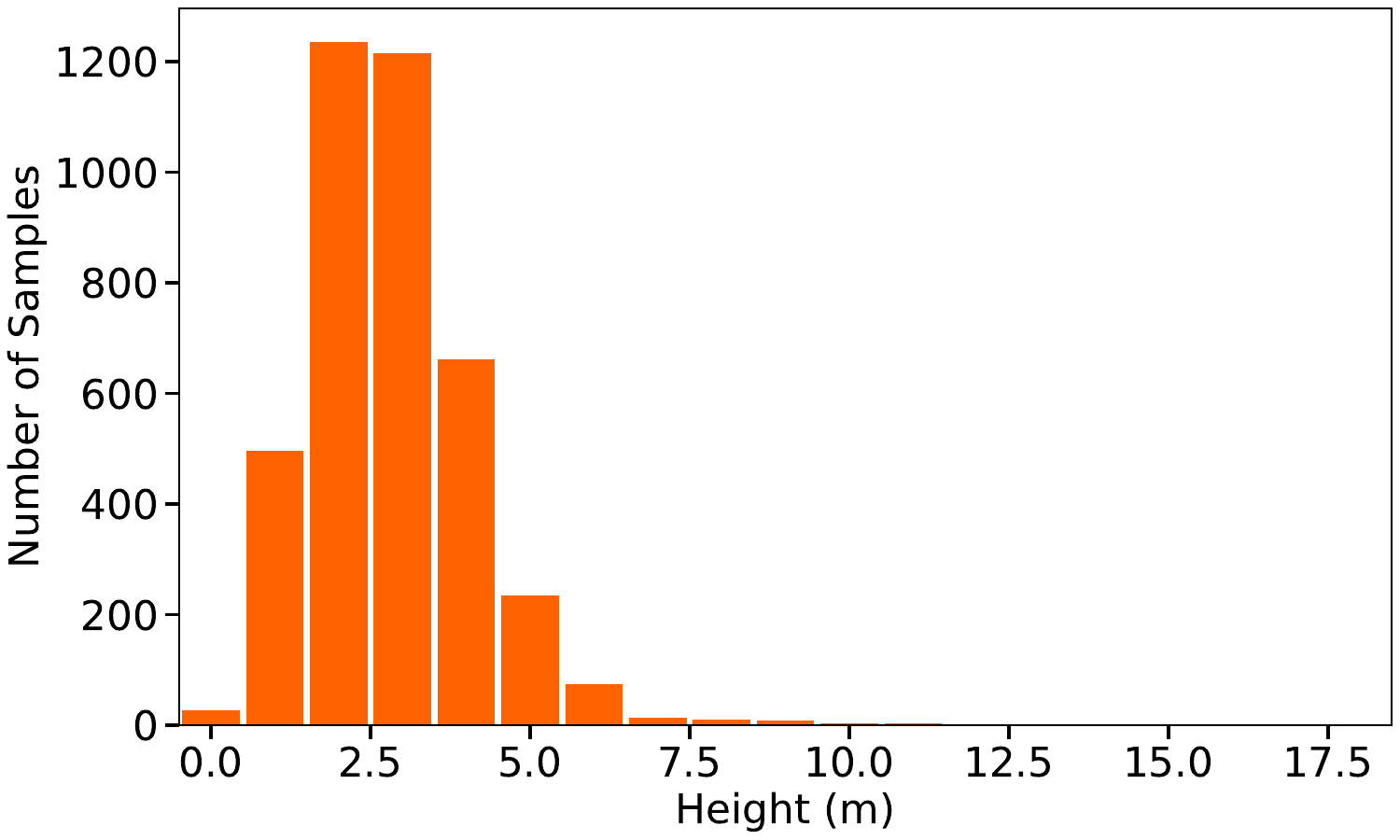}
        \caption{Validation split}
        \label{fig:height_dis_qp_val}
    \end{subfigure}
    \hfill
    \begin{subfigure}[b]{0.325\linewidth}
        \centering
        \includegraphics[width=\linewidth]{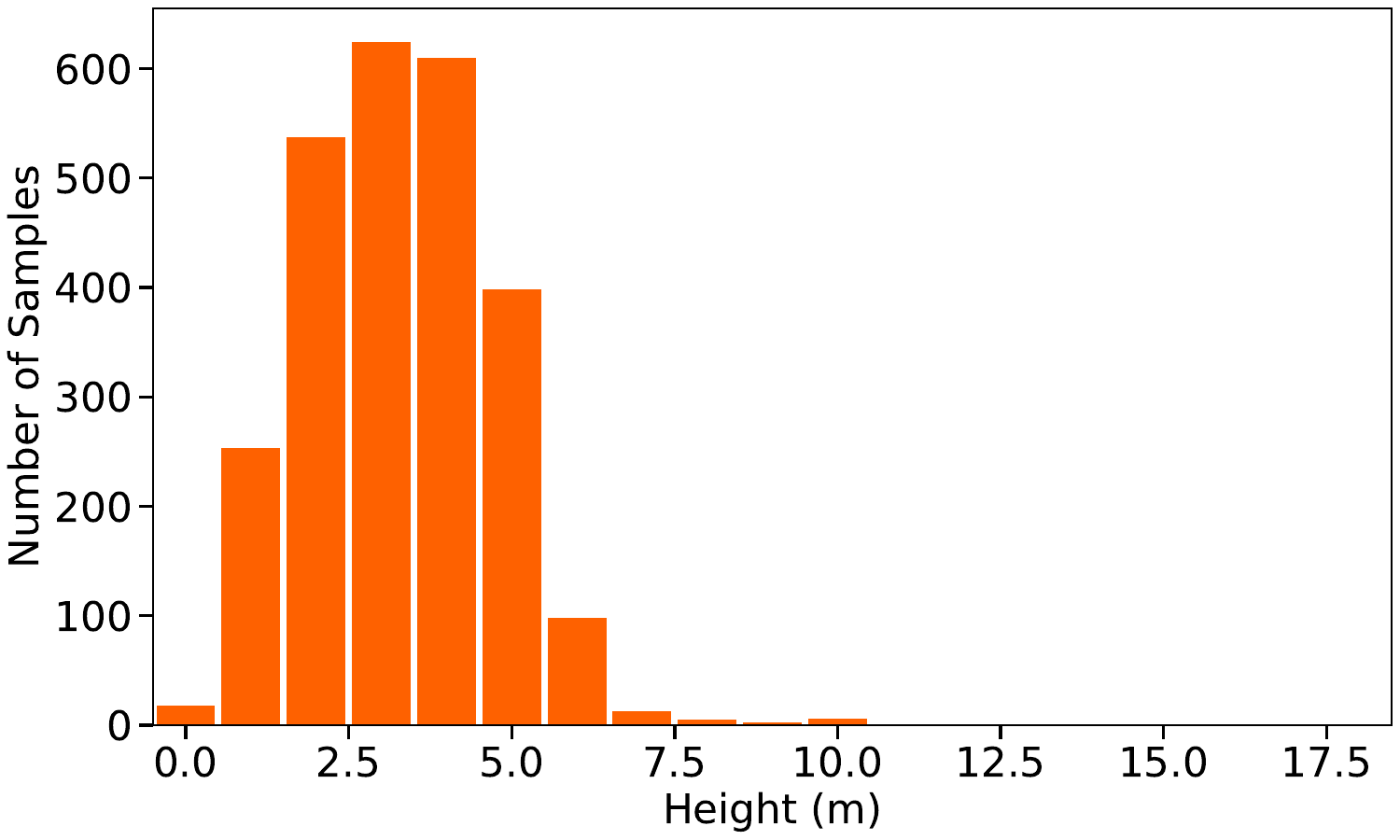}
        \caption{Test split}
        \label{fig:height_dis_qp_test}
    \end{subfigure}
    \caption{
    \textbf{Height distributions across splits for the Quebec Plantations dataset.}
    Separate histograms illustrate the number of samples binned into \SI{1}{m} height intervals within the (a) training, (b) validation, and (c) test splits.
    }
    \label{fig:height_dis_qp_splits}
\end{figure}

\cref{tab:stats_split} presents the height statistics for the three datasets comprising \benchmark.
We observe that trees in tropical forests (BCI) are, on average, more than twice as tall as those in temperate forest (Quebec Trees).
Furthermore, the trees in the boreal plantations of the Quebec Plantations dataset are generally shorter than those in the other two datasets, as the majority has not yet reached maturity (\ie, 10-year old plantations).
As visualized in the height histograms for Quebec Trees (\cref{fig:height_dis_qt_splits}), the modal height bin of the training split (\SI{18}{m}) is significantly higher than that of the test split (\SI{9}{m}).
For the BCI dataset, the height distributions remain relatively consistent across splits (\cref{fig:height_dis_bci_splits}).
However, the test split includes a single outlier measuring \SI{55}{m} in height;
the presence of anomalies facilitates a robust evaluation of height estimation methods under realistic conditions.
Finally, \cref{fig:height_dis_qp_splits} demonstrates that the majority of trees in Quebec Plantations range between \SI{1}{m} and \SI{7}{m}, with height distributions similar across all splits.
\begin{table}[h]
    \caption{
    \textbf{Height statistics across datasets and splits.}
    The table presents the mean, standard deviation, minimum, and maximum height, along with the total number of samples for the training, validation, and test splits, as well as the overall totals for the Quebec Trees, BCI, and Quebec Plantations datasets.
    }
    \label{tab:stats_split}
    \centering
    \setlength{\tabcolsep}{3pt}
    \scriptsize
    \sisetup{table-number-alignment=center, separate-uncertainty=true, retain-zero-uncertainty=true}
    \renewcommand{\arraystretch}{1.1}
    \begin{tabular}{
        l
        c
        S[table-format=2.2]
        S[table-format=1.2]
        S[table-format=2.2]
        S[table-format=2.2]
        S[table-format=2.1]
        }
        \toprule
        \text{Dataset} & \text{Split} & \text{Mean (m)} & \text{Std (m)} & \text{Min (m)} & \text{Max (m)} & \text{Number (K)} \\
        \midrule
        \multirow{3}{*}{Quebec Trees} & Training & 15.37 & 4.63 & 0.16 & 29.45 & 13.3 \\
        & Validation & 13.92 & 4.90 & 0.82 & 29.94 & 3.6 \\
        & Test & 11.54 & 4.53 & 0.83 & 27.77 & 5.4 \\
        & Overall & 14.22 & 4.91 & 0.16 & 29.94 & 22.3 \\
        \midrule
        \multirow{3}{*}{BCI} & Training & 28.58 & 6.60 & 9.69 & 50.17 & 1.4 \\
        & Validation & 29.34 & 6.13 & 14.23 & 49.43 & 0.3 \\
        & Test & 30.80 & 6.55 & 12.58 & 54.61 & 0.3 \\
        & Overall & 29.05 & 6.59 & 9.69 & 54.61 & 2.0 \\
        \midrule
        \multirow{3}{*}{Quebec Plantations} & Training & 3.75 & 1.44 & 0.04 & 18.38 & 11.1 \\
        & Validation & 2.83 & 1.34 & 0.00 & 16.60 & 4.0 \\
        & Test & 3.30 & 1.46 & 0.01 & 15.68 & 2.6 \\
        & Overall & 3.48 & 1.47 & 0.00 & 18.38 & 17.7 \\
        \bottomrule
    \end{tabular}
\end{table}

\clearpage
\clearpage

\section{Evaluation Metrics}
\label[appendix]{app:eval}
This section provide additional details on the evaluation metrics of \cref{subsec:metrics}.
Because accurate predictions across all classes are critical in forest monitoring, we macro-average our classification metrics.
Similar to \cref{subsec:metrics}, we assume a dataset of $N$ images.
Let $\hat{h}_i$ and $\hat{c}_i$ denote the predicted height and class for the $i$-th image, respectively, while $h_i$ and $c_i$ represent the corresponding ground truth.

\paragraph{Mean Absolute Error (MAE).}
The \gls{mae} measures the average magnitude of the absolute errors between the predicted and actual tree heights.
It is defined as:
\begin{equation}
    \text{MAE} = \frac{1}{N} \sum_{i=1}^{N} \left|\hat{h}_i - h_i\right|.
\end{equation}

\paragraph{Root Mean Squared Error (RMSE).}
In contrast to \gls{mae}, the \gls{rmse} disproportionately penalizes large errors because the individual differences are squared before being averaged.
The final square root operation then converts the metric back to the original unit (meters).
\gls{rmse} serves as a standard evaluation metric in tree height estimation \cite{hao2021automated, lang2023high, lefebvre2025suivi}.
We define the \gls{rmse} as follows:
\begin{equation}
    \text{RMSE} = \sqrt{\frac{1}{N} \sum_{i=1}^{N} \left(h_i - \hat{h}_i\right) ^ 2}.
\end{equation}

\paragraph{F1-Score (F1).}
The F1-Score is the harmonic mean of precision and recall.
Given the imbalanced nature of the \benchmark datasets, this metric provides a more robust evaluation than standard accuracy.
Let $\text{TP}_k$, $\text{FP}_k$, and $\text{FN}_k$ denote the true positives, false positives, and false negatives for class $k \in \{1, \dots, C\}$, computed using the indicator function $\mathbbm{1}$ as follows:
\begin{align}
    \text{TP}_k &= \sum_{i=1}^{N} \mathbbm{1}_{[c_i = k]} \cdot \mathbbm{1}_{[\hat{c}_i = k]}, \\
    \text{FP}_k &= \sum_{i=1}^{N} \mathbbm{1}_{[c_i \neq k]} \cdot \mathbbm{1}_{[\hat{c}_i = k]}, \\
    \text{FN}_k &= \sum_{i=1}^{N} \mathbbm{1}_{[c_i = k]} \cdot \mathbbm{1}_{[\hat{c}_i \neq k]}.
\end{align}
The class-specific F1-Score, $\text{F1}_k$, is then calculated as:
\begin{equation}
    \text{F1}_k = \frac{2 \text{TP}_k}{2 \text{TP}_k + \text{FP}_k + \text{FN}_k}.
\end{equation}
To obtain the macro-averaged F1-Score, we compute the unweighted mean of the class-specific scores for all classes $k \in \{1, \dots, C\}$:
\begin{equation}
    \text{F1} = \frac{1}{C} \sum_{k=1}^{C} \text{F1}_k.
\end{equation}

\paragraph{Accuracy (Acc).}
The accuracy for a class $k$ is the proportion of correct predictions where $c_i = k$.
It is defined as:
\begin{equation}
    \text{Acc}_k = \frac{1}{N_k} \sum_{i=1}^{N} \mathbbm{1}_{[c_i = k]} \cdot \mathbbm{1}_{[\hat{c}_i = k]},
\end{equation}
where the number of samples with class $k$, $N_k$, can be expressed as:
\begin{equation}
    N_k = \sum_{i=1}^{N} \mathbbm{1}_{[c_i = k]}.
\end{equation}
The overall macro-averaged accuracy is the average of the class-specific accuracies:
\begin{equation}
    \text{Acc} = \frac{1}{C} \sum_{k=1}^{C} \text{Acc}_k.
\end{equation}

\section{Implementation Details}
\label[appendix]{app:implementation_details}

\paragraph{Benchmark.}
We implement the data extraction pipeline (\cref{subsec:extraction}) in Python using GeoDataset~\cite{baudchon2024geodataset}.
For the Quebec Plantations dataset, following Lefebvre~\etal~\cite{lefebvre2025suivi}, we use the maximum value instead of the 99th percentile due to the higher accuracy of the \gls{chm} for this dataset.
We also apply their proposed linear correction to align the extracted heights with field measurements.
When creating height labels for the Quebec Plantations dataset, a small number of highly non-circular tree crown shapes yielded empty buffer regions ($S_{\text{buf}}=\emptyset$).
To address this, we reduced the scaling factor in \cref{eq:buffer} from $0.1$ to $0.05$. 
Rarely, the linear correction proposed by Lefebvre~\etal~\cite{lefebvre2025suivi} produced negative heights; these samples are marked as undefined and excluded from height training and evaluation.

\paragraph{DINOvTree architecture.}
We use the ViT-B/16 variant of DINOv3 with embedding dimension $768$ and patch size $16$, if not otherwise specified.
The \glspl{mlp} operating on patch tokens follow the standard transformer block architecture~\cite{vaswani2017attention} with pre-norm residual connections.
Patch tokens are first normalized using LayerNorm~\cite{ba2016layer} and processed using two fully-connected layers (expansion ratio of $4$, GELU activation).
Subsequently, the processed patch tokens are residually added to the original tokens.
The resulting sum is normalized by a LayerNorm followed by the cross-attention module.
The cross-attention employs 8 heads with query, key, and value dimensions set to the backbone embedding size of $768$. 
A final LayerNorm is applied directly before the output linear projection.

\paragraph{Training setup.}
We implement our model in PyTorch~\cite{paszke2019pytorch} based on the DINOv3~\cite{simeoni2025dinov3} codebase.
During training, we apply dihedral group $D_4$ augmentations (rotations and reflections) to $512 \times 512$ input images.
Models are trained for $50$ epochs with a batch size of $64$ using AdamW optimizer~\cite{loshchilovdecoupled}.
For BCI, we extend training to $200$ epochs due to limited data.
We use cosine learning rate scheduler~\cite{loshchilov2017sgdr} with a linear warmup, starting at $2e^{-8}$, peaking at $5e^{-5}$, and decaying back to a minimum learning rate of $2e^{-7}$.
To balance pre-trained feature preservation with task-specific adaptation, we apply a $0.5\times$ learning rate multiplier to the backbone. 
All experiments run on a single NVIDIA H100 GPU (80GB). 
We select the checkpoint maximizing the average of macro F1-score and $\delta_{1.25}$ on the validation set.

\paragraph{Mask R-CNN.}
Due to slower convergence, we train this competing method for $100$ epochs. 
To adapt the method to \benchmark, we supervise only the center tree during training, using its instance mask, bounding box, and class label.
At inference, when multiple objects are detected, we select the instance whose centroid is closest to the image center. 
When no object is detected, we assign a random class for classification or the central height bin for regression.

\noindent For all competing methods, we use identical training hyperparameters but train separate models for height estimation and classification.
When memory constraints necessitate smaller batches, we employ gradient accumulation to maintain an effective batch size of $64$.
All competing models are initialized with pre-trained weights and fine-tuned on each dataset.

\section{Additional Experiments}
\label[appendix]{app:additional_experiments}

\subsection{Analysis of Different Classification Loss Functions}
\label[appendix]{app:loss}
We train the classification head of \method with the standard Cross-Entropy (CE) loss (\cref{subsec:loss}), despite the existence of specialized losses for long-tailed distributions.
For instance, the Focal Loss (FL) \cite{lin2017focal} reduces the weight of well-classified samples to force the model to focus on hard examples, while the Class-Balanced (CB) loss \cite{cui2019class} reweights the loss based on the effective number of samples per class.
\begin{table*}[t]
    \caption{
    \textbf{Classification loss analysis of \method-B on the Quebec Trees dataset.}
    We evaluate the default Cross-Entropy (CE) classification loss alongside Focal Loss (FL), both individually and combined with Class-Balanced (CB) reweighting.
    We report the mean ± standard error over 5 seeds.
    \textbf{Bold} and \underline{underline} denote best and second-best.
    Gray corresponds to our design choice.
    }
    \label{tab:cls_loss}
    \centering
    \setlength{\tabcolsep}{3pt}
    \scriptsize
    \sisetup{table-number-alignment=center, separate-uncertainty=true, retain-zero-uncertainty=true}
    \renewcommand{\arraystretch}{1.1}
        \begin{tabular}{
        l
        S[table-format=2.2(2), separate-uncertainty]
        S[table-format=1.2(2), separate-uncertainty]
        S[table-format=1.2(2), separate-uncertainty]
        S[table-format=1.2(2), separate-uncertainty]
        S[table-format=2.2(2), separate-uncertainty]
        S[table-format=2.2(2), separate-uncertainty]
        }
            \toprule
            \textbf{Cls. Loss} & {$\delta_{1.25}$~(\%)$~\uparrow$} & {MSLE~($10^{-2}$)~$\downarrow$} & {MAE~(m)~$\downarrow$} & {RMSE~(m)~$\downarrow$} & {F1~(\%)~$\uparrow$} & {Acc~(\%)~$\uparrow$} \\
            \midrule
            \rowcolor{gray!25} CE & \ulmain{88.29}{0.32} & \bfmainsmall{1.85}{0.05} & \ulmainsmall{1.17}{0.02} & \ulmainsmall{1.54}{0.02} & \ulmain{86.89}{0.31} & \ulmain{85.03}{0.45} \\
            CE, CB & 87.49(43) & 1.97(04) & 1.19(01) & 1.57(01) & \bfmain{87.26}{0.35} & \bfmain{86.43}{0.28} \\
            FL & 86.85(10) & 2.05(02) & 1.22(01) & 1.60(01) & 86.72(49) & 84.26(40) \\
            FL, CB & \bfmain{88.59}{0.12} & \ulmainsmall{1.88}{0.03} & \bfmainsmall{1.14}{0.00} & \bfmainsmall{1.52}{0.01} & 86.19(35) & 84.99(55) \\
            \bottomrule
        \end{tabular}
\end{table*}

As shown in \cref{tab:cls_loss}, the simple CE loss yields the best or second-best results across all metrics.
In contrast, the results for FL are worse than those for CE in every metric.
Interestingly, applying CB reweighting improves the metrics on one task while degrading it on the other, depending on the base loss used.
Specifically, it improves classification when applied to the CE loss, but improves height estimation when applied to FL.

\subsection{Analysis of Different Levels of Parameter Sharing}
\label[appendix]{app:parameter_sharing}
As detailed in \cref{subsec:architecture}, we only share the \gls{vfm} backbone between the height estimation and classification tasks.
However, sharing more parameters reduces the total parameter count.
\begin{table*}[t]
    \caption{
    \textbf{Parameter sharing analysis of \method-B on the Quebec Trees dataset.}
    The table evaluates the impact of sharing specific components---namely the \gls{mlp}, Cross-Attention (CA) module, and task Query (Q)---between both heads.
    We report the mean ± standard error over 5 seeds.
    \textbf{Bold} and \underline{underline} denote best and second-best.
    Gray corresponds to our design choice.
    }
    \label{tab:parameter_sharing_analysis}
    \centering
    \setlength{\tabcolsep}{2pt}
    \scriptsize
    \sisetup{table-number-alignment=center, separate-uncertainty=true, retain-zero-uncertainty=true}
    \renewcommand{\arraystretch}{1.1}
        \begin{tabular}{
        l
        S[table-format=3]
        S[table-format=2.2(2), separate-uncertainty]
        S[table-format=1.2(2), separate-uncertainty]
        S[table-format=1.2(2), separate-uncertainty]
        S[table-format=1.2(2), separate-uncertainty]
        S[table-format=2.2(2), separate-uncertainty]
        }
            \toprule
            \textbf{Shared Comp.} & {P.~(M)~$\downarrow$} & {$\delta_{1.25}$~(\%)$~\uparrow$} & {MSLE~($10^{-2}$)~$\downarrow$} & {MAE~(m)~$\downarrow$} & {RMSE~(m)~$\downarrow$} & {F1~(\%)~$\uparrow$} \\
            \midrule
            \rowcolor{gray!25} None & 100 & \bfmain{88.29}{0.32} & \bfmainsmall{1.85}{0.05} & \bfmainsmall{1.17}{0.02} & \bfmainsmall{1.54}{0.02} & \bfmain{86.89}{0.31} \\
            \gls{mlp} & 95 & 87.46(62) & 1.96(08) & 1.20(02) & 1.57(02) & \ulmain{86.02}{0.30} \\
            \gls{mlp}, CA & \bfseries 93 & 87.77(21) & \ulmainsmall{1.91}{0.02} & \ulmainsmall{1.19}{0.01} & \ulmainsmall{1.55}{0.01} & 85.97(42) \\
            \gls{mlp}, CA, Q & \bfseries 93 & \ulmain{87.90}{0.14} & \ulmainsmall{1.91}{0.02} & \ulmainsmall{1.19}{0.00} & \ulmainsmall{1.55}{0.00} & 85.67(24) \\
            \bottomrule
        \end{tabular}
\end{table*}

In \cref{tab:parameter_sharing_analysis}, we observe that sharing more parameters leads to a significant reduction in the F1-Score (from $\text{F1} \approx 86.9\%$ to $86.0\%$, $86.0\%$, and $85.7\%$) and a marginal reduction in the height threshold accuracy (within the standard error).
Considering the relatively moderate reduction in parameters (from 100M to 95M or 93M), we conclude that sharing more parameters is not beneficial.
For computational efficiency, it is better to reduce the backbone size, as detailed in the App.~\ref{app:backbone}.

\subsection{Analysis of Backbone Selection}
\label[appendix]{app:backbone}
\begin{table}[tb]
    \caption{
    \textbf{Backbone analysis of \method on the Quebec Trees dataset.}
    We compare various \glspl{vit} initializations (DINOv3, PECore) and sizes (ranging from `S' to `H+') utilized as the model backbone.
    We report the mean ± standard error over 5 seeds.
    \textbf{Bold} and \underline{underline} denote best and second-best.
    Gray corresponds to our design choice.
    }
    \label{tab:backbone}
    \centering
    \setlength{\tabcolsep}{3pt}
    \scriptsize
    \sisetup{table-number-alignment=center, separate-uncertainty=true, retain-zero-uncertainty=true}
    \renewcommand{\arraystretch}{1.1}
    \begin{tabular}{
        l
        S[table-format=3]
        S[table-format=2.2(2), separate-uncertainty]
        S[table-format=1.2(2), separate-uncertainty]
        S[table-format=1.2(2), separate-uncertainty]
        S[table-format=1.2(2), separate-uncertainty]
        S[table-format=2.2(2), separate-uncertainty]}
        \toprule
        \textbf{Backbone} & {P.~(M)~$\downarrow$} & {$\delta_{1.25}$~(\%)$~\uparrow$} & {MSLE~($10^{-2}$)~$\downarrow$} & {MAE~(m)~$\downarrow$} & {RMSE~(m)~$\downarrow$} & {F1~(\%)~$\uparrow$} \\
        \midrule
        PECore-B & 107 & 86.55(31) & 2.07(05) & 1.23(01) & 1.60(01) & 85.23(25) \\
        PECore-L & 342 & 88.38(42) & 1.85(05) & 1.18(02) & 1.53(02) & 87.40(49) \\
        DINOv3-S & \bfseries 25 & 87.61(41) & 1.95(03) & 1.18(01) & 1.55(01) & 84.51(38) \\
        DINOv3-S+ & \ulinttwo{32} & 87.41(56) & 1.98(07) & 1.19(03) & 1.56(03) & 84.46(52) \\
        \rowcolor{gray!25} DINOv3-B & 100 & 88.29(32) & 1.85(05) & 1.17(02) & 1.54(02) & 86.89(31) \\
        DINOv3-L & 328 & \ulmain{89.28}{0.28} & \ulmainsmall{1.71}{0.04} & \ulmainsmall{1.12}{0.02} & \ulmainsmall{1.46}{0.02} & \bfmain{88.38}{0.38} \\
        DINOv3-H+ & 880 & \bfmain{90.52}{0.27} & \bfmainsmall{1.56}{0.02} & \bfmainsmall{1.08}{0.01} & \bfmainsmall{1.42}{0.74} & \ulmain{88.15}{0.40} \\
        \bottomrule
    \end{tabular}
\end{table}
\cref{tab:backbone} compares various backbones for \method on the Quebec Trees dataset. 
Scaling up the DINOv3 backbone improves both height estimation and classification, whereas the smaller variants significantly reduce the overall number of parameters while maintaining competitive results.
Replacing DINOv3 with an equivalent-sized PECore backbone degrades predictions: for the Base model, $\delta_{1.25}$ drops from $88.29\%$ to $86.55\%$, and the F1-score falls from $86.89\%$ to $85.23\%$.
This ablation justifies the choice of DINOv3 as the backbone of \method.

\subsection{Separate Model Analysis}
\label[appendix]{app:separate}
\begin{table}[tb]
    \caption{
    \textbf{Separate model of \method analysis on the Quebec Trees dataset.}
    The table compares the use of separate models for the classification and height estimations tasks against a joint approach.
    We denote the joint model for both tasks as `JM' (equivalent to \method) and the deployment of two separate models as `SM'.
    For `SM', `P.' denotes combined parameters of both models.
    We report the mean ± standard error over 5 seeds.
    \textbf{Bold} and \underline{underline} denote best and second-best.
    Gray corresponds to our design choice.
    }
    \label{tab:separate_model}
    \centering
    \setlength{\tabcolsep}{3pt}
    \scriptsize
    \sisetup{table-number-alignment=center, separate-uncertainty=true, retain-zero-uncertainty=true}
    \renewcommand{\arraystretch}{1.1}
    \begin{tabular}{
        l
        S[table-format=3]
        S[table-format=2.2(2), separate-uncertainty]
        S[table-format=1.2(2), separate-uncertainty]
        S[table-format=1.2(2), separate-uncertainty]
        S[table-format=1.2(2), separate-uncertainty]
        S[table-format=2.2(2), separate-uncertainty]}
        \toprule
        \textbf{Variant} & {P.~(M)~$\downarrow$} & {$\delta_{1.25}$~(\%)$~\uparrow$} & {MSLE~($10^{-2}$)~$\downarrow$} & {MAE~(m)~$\downarrow$} & {RMSE~(m)~$\downarrow$} & {F1~(\%)~$\uparrow$} \\
        \midrule
        \rowcolor{gray!25} JM, B & \bfseries 100 & 88.29(32) & 1.85(05) & 1.17(02) & 1.54(02) & 86.89(31) \\
        SM, B & \ulint{186} & \ulmain{88.47}{0.20} & \ulmainsmall{1.83}{0.02} & \ulmainsmall{1.16}{0.01} & \ulmainsmall{1.52}{0.01} & \ulmain{87.53}{0.45} \\
        \rowcolor{gray!25} JM, L & 328 & \bfmain{89.28}{0.28} & \bfmainsmall{1.71}{0.04} & \bfmainsmall{1.12}{0.02} & \bfmainsmall{1.46}{0.02} & \bfmain{88.38}{0.38} \\
        \bottomrule
    \end{tabular}
\end{table}
We compare the \method architecture trained jointly on both tasks (height estimation and classification) against two separate models trained individually for each task, following the protocol for competing methods in \cref{subsec:comp}.
The results in \cref{tab:separate_model} reveal only a marginal improvement (within the standard error) for the two separate versions of \method-B.
However, considering that the combined parameter count of these two separate models falls almost halfway between the Base-sized and Large-sized joint models, it is more effective to increase the backbone size to Large, which leads to significantly better results.
In summary, employing a single joint model for both height estimation and classification yields better performance per parameter.

\subsection{Analysis of Results}
\label[appendix]{app:results_analysis}
In this section, we analyze the height estimation and classification results of \method, structuring our analysis by dataset.

\paragraph{Quebec Trees dataset.}
\cref{fig:true_predicted_height} demonstrates that \method successfully learns to predict individual tree height, rather than merely approximating the mean height of each class.
\begin{figure}[!tb]
    \centering
    \begin{minipage}{0.48\linewidth}
        \centering
        \includegraphics[width=\linewidth]{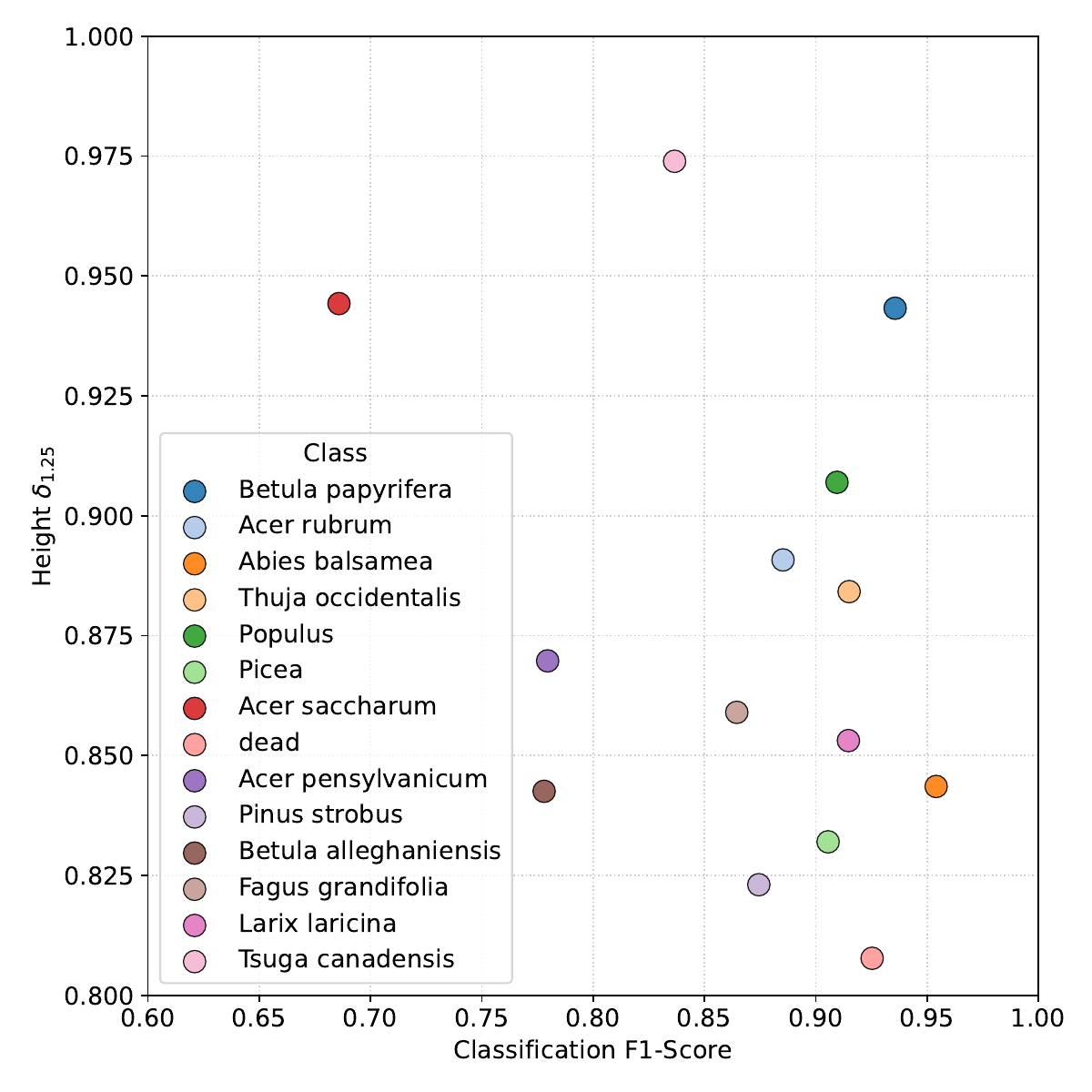}
        \caption{
        \textbf{Comparison of classification and height estimation results} using \method-B on the Quebec Trees dataset.
        We report the F1-Score and threshold accuracy $\delta_{1.25}$ per class averaged over 5 seeds.
        }
        \label{fig:cls_height_qt}
    \end{minipage}
    \hfill
    \begin{minipage}{0.48\linewidth}
        \centering
        \includegraphics[width=\linewidth]{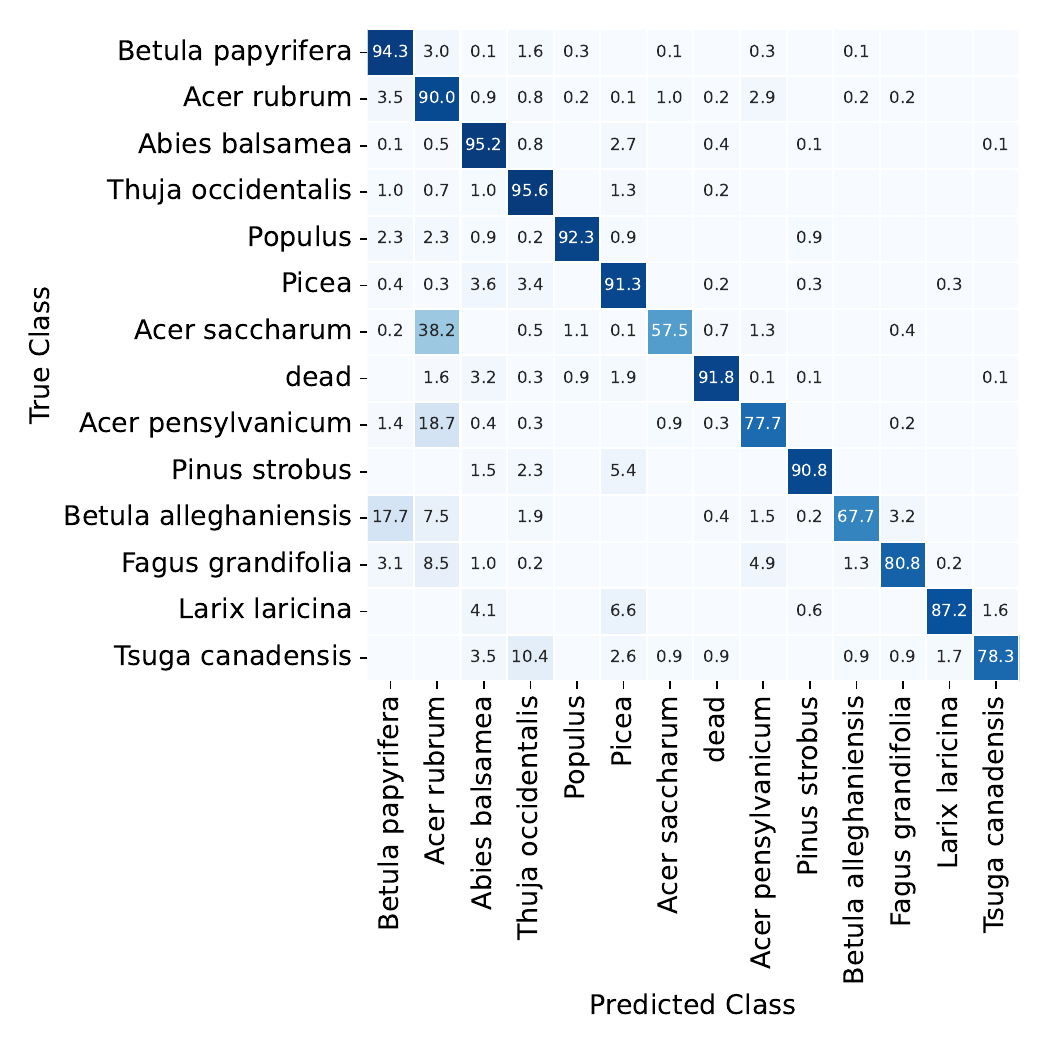}
        \caption{
        \textbf{Confusion matrix of \method-B on Quebec Trees.}
        Predicted and true labels are aggregated across 5 seeds to ensure reliability.
        All entries are row-normalized (per true class) and displayed as percentages.
        Cells with value rounding to $0.0\%$ are left blank for readability.
        }
        \label{fig:cm_qt}
    \end{minipage}
\end{figure}
In \cref{fig:cls_height_qt}, we scatter the mean threshold accuracy for height against the mean F1-Score for classification per class.
The results indicate that while dead trees are easily recognized, estimating their height is particularly challenging.
This difficulty arises from their varying orientations (fallen vs. standing) and narrow treetop profiles, which complicate the identification of the highest point.
Furthermore, the ground truth heights, extracted via the pipeline in \cref{subsec:extraction}, may inherently be less accurate for such thin instance segmentations and treetop occlusions.

Trees within the `Acer' genus (\textit{Acer rubrum}, \textit{Acer saccharum}, and \textit{Acer pensylvanicum}) tend to be recognized more reliably than species from other genera, likely due to the high number of training samples available for these classes.
\cref{fig:cm_qt} displays the normalized confusion matrix of \method-B on the Quebec Trees dataset.
The primary classification errors stem from intra-genus confusion: $38.2\%$ of \textit{Acer saccharum} trees are predicted as \textit{Acer rubrum}, $18.7\%$ of \textit{Acer pensylvanicum} as \textit{Acer rubrum}, and $17.7\%$ of \textit{Betula alleghaniensis} as \textit{Betula papyrifera}.
Due to their highly similar visual appearance, these species are difficult to distinguish on remote sensing imagery even for expert biologists.

\paragraph{BCI dataset.}
\begin{figure}[!tb]
    \centering
    \begin{subfigure}[b]{0.48\linewidth}
        \centering
        \includegraphics[width=\linewidth]{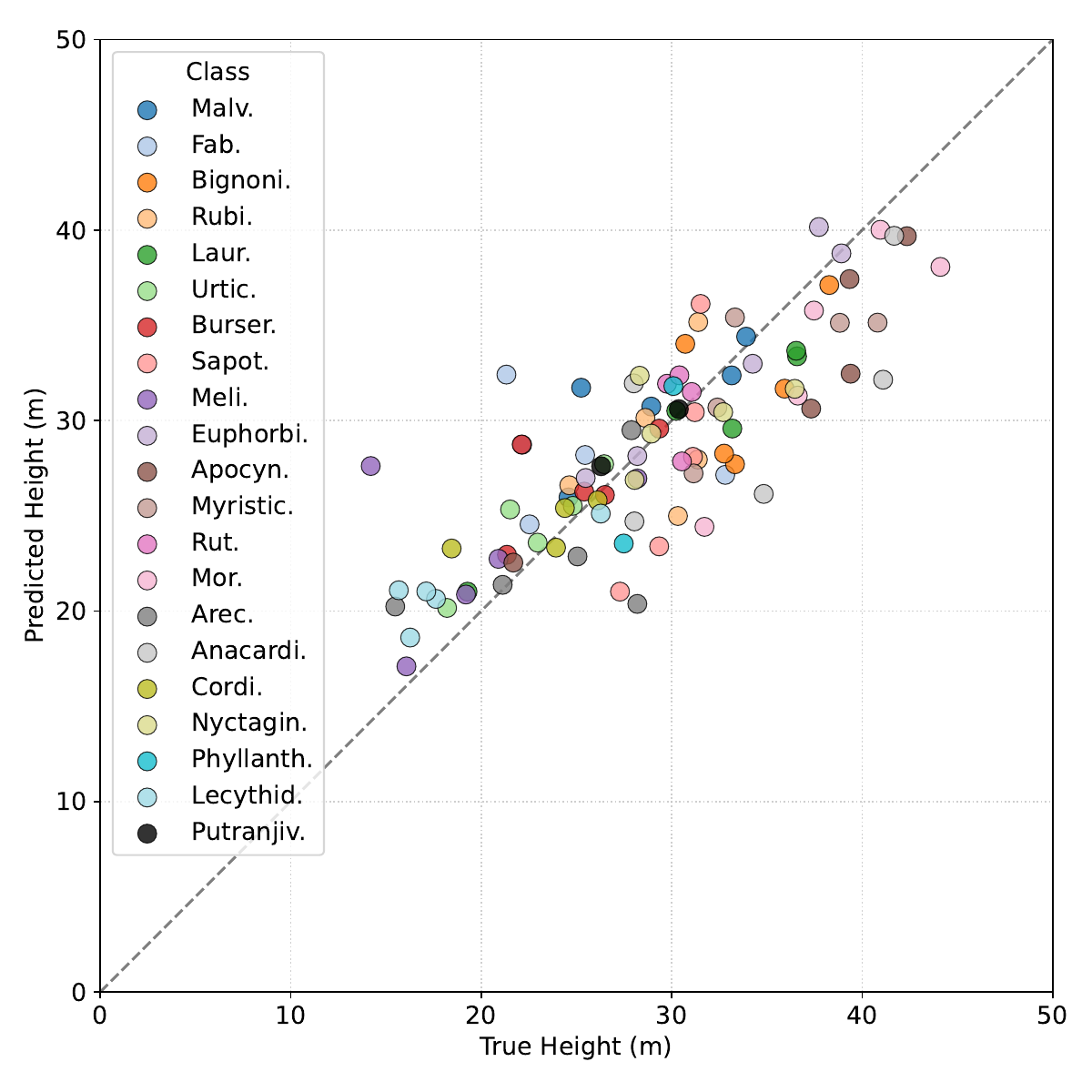}
        \caption{BCI}
        \label{fig:true_predicted_height_bci}
    \end{subfigure}
    \hfill
    \begin{subfigure}[b]{0.48\linewidth}
        \centering
        \includegraphics[width=\linewidth]{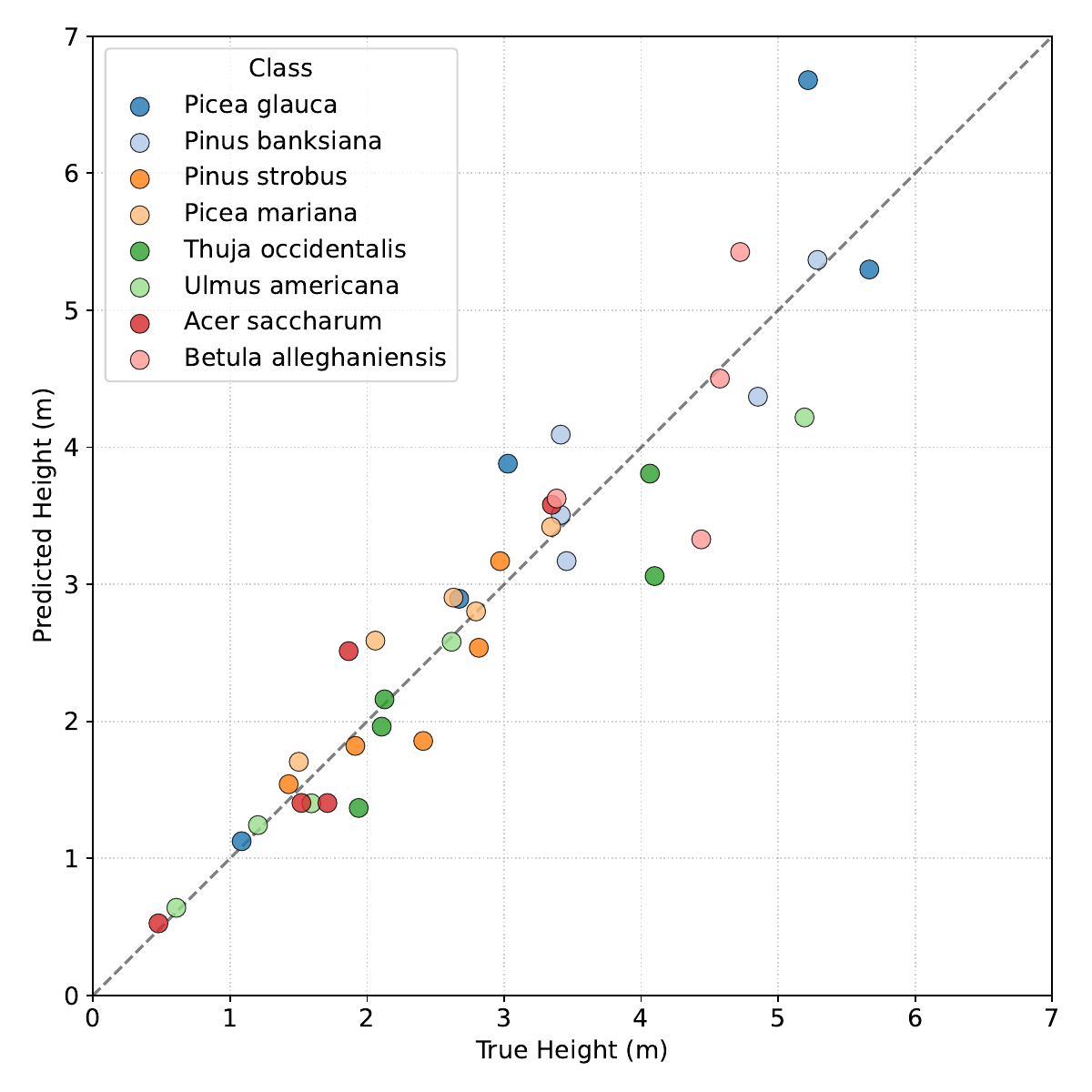}
        \caption{Quebec Plantations}
        \label{fig:true_predicted_height_qp}
    \end{subfigure}
    \caption{
    \textbf{Comparison of true vs. predicted height}
    using \method-B on the BCI and Quebec Plantations test splits. 
    We select the random seed yielding the highest $\delta_{1.25}$ and visualize up to 5 samples per class.
    The dashed diagonal line represents the ideal prediction ($y=x$).
    \label{fig:true_predicted_height_qp_bci}}
\end{figure}
Similar to the observations for the Quebec Trees dataset, \cref{fig:true_predicted_height_bci} illustrates that, on average, \method overestimates the height of small trees and underestimates the height of large trees.
\begin{figure}[!tb]
    \centering
    \begin{minipage}{0.48\linewidth}
        \centering
        \includegraphics[width=\linewidth]{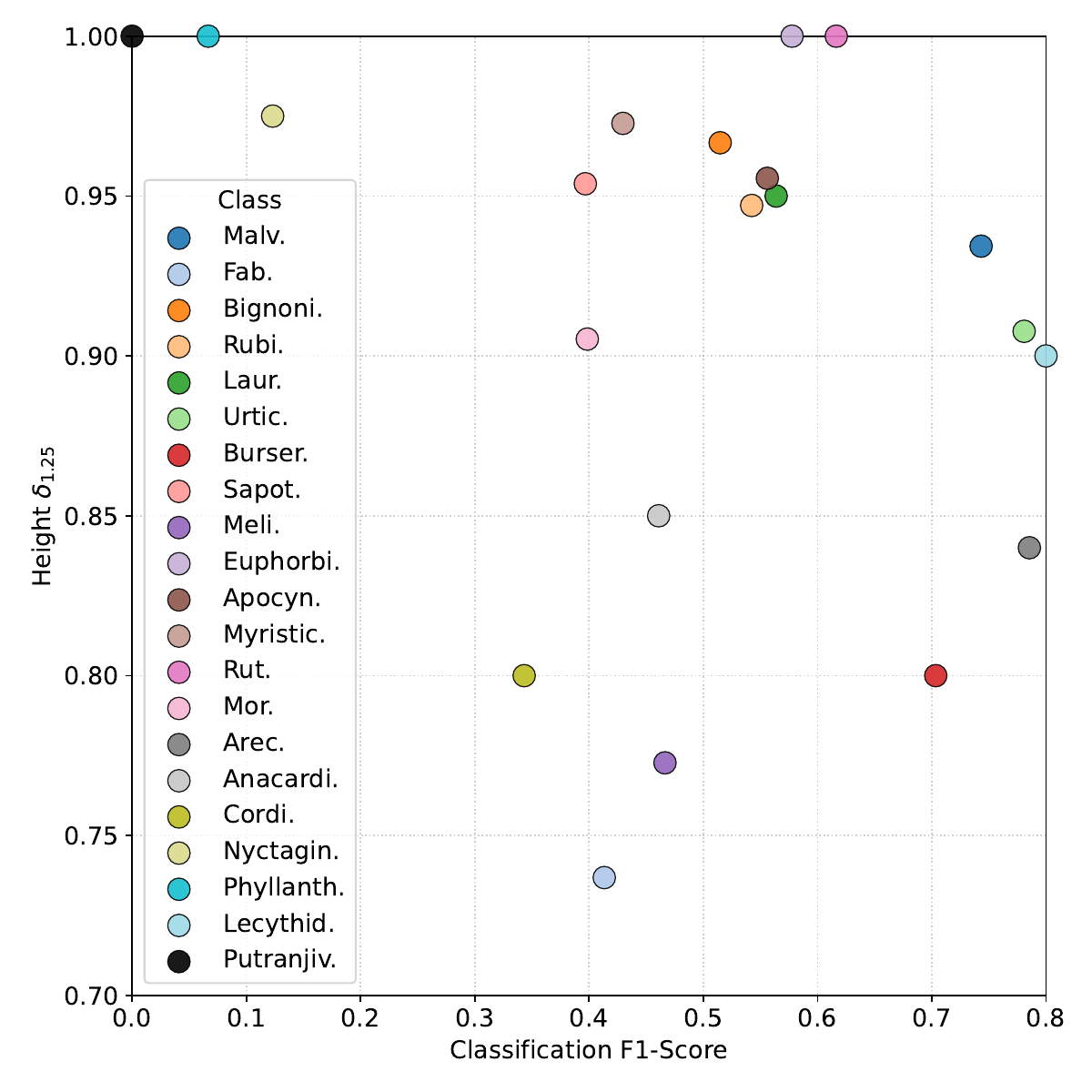}
        \caption{
        \textbf{Comparison of classification and height estimation results} using \method-B on the BCI dataset.
        We report the F1-Score and threshold accuracy $\delta_{1.25}$ per class averaged over 5 seeds.
        }
        \label{fig:cls_height_bci}
    \end{minipage}
    \hfill
    \begin{minipage}{0.48\linewidth}
        \centering
        \includegraphics[width=\linewidth]{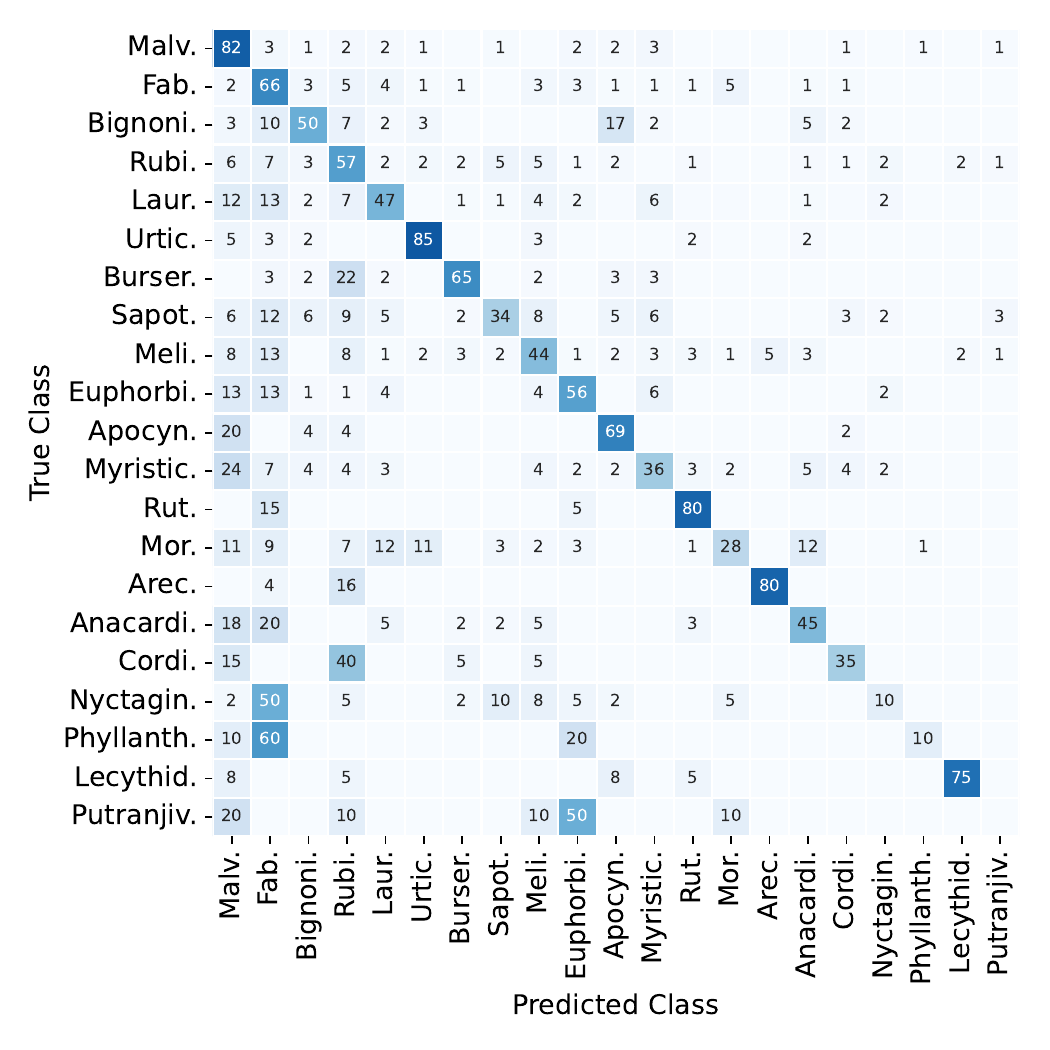}
        \caption{
        \textbf{Confusion matrix of \method-B on BCI.}
        Predicted and true labels are aggregated across 5 seeds to ensure reliability.
        All entries are row-normalized (per true class) and displayed as percentages.
        Cells with value rounding to $0\%$ are left blank for readability.
        }
        \label{fig:cm_bci}
    \end{minipage}
\end{figure}
Furthermore, \cref{fig:cls_height_bci} indicates no strong correlation between height estimation and classification results per tree class.
For example, the class `Putranjivaceae' achieves a threshold accuracy of $100\%$, while having a F1-Score of $0\%$.
Conversely, the class `Rutaceae' achieves the same $100\%$ threshold accuracy while maintaining a high F1-Score ($\text{F1} > 60\%$).
The confusion matrix of \method-B on BCI (\cref{fig:cm_bci}) reveals that recall (the diagonal values) generally decreases for less frequent classes.
Consequently, classes with few training images are often misclassified as instances of more frequent classes.
For instance, $50\%$ of `Nyctaginaceae' trees and $60\%$ of `Phyllanthaceae' trees (both among the four least frequent classes) are incorrectly classified as `Fab.' trees, which is the second most frequent class in the training set.
\clearpage

\paragraph{Quebec Plantations dataset.}
\begin{figure}[!tb]
    \centering
    \begin{minipage}{0.48\linewidth}
        \centering
        \includegraphics[width=\linewidth]{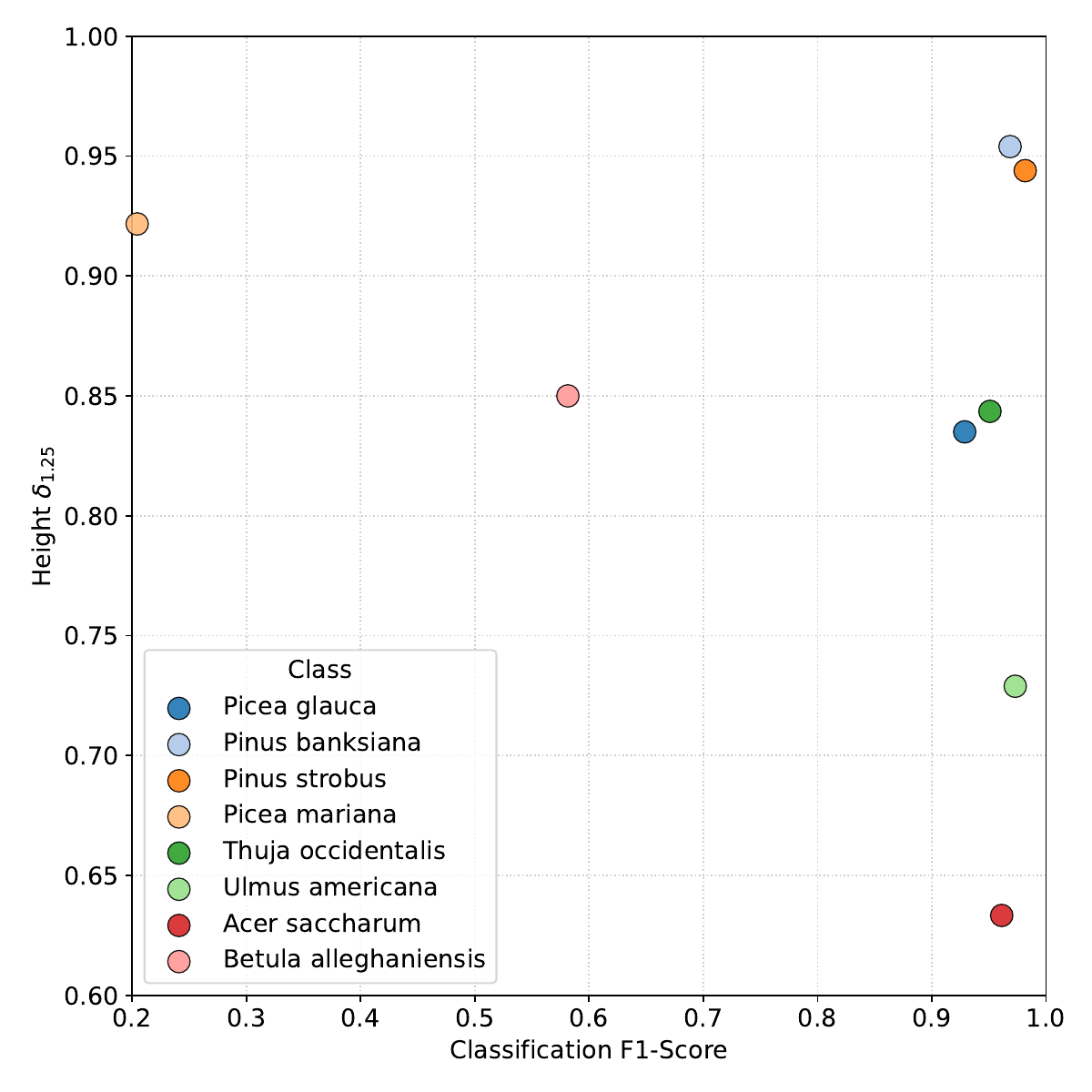}
        \caption{
        \textbf{Comparison of classification and height estimation results} using \method-B on the Quebec Plantations dataset.
        We report the F1-Score and threshold accuracy $\delta_{1.25}$ per class averaged over 5 seeds.
        }
        \label{fig:cls_height_qp}
    \end{minipage}
    \hfill
    \begin{minipage}{0.48\linewidth}
        \centering
        \includegraphics[width=\linewidth]{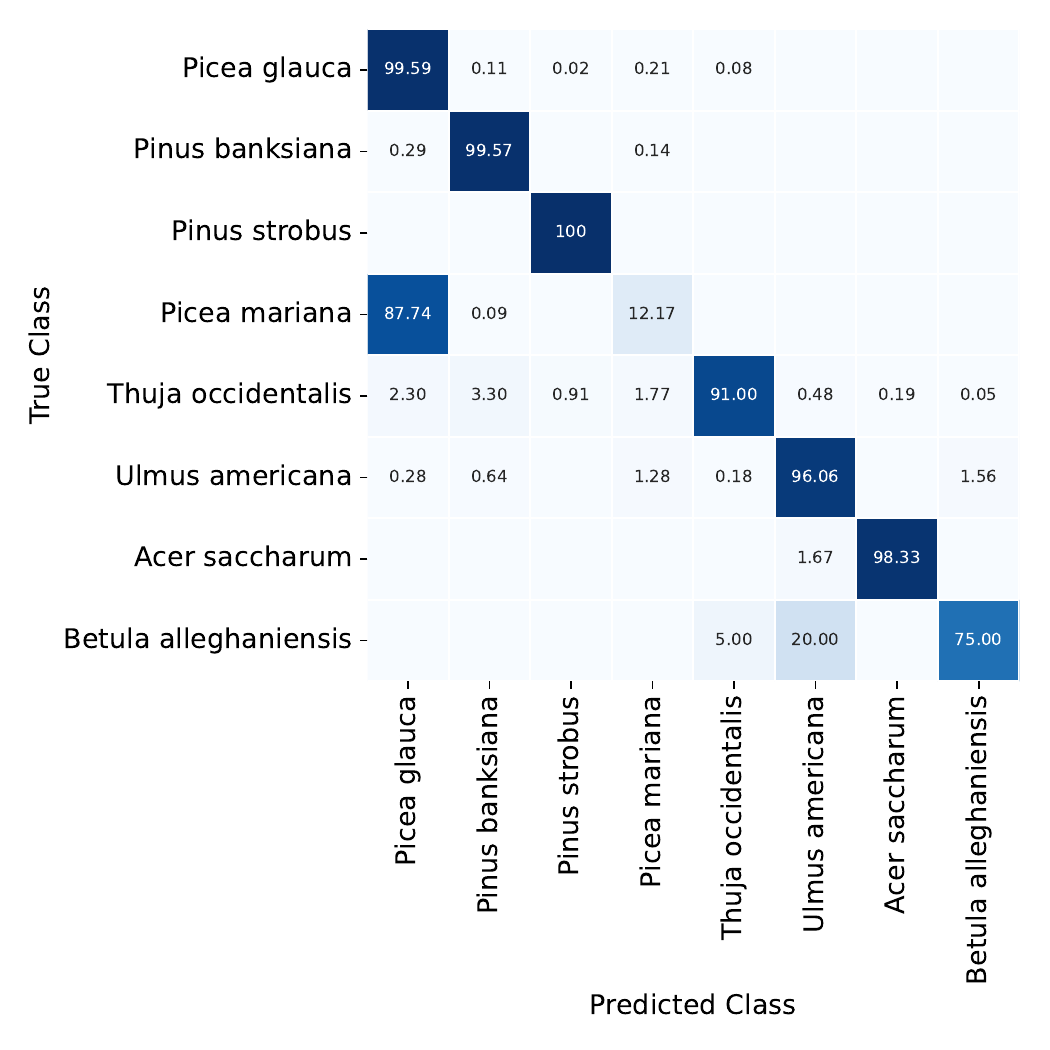}
        \caption{
        \textbf{Confusion matrix of \method-B on Quebec Plantations.}
        Predicted and true labels are aggregated across 5 seeds to ensure reliability.
        All entries are row-normalized (per true class) and displayed as percentages.
        Cells with value rounding to $0.00\%$ are left blank for readability.
        }
        \label{fig:cm_qp}
    \end{minipage}
\end{figure}
Consistent with the negligibly small average MSD for Quebec Plantations (\cref{tab:height_bias}), \cref{fig:true_predicted_height_qp} displays no height- or species-specific over- or underestimation trends.
\cref{fig:cls_height_qp} reveals that the classification results are worst for \textit{Picea mariana}, while \textit{Acer saccharum} exhibits the most inaccurate height predictions.
On average, `Pinus' trees (\textit{Pinus banksiana} and \textit{Pinus strobus}) yield the most accurate height predictions ($\delta_{1.25} \approx 95\%$).
In the confusion matrix of \method-B (\cref{fig:cm_qp}) we observe that recall is higher than $90\%$ for all species except \textit{Picea mariana} and \textit{Betula alleghaniensis}.
Instances of \textit{Picea mariana} are predominately misclassified as \textit{Picea glauca}, a more frequent species within the same genus that shares a similar visual appearance.
Furthermore, \textit{Betula alleghaniensis} is difficult to classify due severe class imbalance, with only 11 example images in the training set (representing less than $0.1\%$ of all training samples).
Consequently, this species is occasionally misclassified as \textit{Ulmus americana} or \textit{Thuja occidentalis}.

\end{document}